\documentclass{article}

\usepackage{iclr2017_arXiv,times}
\usepackage{hyperref}
\usepackage{url}

\usepackage{tabularx}
\usepackage{multirow}

\usepackage{algorithm2e}

\usepackage{graphicx} 

\usepackage{bm}
\usepackage{amsmath}

\DeclareMathOperator*{\argmax}{arg\,max}

\usepackage{xcolor}

\SetCommentSty{mycommfont}

\usepackage{hyperref}
\hypersetup{
    colorlinks,
    citecolor=black,
    filecolor=black,
    linkcolor=black,
    urlcolor=black
}

\usepackage{multirow}

\usepackage{amssymb} 
\usepackage{rotating} 

\title{Deep Reinforcement Learning: An Overview}

\author{
  Yuxi Li (yuxili@gmail.com)\\
  }

\begin{document} 

\maketitle

\begin{abstract} 

We give an overview of recent exciting achievements of deep reinforcement learning (RL). 
We discuss six core elements, six important mechanisms, and twelve applications.
We start with background of machine learning, deep learning and reinforcement learning.
Next we discuss core RL elements, including
value function, in particular, Deep Q-Network (DQN),
policy,
reward,
model and planning,
exploration,
and knowledge.
After that, we discuss important mechanisms for RL, including
attention and memory,
unsupervised learning,
transfer learning,
multi-agent RL,
hierarchical RL, 
and learning to learn.
Then we discuss various applications of RL, 
including games, in particular, AlphaGo,
robotics,
natural language processing, including
dialogue systems,
machine translation,
and text generation,
computer vision,
business management,
finance,
healthcare, 
education,
Industry 4.0,
smart grid,
intelligent transportation systems,
and computer systems.
We mention topics not reviewed yet,
and list a collection of RL resources.
After presenting a brief summary,
we close with discussions.


This is the first overview about deep reinforcement learning publicly available online. It is comprehensive. Comments and criticisms are welcome. (This particular version is incomplete.)

\vspace{3mm}

\textbf{Please see 
Deep Reinforcement Learning, 
\url{https://arxiv.org/abs/1810.06339},
for a significant update to this manuscript.}



\end{abstract}   

\newpage

\tableofcontents

\newpage

\section{Introduction}
\label{introduction}

Reinforcement learning (RL) is about an agent interacting with the environment, learning an optimal policy, by trial and error, for sequential decision making problems in a wide range of fields in both natural and social sciences, and engineering~\citep{Sutton98, Sutton2018, Bertsekas96, Bertsekas12, Szepesvari2010, Powell11}. 

The integration of reinforcement learning and neural networks has a long history~\citep{Sutton2018,Bertsekas96, Schmidhuber2015-DL}. With recent exciting achievements of deep learning~\citep{LeCun2015, Goodfellow2016}, benefiting from big data, powerful computation, new algorithmic techniques, mature software packages and architectures, and strong financial support, we have been witnessing the renaissance of reinforcement learning~\citep{Krakovsky2016}, especially, the combination of deep neural networks and reinforcement learning, i.e., deep reinforcement learning (deep RL). 

Deep learning, or deep neural networks, has been prevailing in reinforcement learning in the last several years, in  games, robotics, natural language processing, etc. We have been witnessing breakthroughs, like deep Q-network~\citep{Mnih-DQN-2015} and AlphaGo~\citep{Silver-AlphaGo-2016}; and novel architectures and applications, like differentiable neural computer~\citep{Grave-DNC-2016}, asynchronous methods~\citep{Mnih-A3C-2016}, dueling network architectures~\citep{Wang-Dueling-2016}, value iteration networks~\citep{Tamar2016}, unsupervised reinforcement and auxiliary learning~\citep{Jaderberg2017, Mirowski2017}, neural architecture design~\citep{Zoph2017}, dual learning for machine translation~\citep{He2016}, spoken dialogue systems~\citep{Su2016}, information extraction~\citep{Narasimhan2016},  guided policy search~\citep{Levine2016}, and generative adversarial imitation learning~\citep{Ho2016}, etc. Creativity would push the frontiers of deep RL further with respect to core elements, mechanisms, and applications.

Why has deep learning been helping reinforcement learning make so many and so enormous achievements? Representation learning with deep learning enables automatic feature engineering and end-to-end learning through gradient descent, so that reliance on domain knowledge is significantly reduced or even removed. Feature engineering used to be done manually and is usually time-consuming,  over-specified, and incomplete.  Deep, distributed representations exploit the hierarchical composition of factors in data to combat the exponential challenges of the curse of dimensionality. 
 Generality, expressiveness and flexibility of deep neural networks  make some tasks easier or possible, e.g., in the breakthroughs and novel architectures and applications discussed above. 
 
 Deep learning, as a specific class of machine learning, is not without limitations, e.g., as a black-box lacking interpretability, as an "alchemy" without clear and sufficient scientific principles to work with, and without human intelligence not able to competing with a baby in some tasks. However, there are lots of works to improve deep learning, machine learning, and AI in general.

Deep learning  and reinforcement learning, being selected as one of the  MIT Technology Review 10 Breakthrough Technologies in 2013 and 2017 respectively,  will play their crucial role in achieving artificial general intelligence. David Silver, the major contributor of AlphaGo~\citep{Silver-AlphaGo-2016, Silver-AlphaGo-2017}, even made a formula:  artificial intelligence = reinforcement learning + deep learning~\citep{Silver2016Tutorial}. 

The outline of this overview follows. First we discuss background of machine learning, deep learning and reinforcement learning in Section~\ref{background}.
Next we discuss core RL elements, including
value function in Section~\ref{value}, 
policy in Section~\ref{policy}, 
reward in Section~\ref{reward},
model and planning in Section~\ref{modelplanning},
exploration in Section~\ref{exploration},
and knowledge in Section~\ref{knowledge}.
Then we discuss important mechanisms for RL, including
attention and memory  in Section~\ref{attention}, 
unsupervised learning in Section~\ref{unsupervised},
transfer learning in Section~\ref{transfer},
multi-agent RL in Section~\ref{MARL},
hierarchical RL in Section~\ref{hierarchical}, and, 
learning to learn in Section~\ref{learning2learn}.
After that, we discuss various RL applications, including
games in Section~\ref{games}, 
robotics in Section~\ref{robotics},
natural language processing in Section~\ref{NLP},
computer vision in Section~\ref{CV},
business management in Section~\ref{business},
finance  in Section~\ref{fin}, 
healthcare in Section~\ref{healthcare},
education in Section~\ref{education},
Industry 4.0 in Section~\ref{Industry40},
smart grid  in Section~\ref{smartgrid},
intelligent transportation systems  in Section~\ref{ITS},
and computer systems in Section~\ref{systems}.
We present a list of topics not reviewed yet in Section~\ref{todo},
give a brief summary in Section~\ref{summary},
and close with discussions in Section~\ref{discussion}.



In Section~\ref{resources}, we list a collection of RL resources including books, surveys, reports, online courses,  tutorials, conferences, journals and workshops, blogs, and open sources.  If picking a single RL resource, it is Sutton and Barto's RL book~\citep{Sutton2018},  2nd edition in preparation. It covers RL fundamentals and reflects new progress, e.g., in deep Q-network, AlphaGo, policy gradient methods, as well as in psychology and neuroscience. 
\citet{Deng2014} and \citet{Goodfellow2016} are recent deep learning books.
\citet{Bishop2011},  \citet{Hastie2009}, and \citet{Murphy2012} are popular machine learning textbooks;
\citet{James2013} gives an introduction to machine learning;
\citet{Provost2013} and \citet{Kuhn2013}  discuss practical issues in machine learning applications; and 
\citet{Simeone2017} is a brief introduction to machine learning for engineers.

\begin{figure}
\includegraphics[width=1.0\linewidth]{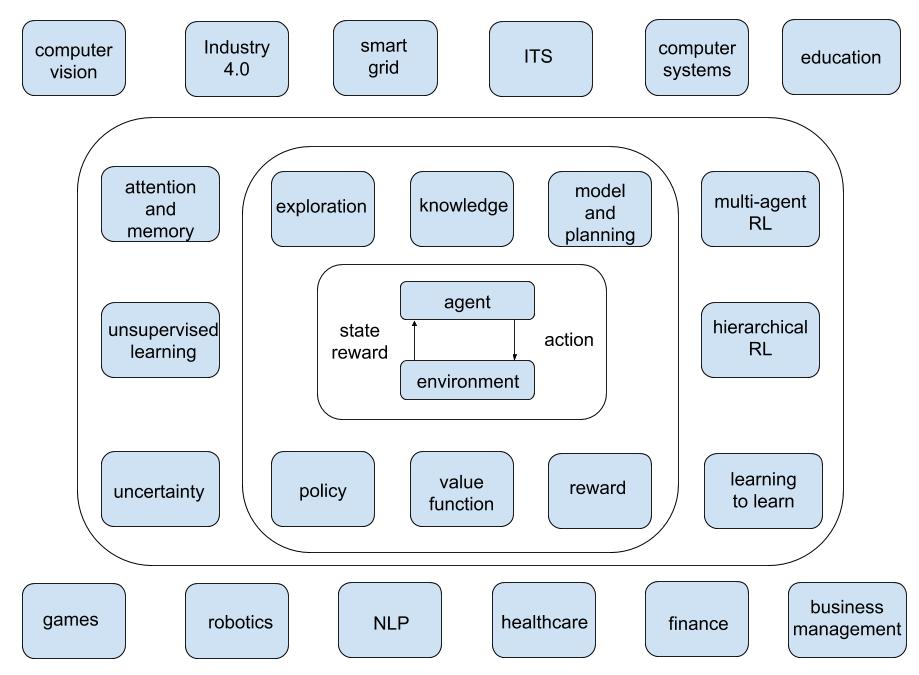}
\caption{Conceptual Organization of the Overview}
\label{organization}
\end{figure}

Figure~\ref{organization} illustrates the conceptual organization of the overview. The agent-environment interaction sits in the center, around which are core elements: value function, policy, reward, model and planning, exploration, and knowledge. Next come important mechanisms: attention and memory, unsupervised learning, transfer learning, multi-agent RL, hierarchical RL, and learning to learn. Then come various applications: games, robotics, NLP (natural language processing),  computer vision, business management, finance, healthcare, education, Industry 4.0, smart grid, ITS (intelligent transportation systems), and computer systems.

The main readers of this overview would be those who want to get more familiar with deep reinforcement learning.
We endeavour to provide as much relevant information as possible.  
For reinforcement learning experts, as well as new comers, we hope this overview would be helpful as a reference.
In this overview, we mainly focus on contemporary work in recent couple of years, by no means complete, and make slight effort for discussions of historical context, for which the best material to consult is~\citet{Sutton2018}. 

In this version, we endeavour to provide a wide coverage of fundamental and contemporary RL issues, about core elements, important mechanisms, and applications. In the future, besides further refinements for the width, we will also improve the depth by conducting deeper analysis of the issues involved and the papers discussed. Comments and criticisms are welcome.


\section{Background}
\label{background}

In this section, we briefly introduce  concepts and fundamentals in machine learning, deep learning~\citep{Goodfellow2016} and reinforcement learning~\citep{Sutton2018}. We do not give detailed background introduction for machine learning and deep learning. Instead, we recommend the following recent Nature/Science survey papers: \citet{Jordan2015} for machine learning, and \citet{LeCun2015} for deep learning. We cover some RL basics. However,  we recommend the textbook, \citet{Sutton2018}, and the recent Nature survey paper, \citet{Littman2015}, for reinforcement learning. We also collect relevant resources in Section~\ref{resources}.

\subsection{Machine Learning}
\label{machinelearning}

Machine learning is about learning from data and making predictions and/or decisions. 

Usually we categorize machine learning as supervised, unsupervised, and reinforcement learning.\footnote{Is reinforcement learning part of machine learning, or more than it, and somewhere close to artificial intelligence? We raise this question without elaboration.} In supervised learning, there are labeled data; in unsupervised learning, there are no labeled data; and in reinforcement learning, there are evaluative feedbacks, but no supervised signals. Classification and regression are two types of supervised learning problems, with categorical and numerical outputs respectively.

Unsupervised learning attempts to extract information from data without labels, e.g., clustering and density estimation. Representation learning is a classical type of unsupervised learning. However, training  feedforward networks or convolutional neural networks with supervised learning is a kind of representation learning.
Representation learning finds a representation to preserve as much information about the original data as possible, at the same time, to keep the representation simpler or more accessible than the original data, with low-dimensional, sparse, and independent representations. 

Deep learning, or deep neural networks, is a particular machine learning scheme, usually for supervised or unsupervised learning, and can be integrated with reinforcement learning, usually as a function approximator.  
Supervised and unsupervised learning are usually one-shot, myopic, considering instant reward; while reinforcement learning is sequential, far-sighted, considering long-term accumulative reward. 

Machine learning is  based on probability theory and statistics~\citep{Hastie2009} and optimization~\citep{Boyd04}, is the basis for big data, data science~\citep{Blei2017,Provost2013}, predictive modeling~\citep{Kuhn2013}, data mining, information retrieval~\citep{Manning2008}, etc, and becomes a critical ingredient for computer vision, natural language processing, robotics, etc. Reinforcement learning is kin to optimal control~\citep{Bertsekas12}, and operations research and management~\citep{Powell11}, and is also related to psychology and neuroscience~\citep{Sutton2018}. Machine learning is a subset of artificial intelligence (AI), and is evolving to be critical for all fields of AI.


A machine learning algorithm is composed of a dataset, a cost/loss function, an optimization procedure, and a model~\citep{Goodfellow2016}. 
A dataset is divided into non-overlapping training, validation, and testing subsets. 
A cost/loss function measures the model performance, e.g., with respect to accuracy, like mean square error in regression and classification error rate. 
Training error measures the error on the training data, minimizing which is an optimization problem.
Generalization error, or test error, measures the error on new input data, which differentiates machine learning from optimization.
A machine learning algorithm tries to make the training error, and the gap between training error and testing error small.
A model is under-fitting if it can not achieve a low training error; a model is over-fitting if the gap between training error and test error is large.

A model's capacity measures the range of functions it can fit.
VC dimension measures the capacity of a binary classifier.
Occam's Razor states that, with the same expressiveness, simple models are preferred. 
Training error and generalization error versus model capacity usually form a U-shape relationship. We find the optimal capacity to achieve low training error and small gap between training error and generalization error.
Bias measures the expected deviation of the estimator from the true value; while variance measures the deviation of the estimator from the expected value, or variance of the estimator. 
As model capacity increases, bias tends to decrease, while variance tends to increase, yielding another U-shape relationship between generalization error versus model capacity. 
We try to find the optimal capacity point, of which under-fitting occurs on the left and over-fitting occurs on the right. 
Regularization add a penalty term to the cost function, to reduce the generalization error, but not training error.
No free lunch theorem states that there is no universally best model, or best regularizor. 
An implication is that deep learning may not be the best model for some problems.
There are model parameters, and hyperparameters for model capacity and regularization.
Cross-validation is used to tune hyperparameters, to strike a balance between bias and variance, and to select the optimal model.

Maximum likelihood estimation (MLE) is a common approach to derive good estimation of parameters. 
For issues like numerical underflow, the product in MLE is converted to summation to obtain negative log-likelihood (NLL). 
MLE is equivalent to minimizing KL divergence, the dissimilarity between the empirical distribution defined by the training data and the model distribution.
Minimizing KL divergence between two distributions corresponds to minimizing the cross-entropy between the distributions. 
In short, maximization of likelihood becomes minimization of the negative log-likelihood (NLL), or equivalently, minimization of cross entropy.

Gradient descent is a common approach to solve optimization problems. 
Stochastic gradient descent extends gradient descent by working with a single sample each time, and usually with minibatches.

Importance sampling is a technique to estimate properties of a particular distribution, by samples from a different distribution, to lower the variance of the estimation, or when sampling from the distribution of interest is difficult.

Frequentist statistics estimates a single value, and characterizes variance by confidence interval; Bayesian statistics considers the distribution of an estimate when making predictions and decisions. 

generative vs discriminative

\subsection{Deep Learning}
\label{deeplearning}

Deep learning is in contrast to "shallow" learning. For many machine learning algorithms, e.g., linear regression, logistic regression, support vector machines (SVMs), decision trees, and boosting, we have input layer and output layer, and the inputs may be transformed with manual feature engineering before training. In deep learning, between input and output layers, we have one or more hidden layers. At each layer except input layer, we compute the input to each unit, as the weighted sum of units from the previous layer; then we usually use nonlinear transformation, or activation function, such as logistic, tanh, or more popular recently, rectified linear unit (ReLU), to apply to the input of a unit, to obtain a new representation of the input from previous layer. We have weights on links between units from layer to layer. After computations flow forward from input to output, at output layer and each hidden layer, we can compute error derivatives backward, and backpropagate gradients towards the input layer, so that weights can be updated to optimize some loss function.  

A feedforward deep neural network or multilayer perceptron (MLP) is to map a set of input values to output values
with a mathematical function formed by composing many simpler functions at each layer.  A convolutional neural network (CNN) is a feedforward deep neural network, with convolutional layers, pooling layers and fully connected layers. CNNs are designed to process data with multiple arrays, e.g., colour image, language, audio spectrogram, and video, benefit from the properties of such signals: local connections, shared weights, pooling and the use of many layers, and are inspired by simple cells and complex cells in visual neuroscience~\citep{LeCun2015}.  
ResNets~\citep{He2016-ResNets} are designed to ease the training of very deep neural networks by adding shortcut connections to learn residual functions with reference to the layer inputs.
A recurrent neural network (RNN) is often used to process sequential inputs like speech and language, element by element, with hidden units to store history of past elements. A RNN can be seen as a multilayer neural network with all layers sharing the same weights, when being unfolded in time of forward computation. It is hard for RNN to store information for very long time and the gradient may vanish. 
Long short term memory networks (LSTM)~\citep{Hochreiter1997} and gated recurrent unit (GRU)~\citep{Chung2014}  were proposed to address such issues, with gating mechanisms to manipulate information through recurrent cells. Gradient backpropagation or its variants can be used for training all deep neural networks mentioned above.


Dropout~\citep{Srivastava2014} is a regularization strategy to train an ensemble of sub-networks by removing non-output units randomly from the original network. Batch normalization~\citep{Ioffe2015} performs the normalization for each training mini-batch, to accelerate training by reducing internal covariate shift, i.e., the change of parameters of previous layers will change each layer's inputs distribution.

Deep neural networks learn representations automatically from raw inputs to recover the compositional hierarchies in many natural signals, i.e., higher-level features are composed of lower-level ones, e.g., in images, the hierarch of objects, parts, motifs, and local combinations of edges. Distributed representation is a central idea in deep learning, which implies that many features may represent each input, and each feature may represent many inputs. The exponential advantages of deep, distributed representations combat the exponential challenges of the curse of dimensionality. The notion of end-to-end training refers to that a learning model uses raw inputs without manual feature engineering to generate outputs, e.g., AlexNet~\citep{Krizhevsky2012} with raw pixels for image classification, Seq2Seq~\citep{Sutskever2014} with raw sentences for machine translation, and DQN~\citep{Mnih-DQN-2015} with raw pixels and score to play games. 

\subsection{Reinforcement Learning}
\label{RL}

We provide background of reinforcement learning briefly in this section. After setting up the RL problem, we discuss value function, temporal difference learning, function approximation, policy optimization, deep RL, RL parlance, and close this section with a brief summary. To have a good understanding of deep reinforcement learning, it is essential to have a good understanding of reinforcement learning first.

\subsubsection{Problem Setup}

A RL agent interacts with an environment over time. At each time step $t$, the agent receives a state $s_t$ in a state space $\mathcal{S}$ and selects an action $a_t$ from an action space $\mathcal{A}$, following a policy $\pi(a_t|s_t)$, which is the agent's behavior, i.e., a mapping from state $s_t$ to actions $a_t$, receives a scalar reward $r_t$, and transitions to the next state $s_{t+1}$, according to the environment dynamics, or model, for reward function $\mathcal{R}(s,a)$ and state transition probability $\mathcal{P}(s_{t+1}|s_t, a_t)$ respectively. In an episodic problem, this process continues until the agent reaches a terminal state and then it restarts. The return $R_t = \sum_{k=0}^{\infty} \gamma^k r_{t+k}$ is the discounted, accumulated reward with the discount factor $\gamma \in (0,1]$. The agent aims to maximize the expectation of such long term return from each state.  The problem is set up in discrete state and action spaces. It is not hard to extend it to continuous spaces.   

\subsubsection{Exploration vs Exploitation}

multi-arm bandit

various exploration techniques

\subsubsection{Value Function}

A value function is a prediction of the expected, accumulative, discounted, future reward, measuring how good each state, or state-action pair, is. 
The state value $v_{\pi}(s) = E[R_t | s_t = s]$ is the expected return for following policy $\pi$ from state $s$. 
$v_{\pi}(s)$ decomposes into the Bellman equation: $v_{\pi}(s) = \sum_a \pi(a|s) \sum_{s',r}p(s',r|s,a)[r + \gamma v_{\pi}(s')]$.
An optimal state value $v_{*}(s) = \max_{\pi} v_{\pi}(s) = \max_{a}q_{\pi ^{*}}(s,a)$ is the maximum state value achievable by any policy for state $s$.
$v_{*}(s)$ decomposes into the Bellman equation: $v_{*}(s) = \max_a \sum_{s',r}p(s',r|s,a)[r + \gamma v_{*}(s')]$.
The action value $q_{\pi}(s, a) = E[R_t | s_t = s, a_t = a]$ is the expected return for selecting action $a$ in state $s$ and then following policy $\pi$. 
$q_{\pi}(s, a)$ decomposes into the Bellman equation: $q_{\pi}(s, a) = \sum_{s',r}p(s',r|s,a)[r + \gamma \sum_{a'} \pi(a'|s')q_{\pi}(s', a')]$.
An optimal action value function $q_{*}(s, a) = \max_{\pi} q_{\pi}(s, a)$ is the maximum action value achievable by any policy for state $s$ and action $a$. 
$q_{*}(s, a)$ decomposes into the Bellman equation: $q_{*}(s, a) = \sum_{s',r}p(s',r|s,a)[r + \gamma \max_{a'}q_{*}(s', a')]$.
We denote an optimal policy by $\pi^{*}$. 

\subsubsection{Dynamic Programming}

\subsubsection{Temporal Difference Learning}

When a RL problem satisfies the Markov property, i.e., the future depends only on the current state and action, but not on the past, it is formulated as a Markov Decision Process (MDP), defined by the 5-tuple $(\mathcal{S}, \mathcal{A}, \mathcal{P},\mathcal{R}, \gamma)$. When the system model is available, we use dynamic programming methods: policy evaluation to calculate value/action value function for a policy, value iteration and policy iteration for finding an optimal policy. When there is no model, we resort to RL methods. RL methods also work when the model is available. Additionally, a RL environment can be a multi-armed bandit, an MDP, a POMDP, a game, etc.

Temporal difference (TD) learning is central in RL. TD learning is usually refer to the learning methods for value function evaluation in~\citet{Sutton1988}. SARSA~\citep{Sutton2018} and Q-learning~\citep{Watkins1992} are also regarded as temporal difference learning. 

TD learning~\citep{Sutton1988} learns value function $V(s)$ directly from experience with TD error, with bootstrapping, in a model-free, online, and fully incremental way.  TD learning is a prediction problem. The update rule is $V(s) \leftarrow V(s) + \alpha [r + \gamma V(s') - V(s)]$, where $\alpha$ is a learning rate, and $r + \gamma V(s') - V(s)$ is called TD error. Algorithm~\ref{TD-algo} presents the pseudo code for tabular TD learning. Precisely, it is tabular TD(0) learning, where "0" indicates it is based on one-step return.

Bootstrapping, like the TD update rule, estimates state or action value based on subsequent estimates, is common in RL, like TD learning, Q learning, and actor-critic.
 Bootstrapping methods are usually faster to learn, and enable learning to be online and continual. Bootstrapping methods are not instances of true gradient decent, since the target depends on the weights to be estimated. The concept of semi-gradient descent is then introduced~\citep{Sutton2018}.  

\begin{algorithm}[h]
\SetAlgoNoLine
\textbf{Input: } the policy $\pi$ to be evaluated\\
\textbf{Output: } value function $V$\\
initialize $V$ arbitrarily, e.g., to 0 for all states\\
\For{each episode}{
	initialize state $s$\\
	\For{each step of episode, state $s$ is not terminal}{
	        $a \leftarrow $ action given by $\pi$ for $s$\\
	        take action $a$, observe $r$, $s'$\\
	        $V(s) \leftarrow V(s) + \alpha [r + \gamma V(s') - V(s)]$\\
	        $s \leftarrow s'$
	}
}
\caption{TD learning, adapted from \citet{Sutton2018}}
\label{TD-algo}
\end{algorithm}

\begin{algorithm}[h]
\SetAlgoNoLine
\textbf{Output: } action value function $Q$\\
initialize $Q$ arbitrarily, e.g., to 0 for all states, set action value for terminal states as 0\\
\For{each episode}{
	initialize state $s$\\
	\For{each step of episode, state $s$ is not terminal}{
	        $a \leftarrow $ action for $s$ derived by $Q$, e.g., $\epsilon$-greedy\\
	        take action $a$, observe $r$, $s'$\\
	        $a' \leftarrow $ action for $s'$ derived by $Q$, e.g., $\epsilon$-greedy\\
	        $Q(s, a) \leftarrow Q(s, a) + \alpha [r + \gamma Q(s', a') - Q(s,a)]$\\
	        $s \leftarrow s'$, $a \leftarrow a'$
	}
}
\caption{SARSA, adapted from \citet{Sutton2018}}
\label{SARSA-algo}
\end{algorithm}

\begin{algorithm}[H]
\SetAlgoNoLine
\textbf{Output: } action value function $Q$\\
initialize $Q$ arbitrarily, e.g., to 0 for all states, set action value for terminal states as 0\\
\For{each episode}{
	initialize state $s$\\
	\For{each step of episode, state $s$ is not terminal}{
	        $a \leftarrow $ action for $s$ derived by $Q$, e.g., $\epsilon$-greedy\\
	        take action $a$, observe $r$, $s'$\\
	        $Q(s, a) \leftarrow Q(s, a) + \alpha [r + \gamma  \max_{a'} Q(s', a') - Q(s, a)]$\\
	        $s \leftarrow s'$
	}
}
\caption{Q learning, adapted from \citet{Sutton2018}}
\label{Q-algo}
\end{algorithm}

SARSA, representing state, action, reward, (next) state, (next) action, is an on-policy control method to find the optimal policy, with the update rule, $Q(s, a) \leftarrow Q(s, a) + \alpha [r + \gamma Q(s', a') - Q(s,a)]$.  Algorithm~\ref{SARSA-algo} presents the pseudo code for tabular SARSA, precisely tabular SARSA(0).

Q-learning is an off-policy control method to find the optimal policy. Q-learning learns action value function, with the update rule, $Q(s, a) \leftarrow Q(s, a) + \alpha [r + \gamma  \max_{a'} Q(s', a') - Q(s, a)]$. Q learning refines the policy greedily  with respect to action values by the max operator. Algorithm~\ref{Q-algo} presents the pseudo code for Q learning, precisely, tabular Q(0) learning.

TD-learning, Q-learning and SARSA converge under certain conditions. From an optimal action value function, we can derive an optimal policy. 

\subsubsection{Multi-step Bootstrapping}

The above algorithms are referred to as TD(0) and Q(0), learning with one-step return. We have TD learning and Q learning variants and Monte-Carlo approach with multi-step return in the forward view. The eligibility trace from the backward view provides an online, incremental implementation, resulting in TD($\lambda$) and Q($\lambda$) algorithms, where $\lambda \in[0,1]$. TD(1) is the same as the Monte Carlo approach. 

Eligibility trace is a short-term memory, usually lasting within an episode, assists the learning process, by affecting the weight vector. The weight vector is a long-term memory, lasting the whole duration of the system, determines the estimated value. Eligibility trace helps with the issues of long-delayed rewards and non-Markov tasks~\citep{Sutton2018}.

TD($\lambda$) unifies one-step TD prediction, TD(0), with Monte Carlo methods, TD(1), using eligibility traces and the decay parameter $\lambda$, for prediction algorithms.
\citet{DeAsis2018} made unification for multi-step TD control algorithms. 

\subsubsection{Function Approximation}

We discuss the tabular cases above, where a value function or a policy is stored in a tabular form. Function approximation is a way for generalization when the state and/or action spaces are large or continuous. Function approximation aims to generalize from examples of a function to construct an approximate of the entire function; it is usually a concept in supervised learning, studied in the fields of machine learning, patten recognition, and statistical curve fitting; function approximation in reinforcement learning usually treats each backup as a training example, and encounters new issues like nonstationarity, bootstrapping, and delayed targets~\citep{Sutton2018}. Linear function approximation is a popular choice, partially due to its desirable theoretical properties, esp. before the work of Deep Q-Network~\citep{Mnih-DQN-2015}.  However, the integration of reinforcement learning and neural networks dated back a long time ago~\citep{Sutton2018,Bertsekas96, Schmidhuber2015-DL}. 

Algorithm~\ref{TD-FA-algo} presents the pseudo code for TD(0) with function approximation. $\hat{v}(s, \bm{w})$ is the approximate value function, $\bm{w}$ is the value function weight vector, $\nabla \hat{v}(s, \bm{w})$ is the gradient of the approximate value function with respect to the weight vector, and the weight vector is updated following the update rule, $\bm{w} \leftarrow \bm{w} + \alpha [r + \gamma  \hat{v}(s', \bm{w}) - \hat{v}(s, \bm{w})] \nabla \hat{v}(s, \bm{w}) $. 

\begin{algorithm}[h]
\SetAlgoNoLine
\textbf{Input: } the policy $\pi$ to be evaluated\\
\textbf{Input: } a differentiable value function $\hat{v}(s, \bm{w})$, $\hat{v}(terminal, \cdot) = 0$\\
\textbf{Output: } value function $\hat{v}(s, \bm{w})$\\
initialize value function weight $\bm{w}$ arbitrarily, e.g., $\bm{w} = 0$\\
\For{each episode}{
	initialize state $s$\\
	\For{each step of episode, state $s$ is not terminal}{
	        $a \leftarrow \pi(\cdot|s)$\\
	        take action $a$, observe $r$, $s'$\\
	        $\bm{w} \leftarrow \bm{w} + \alpha [r + \gamma  \hat{v}(s', \bm{w}) - \hat{v}(s, \bm{w})] \nabla \hat{v}(s, \bm{w}) $\\
	        $s \leftarrow s'$
	}
}
\caption{TD(0) with function approximation, adapted from \citet{Sutton2018}}
\label{TD-FA-algo}
\end{algorithm}


When combining off-policy, function approximation, and bootstrapping, instability and divergence may occur~\citep{Tsitsiklis97}, which is called the deadly triad issue~\citep{Sutton2018}. All these three elements are necessary: function approximation for scalability and generalization, bootstrapping for computational and data efficiency, and off-policy learning for freeing behaviour policy from target policy.  What is the root cause for the instability? Learning or sampling are not, since dynamic programming suffers from divergence with function approximation; exploration, greedification, or control are not, since prediction alone can diverge; local minima or complex non-linear function approximation are not, since linear function approximation can produce instability~\citep{Sutton2016Course}. It is unclear what is the root cause for instability -- each single factor mentioned above is not -- there are still many open problems in off-policy learning~\citep{Sutton2018}. 

Table~\ref{table-rl}  presents various algorithms that tackle various issues~\citep{Sutton2016Course}.  Deep RL algorithms like Deep Q-Network~\citep{Mnih-DQN-2015} and A3C~\citep{Mnih-A3C-2016} are not presented here, since they do not have theoretical guarantee, although they achieve stunning performance empirically.

Before explaining Table~\ref{table-rl}, we introduce some background definitions. 
Recall that Bellman equation for value function is $v_{\pi}(s) = \sum_a \pi(a|s) \sum_{s',r}p(s',r|s,a)[r + \gamma v_{\pi}(s')]$.
Bellman operator is defined as $(B_{\pi}v)(s) \doteq \sum_a \pi(a|s) \sum_{s',r}p(s',r|s,a)[r + \gamma v_{\pi}(s')]$.
TD fix point is then $v_{\pi} = B_{\pi}v_{\pi} $.
Bellman error for the function approximation case is then $\sum_a \pi(a|s) \sum_{s',r}p(s',r|s,a)[r + \gamma \hat{v}^{\pi}(s', \bm{w})] - \hat{v}^{\pi}(s, \bm{w})$, the right side of Bellman equation with function approximation minus the left side. 
It can be written as $B_{\pi}v_{\bm{w}} - v_{\bm{w}}$.
Bellman error is the expectation of the TD error.

ADP algorithms refer to dynamic programming algorithms like policy evaluation, policy iteration, and value iteration, with function approximation. 
Least square temporal difference (LSTD)~\citep{Bradtke96} computes TD fix-point directly in batch mode. LSTD is data efficient, yet with squared time complexity. LSPE~\citep{Nedic2003} extended LSTD. 
Fitted-Q algorithms~\citep{Ernst2005, Riedmiller2005} learn action values in batch mode.
Residual gradient algorithms~\citep{Baird1995} minimize Bellman error. 
Gradient-TD~\citep{Sutton2009GTD-ICML, Sutton2009GTD-NIPS, Mahmood2014} methods are true gradient algorithms, perform SGD in the projected Bellman error (PBE), converge robustly under off-policy training and non-linear function approximation. Emphatic-TD~\citep{Sutton2016} emphasizes some updates and de-emphasizes others by reweighting, improving computational efficiency, yet being a semi-gradient method. See~\citet{Sutton2018} for more details. \citet{Du2017PE} proposed variance reduction techniques for policy evaluation to achieve fast convergence. \citet{White2016} performed empirical comparisons of linear TD methods, and made suggestions about their practical use.

\begin{table}[h]
\centering
\begin{tabular}{cc|c|c|c|c|c|c|l}
\cline{3-8}
& & \multicolumn{6}{ c| }{algorithm} \\ \cline{3-8}
&  & \shortstack{TD($\lambda$)\\ SARSA($\lambda$)} &  \shortstack{ADP} & \shortstack{LSTD($\lambda$)\\ LSPE($\lambda$)} & Fitted-Q & \shortstack{Residual\\ Gradient} &  \shortstack{GTD($\lambda$)\\ GQ($\lambda$)}\\ \cline{1-8}
\multicolumn{1}{ |c  }{ } &
\multicolumn{1}{ |c| }{\shortstack{linear\\ computation}} & $\checkmark$  & $\checkmark$ &  &  & $\checkmark$  &  $\checkmark$&    \\ \cline{2-8}
\multicolumn{1}{ |c  }{\multirow{4}{*}{\begin{turn}{-270}issue\end{turn}}}                        & 
\multicolumn{1}{ |c| }{\shortstack{nonlinear\\ convergent}} &  &  &  & $\checkmark$ & $\checkmark$& $\checkmark$&     \\ \cline{2-8}
\multicolumn{1}{ |c  }{}                        & 
\multicolumn{1}{ |c| }{\shortstack{off-policy\\ convergent}} &  &  &  $\checkmark$&  & $\checkmark$& $\checkmark$&     \\ \cline{2-8}
\multicolumn{1}{ |c  }{}                        & 
\multicolumn{1}{ |c| }{\shortstack{model-free,\\ online}} & $\checkmark$ &  & $\checkmark$ &  &$\checkmark$ & $\checkmark$&     \\ \cline{2-8}
\multicolumn{1}{ |c  }{}                        & 
\multicolumn{1}{ |c| }{\shortstack{converges to\\ PBE = 0}} &  $\checkmark$&  $\checkmark$&  $\checkmark$&  $\checkmark$& &  $\checkmark$&     \\ \cline{1-8}
\end{tabular}
\caption{RL Issues vs. Algorithms}
\label{table-rl}
\end{table}


\subsubsection{Policy Optimization}

In contrast to value-based methods like TD learning and Q-learning, policy-based methods optimize the policy $\pi(a|s; \bm{\theta})$ (with function approximation) directly, and update the parameters $\bm{\theta}$ by gradient ascent on $E[R_t]$. REINFORCE~\citep{Williams1992} is a policy gradient method, updating $\bm{\theta}$ in the direction of $\nabla_{\bm{\theta}} \log \pi(a_t|s_t; \bm{\theta}) R_t$. Usually a baseline $b_t(s_t)$ is subtracted from the return to reduce the variance of gradient estimate, yet keeping its unbiasedness, to yield the gradient direction $\nabla_{\bm{\theta}} \log \pi(a_t|s_t; \bm{\theta}) (R_t - b_t(s_t))$. Using $V(s_t)$ as the baseline $b_t(s_t)$, we have the advantage function $A(a_t, s_t) = Q(a_t, s_t) - V(s_t)$, since $R_t$ is an estimate of $Q(a_t, s_t)$. Algorithm~\ref{REINFORCE} presents the pseudo code for REINFORCE algorithm in the episodic case.

\begin{algorithm}[h]
\SetAlgoNoLine
\textbf{Input: } policy $\pi(a|s, \bm{\theta})$, $\hat{v}(s, \bm{w})$ \\
\textbf{Parameters: } step sizes, $\alpha > 0$, $\beta > 0$\\
\textbf{Output: } policy $\pi(a|s, \bm{\theta})$\\
initialize policy parameter $\bm{\theta}$ and state-value weights $\bm{w}$\\
\For{true}{
	generate an episode $s_0$, $a_0$, $r_1$,  $\cdots$, $s_{T-1}$, $a_{T-1}$, $r_T$, following $\pi(\cdot|\cdot, \bm{\theta})$\\
	\For{each step $t$ of episode 0, $\cdots$, $T-1$}{
	        $G_t \leftarrow $ return from step $t$\\
	        $\delta \leftarrow G_t - \hat{v}(s_t, \bm{w})$\\
	        	$\bm{w} \leftarrow \bm{w} + \beta \delta \nabla_{\bm{w}} \hat{v}(s_t, \bm{w}) $\\
		$\bm{\theta} \leftarrow \bm{\theta} + \alpha \gamma^t \bm{\delta} \nabla_{\bm{\theta}}  log \pi(a_t|s_t, \bm{\theta})$
	}
}
\caption{REINFORCE with baseline (episodic), adapted from \citet{Sutton2018}}
\label{REINFORCE}
\end{algorithm}

In actor-critic algorithms, the critic updates action-value function parameters, and the actor updates policy parameters, in the direction suggested by the critic. Algorithm~\ref{Actor-Critic} presents the pseudo code for one-step actor-critic algorithm in the episodic case.

\begin{algorithm}[h]
\SetAlgoNoLine
\textbf{Input: } policy $\pi(a|s, \bm{\theta})$, $\hat{v}(s, \bm{w})$ \\
\textbf{Parameters: } step sizes, $\alpha > 0$, $\beta > 0$\\
\textbf{Output: } policy $\pi(a|s, \bm{\theta})$\\
initialize policy parameter $\bm{\theta}$ and state-value weights $\bm{w}$\\
\For{true}{
        initialize $s$, the first state of the episode\\
        $I \leftarrow 1$\\
	\For{$s$ is not terminal}{
	        $a \sim \pi(\cdot|s, \bm{\theta})$\\
	        take action $a$, observe $s'$, $r$\\
	        $\delta \leftarrow r + \gamma  \hat{v}(s', \bm{w}) - \hat{v}(s, \bm{w})$  (if $s'$ is terminal, $\hat{v}(s', \bm{w}) \doteq 0$)\\
	        	$\bm{w} \leftarrow \bm{w} + \beta \delta \nabla_{\bm{w}} \hat{v}(s_t, \bm{w}) $\\
		$\bm{\theta} \leftarrow \bm{\theta} + \alpha I \delta \nabla_{\bm{\theta}}  log \pi(a_t|s_t, \bm{\theta})$\\
		$I \leftarrow \gamma I$\\
		$s \leftarrow s'$
	}
}
\caption{Actor-Critic (episodic), adapted from \citet{Sutton2018}}
\label{Actor-Critic}
\end{algorithm}

Policy iteration alternates between policy evaluation and policy improvement, to generate a sequence of improving policies. In policy evaluation, the value function of the current policy is estimated from the outcomes of sampled trajectories. In policy improvement, the current value function is used to generate a better policy, e.g., by selecting actions greedily with respect to the value function.

\subsubsection{Deep Reinforcement Learning}

We obtain deep reinforcement learning (deep RL) methods when we use deep neural networks to approximate any of the following components of reinforcement learning: value function, $\hat{v}(s; \bm{\theta})$ or $\hat{q}(s,a; \bm{\theta})$, policy $\pi(a|s; \bm{\theta})$, and model (state transition function and reward function). Here, the parameters $\bm{\theta}$ are the weights in deep neural networks. When we use "shallow" models, like linear function, decision trees, tile coding and so on as the function approximator, we obtain "shallow" RL, and the parameters $\bm{\theta}$ are the weight parameters in these models. Note, a shallow model, e.g., decision trees, may be non-linear. The distinct difference between deep RL and "shallow" RL is what function approximator is used. This is similar to the difference between deep learning and "shallow" machine learning. We usually utilize stochastic gradient descent to update weight parameters in deep RL. When off-policy, function approximation, in particular, non-linear function approximation, and bootstrapping are combined together, instability and divergence may occur~\citep{Tsitsiklis97}. However, recent work like Deep Q-Network~\citep{Mnih-DQN-2015} and AlphaGo~\citep{Silver-AlphaGo-2016} stabilized the learning and achieved outstanding results.

\subsubsection{RL Parlance}

We explain some terms in RL parlance. 

The prediction problem, or policy evaluation, is to compute the state or action value function for a policy. The control problem is to find the optimal policy. Planning constructs a value function or a policy with a model. 

On-policy methods evaluate or improve the behavioural policy, e.g., SARSA fits the action-value function to the current policy, i.e., SARSA evaluates the policy based on samples from the same policy, then refines the policy greedily with respect to action values. In off-policy methods, an agent learns an optimal value function/policy, maybe following an unrelated behavioural policy, e.g., Q-learning attempts to find action values for the optimal policy directly, not necessarily fitting to the policy generating the data, i.e., the policy Q-learning obtains is usually different from the policy that generates the samples.  The notion of on-policy and off-policy can be understood as same-policy and different-policy.

The exploration-exploitation dilemma is about the agent needs to exploit the currently best action to maximize rewards greedily, yet it has to explore the environment to find better actions, when the policy is not optimal yet, or the system is non-stationary.  

In model-free methods, the agent learns with trail-and-error from experience explicitly; the model (state transition function) is not known or learned from experience. RL methods that use models are model-based methods. 

In online mode, training algorithms are executed on data acquired in sequence. In offline mode, or batch mode, models are trained on the entire data set. 

With bootstrapping, an estimate of state or action value is updated from subsequent estimates.

\subsubsection{Brief Summary}

A RL problem is formulated as an MDP when the observation about the environment satisfies the Markov property. An MDP is defined by the 5-tuple $(\mathcal{S}, \mathcal{A}, \mathcal{P},\mathcal{R}, \gamma)$. A central concept in RL is value function. Bellman equations are cornerstone for developing RL algorithms. Temporal difference learning algorithms are fundamental for evaluating/predicting value functions.  Control algorithms find optimal policies. Reinforcement learning algorithms may be based on value function and/or policy, model-free or model-based, on-policy or off-policy, with function approximation or not, with sample backups (TD and Monte Carlo) or full backups (dynamic programming and exhaustive search), and about the depth of backups, either one-step return (TD(0) and dynamic programming) or multi-step return (TD($\lambda$), Monte Carlo, and exhaustive search). When combining off-policy, function approximation, and bootstrapping, we face instability and divergence~\citep{Tsitsiklis97}, the deadly triad issue~\citep{Sutton2018}. Theoretical guarantee has been established for linear function approximation, e.g., Gradient-TD~\citep{Sutton2009GTD-ICML, Sutton2009GTD-NIPS, Mahmood2014}, Emphatic-TD~\citep{Sutton2016} and \citet{Du2017PE}. With non-linear function approximation, in particular deep learning,  algorithms like Deep Q-Network~\citep{Mnih-DQN-2015} and AlphaGo~\citep{Silver-AlphaGo-2016, Silver-AlphaGo-2017} stabilized the learning and achieved stunning results, which is the focus of this overview.




\section{Core Elements}

A RL agent executes a sequence of actions and observe states and rewards,  with major components of value function, policy and model. A RL problem may be formulated as a prediction, control or planning problem, and solution methods may be model-free or model-based, with value function and/or policy. Exploration-exploitation is a fundamental tradeoff in RL. Knowledge would be critical for RL.
In this section, we discuss core RL elements: value function in Section~\ref{value}, policy in Section~\ref{policy}, reward in Section~\ref{reward}, model and planning in Section~\ref{modelplanning},  exploration in Section~\ref{exploration}, and knowledge in Section~\ref{knowledge}.


\subsection{Value Function}
\label{value}

Value function is a fundamental concept in reinforcement learning, and temporal difference (TD) learning~\citep{Sutton1988} and its extension, Q-learning~\citep{Watkins1992}, are classical algorithms for learning state and action value functions respectively. In the following, we focus on Deep Q-Network~\citep{Mnih-DQN-2015}, a recent breakthrough, and its extensions.

\subsubsection{Deep Q-Network (DQN) And Extensions}
\label{DQN}

\citet{Mnih-DQN-2015} introduced Deep Q-Network (DQN)  and ignited the field of deep RL.  
We present DQN pseudo code in Algorithm~\ref{DQN-algo}. 

\begin{algorithm}[h]
\SetAlgoNoLine
\textbf{Input: } the pixels and the game score\\
\textbf{Output: } Q action value function (from which we obtain policy and select action)\\
Initialize replay memory $D$\\
Initialize action-value function $Q$ with random weight $\bm{\theta}$\\
Initialize target action-value function $\hat{Q}$ with weights $\bm{\theta}^{-} = \bm{\theta}$\\    
\For{episode = 1 to $M$}{
	Initialize sequence $s_1 = \{x_1\}$ and preprocessed sequence $\phi_1 = \phi(s_1)$\\
	\For{t = 1 to $T$}{
		Following $\epsilon$-greedy policy, select $a_t =
  		\begin{cases}
    		\text{a random action}      & \text{with probability } \epsilon\\
    		\argmax_{a}Q(\phi(s_t), a; \bm{\theta})   & \text{otherwise}\\
  		\end{cases} $\\
		Execute action $a_i$ in emulator and observe reward $r_t$ and image $x_{t+1}$\\
		Set $s_{t+1} = s_t, a_t, x_{t+1}$ and preprocess $\phi_{t+1} = \phi(s_{t+1})$\\
		Store transition $(\phi_t, a_t, r_t, \phi_{t+1})$ in $D$\\
		\tcp{experience replay}
		Sample random minibatch of transitions $(\phi_j, a_j, r_j, \phi_{j+1})$ from $D$\\
		Set $y_j =
  		\begin{cases}
    		r_j       & \text{if episode terminates at step } j+1\\
    		r_j + \gamma \max_{a'} \hat{Q}(\phi_{j+1}, a'; \bm{\theta}^-)   & \text{otherwise}\\
  		\end{cases} $\\
		Perform a gradient descent step on $(y_j - Q(\phi_j, a_j; \bm{\theta}))^2$ w.r.t. the network parameter $\bm{\theta}$\\
		\tcp{periodic update of target network}
		Every $C$ steps reset $\hat{Q} = Q$, i.e., set $\bm{\theta}^- = \bm{\theta}$
	}
}
\caption{Deep Q-Nework (DQN), adapted from \citet{Mnih-DQN-2015}}
\label{DQN-algo}
\end{algorithm}

Before DQN, it is well known that RL is unstable or even divergent when action value function is approximated with a nonlinear function like neural networks.  DQN made several important contributions: 1)  stabilize the training of action value function approximation with deep neural networks (CNN) using experience replay~\citep{Lin1992} and target network; 2) designing an end-to-end RL approach, with only the pixels and the game score as inputs, so that only minimal domain knowledge is required; 3) training a flexible network with the same algorithm, network architecture and hyperparameters to perform well on many different tasks, i.e., 49 Atari games~\citep{Bellemare2013}, and outperforming previous algorithms and performing comparably to a human professional tester. 

See Chapter 16 in \citet{Sutton2018} for a detailed and intuitive description of Deep Q-Network.   See Deepmind's description of DQN at https://deepmind.com/research/dqn/. 


\subsubsection*{Double DQN}
\label{D-DQN}

\citet{vanHasselt2016} proposed Double DQN (D-DQN) to tackle the over-estimate problem in Q-learning. In standard Q-learning, as well as in DQN, the parameters are updated as follows:  
\[\bm{\theta}_{t+1} = \bm{\theta}_t + \alpha (y_t ^Q- Q(s_t, a_t; \bm{\theta}_t) ) \nabla_{\bm{\theta}_t}  Q(s_t, a_t; \bm{\theta}_t),\] 
where 
\[y_t^Q = r_{t+1} + \gamma \max_{a}Q(s_{t+1}, a; \bm{\theta}_t),\] 
so that the max operator uses the same values to both select and evaluate an action. As a consequence, it is more likely to select over-estimated values, and results in over-optimistic value estimates. \citet{vanHasselt2016}  proposed to evaluate the greedy policy according to the online network, but to use the target network to estimate its value. This can be achieved with a minor change to the DQN algorithm, replacing $y_t^Q$ with 
\[y_t^{D-DQN} = r_{t+1} + \gamma Q(s_{t+1}, \argmax_{a} Q(s_{t+1}, a_t; \bm{\theta}_t); \bm{\theta}_t^-),\]
where $\bm{\theta}_t$ is the parameter for online network and $\bm{\theta}_t^-$ is the parameter for target network. For reference, $y_t^Q$ can be written as 
\[y_t^{Q} = r_{t+1} + \gamma Q(s_{t+1}, \argmax_{a} Q(s_{t+1}, a_t; \bm{\theta}_t); \bm{\theta}_t).\]

D-DQN found better policies than DQN on Atari games. 

\subsubsection*{Prioritized Experience Replay} 

In DQN, experience transitions are uniformly sampled from the replay memory, regardless of the significance of experiences. \citet{Schaul2016} proposed to prioritize experience replay, so that important experience transitions can be replayed more frequently, to learn more efficiently. The importance of experience transitions are measured by TD errors. The authors designed a stochastic prioritization based on the TD errors, using importance sampling to avoid the bias in the update distribution. The authors used prioritized experience replay in DQN and D-DQN, and improved their performance on Atari games. 

\subsubsection*{Dueling Architecture} 
\label{dueling}

\citet{Wang-Dueling-2016} proposed the dueling network architecture to estimate state value function $V(s)$ and associated advantage function $A(s,a)$, and then combine them to estimate action value function $Q(s,a)$, to converge faster than Q-learning. In DQN, a CNN layer is followed by a fully connected (FC) layer. In dueling architecture, a CNN layer is followed by two streams of FC layers, to estimate value function and advantage function separately; then the two streams are combined to estimate action value function. Usually we use the following to combine $V(s)$ and $A(s,a)$ to obtain $Q(s,a)$,
\[Q(s,a; \bm{\theta}, \alpha, \beta) = V(s; \bm{\theta}, \beta) + \big( A(s,a; \bm{\theta}, \alpha) - \max_{a'}A(s,a'; \bm{\theta}, \alpha)\big) \]
where $\alpha$ and $\beta$ are parameters of the two streams of FC layers. \citet{Wang-Dueling-2016} proposed to replace max operator with average as the following for better stability,
\[Q(s,a; \bm{\theta}, \alpha, \beta) = V(s; \bm{\theta}, \beta) + \big( A(s,a; \bm{\theta}, \alpha) - \frac{a}{|\mathcal{A}|}A(s,a'; \bm{\theta}, \alpha)\big) \]
Dueling architecture implemented with D-DQN and prioritized experience replay improved previous work, DQN and D-DQN with prioritized experience replay, on Atari games.

\subsubsection*{Distributional Value Function} 

\citet{Bellemare2017Distributional}

\subsubsection*{Rainbow} 
\label{rainbow}

\citet{Hessel2018}


\subsubsection*{More DQN Extensions}

DQN has been receiving much attention. We list several extensions/improvements here.

\begin{itemize}
\item \citet{Anschel2017} proposed to reduce variability and instability by an average of previous Q-values estimates. 
\item \citet{He2017} proposed  to accelerate DQN by optimality tightening, a constrained optimization approach,  to propagate reward faster, and to improve accuracy over DQN.  
\item \citet{Liang2016} attempted to understand the success of DQN and reproduced results with shallow RL. 
\item \citet{ODonoghue2017} proposed policy gradient and Q-learning (PGQ), as discussed in Section~\ref{PGQ}. 
\item \citet{Oh2015} proposed spatio-temporal video prediction conditioned on actions and previous video frames with deep neural networks in Atari games. 
\item \citet{Osband2016} designed better exploration strategy to improve DQN. 
\item \citet{Hester2018} proposed to learn from demonstration with new loss functions, as discussed in Section~\ref{unsupervised}.
\end{itemize}

\subsection{Policy}
\label{policy}


A policy maps state to action, and policy optimization is to find an optimal mapping. 
As in~\citet{Peters2015}, the spectrum from direct policy search to value-based RL includes:
evolutionary strategies, CMA-ES (covariance matrix adaptation evolution strategy), 
episodic REPS (relative entropy policy search),
policy gradients, 
PILCO (probabilistic inference for learning control)~\citep{Deisenroth2011}, 
model-based REPS, policy search by trajectory optimization,
actor critic, natural actor critic,
eNAC (episodic natural actor critic),
advantage weighted regression,
conservative policy iteration,
LSPI (least square policy iteration)~\citep{Lagoudakis03},
Q-learning, and fitted Q, 
as well as important extensions,
contextual policy search, 
and hierarchical policy search.

We discuss actor-critic~\citep{Mnih-A3C-2016}. Then we discuss policy gradient, including deterministic policy gradient~\citep{Silver-DPG-2014, Lillicrap2016}, trust region policy optimization~\citep{Schulman2015}, and, benchmark results~\citep{Duan2016}.   Next we discuss the combination of policy gradient and off-policy RL~\citep{ODonoghue2017, Nachum2017Gap, Gu2017QProp}. 

See Retrace algorithm~\citep{Remi2016}, a safe and efficient return-based off-policy control algorithm,  and its actor-critic extension, Reactor~\citep{Gruslys2017}, for Retrace-actor. See distributed proximal policy optimization~\citep{Heess2017DPPO}.
\citet{McAllister2017} extended PILCO to POMDPs.


\subsubsection{Actor-Critic}
\label{A3C}

An actor-critic algorithm learns both a policy and a state-value function, and the value function is used for bootstrapping, i.e.,  updating a state from subsequent estimates, to reduce variance and accelerate learning~\citep{Sutton2018}. In the following, we focus on asynchronous advantage actor-critic (A3C)~\citep{Mnih-A3C-2016}. 
\citet{Mnih-A3C-2016} also discussed asynchronous one-step SARSA, one-step Q-learning and n-step Q-learning. 

In A3C, parallel actors employ different exploration policies to stabilize training, so that experience replay is not utilized. Different from most deep learning algorithms, asynchronous methods can run on a single multi-core CPU. For Atari games, A3C ran much faster yet performed better than or comparably with DQN, Gorila~\citep{Nair2015}, D-DQN, Dueling D-DQN, and Prioritized D-DQN. A3C also succeeded on continuous motor control problems: TORCS car racing games and MujoCo physics manipulation and locomotion, and Labyrinth, a navigating task in random 3D mazes using visual inputs, in which an agent will face a new maze in each new episode, so that it needs to learn a general strategy to explore random mazes.

 \begin{algorithm}[h]
\SetAlgoNoLine
Global shared parameter vectors $\bm{\theta}$ and $\bm{\theta}_v$, thread-specific parameter vectors $\bm{\theta}'$ and $\bm{\theta}'_v$\\
Global shared counter $T=0$, $T_{max}$\\
Initialize step counter $t \leftarrow 1$\\
\For{$T \leq T_{max}$}{
	Reset gradients, $d \bm{\theta} \leftarrow 0$ and $d \bm{\theta}_v \leftarrow 0$\\
	Synchronize thread-specific parameters $\bm{\theta}' = \bm{\theta}$ and $\bm{\theta}'_v = \bm{\theta}_v$\\
	Set $t_{start} = t$, get state $s_t$\\
	\For{$s_t$ $\text{ not terminal and }$ $t-t_{start} \leq t_{max}$}{
		Take $a_t$ according to policy $\pi(a_t|s_t; \bm{\theta}')$\\
		Receive reward $r_t$ and new state $s_{t+1}$\\
		$t \leftarrow t + 1$, $T \leftarrow T + 1$
		}
	$R =
  	\begin{cases}
    	0      & \text{for terminal } s_t\\
    	V(s_t, \bm{\theta}'_v)   & \text{otherwise}\\
  	\end{cases} $\\
	\For{$i \in \{t-1, ..., t_{start}\}$}{
		$R \leftarrow r_i + \gamma R$\\
		accumulate gradients wrt $\bm{\theta}'$: $d\bm{\theta} \leftarrow d\bm{\theta} + \nabla_{\bm{\theta}'} \log  \pi(a_i|s_i; \bm{\theta}') (R - V(s_i; \bm{\theta}'_v))$\\
		accumulate gradients wrt $\bm{\theta}'_v$: $d\bm{\theta}_v \leftarrow d\bm{\theta}_v + \nabla_{\bm{\theta}'_v} (R - V(s_i; \bm{\theta}'_v))^2$
		}
	Update asynchronously $\bm{\theta}$ using $d\bm{\theta}$, and $\bm{\theta}_v$ using $d\bm{\theta}_v$
}
\caption{A3C, each actor-learner thread, based on \citet{Mnih-A3C-2016}}
\label{A3C-algo}
\end{algorithm}

We present pseudo code for asynchronous advantage actor-critic for each actor-learner thread in Algorithm~\ref{A3C-algo}. A3C maintains a policy $\pi(a_t|s_t;\bm{\theta})$ and an estimate of the value function $V(s_t; \bm{\theta}_v)$, being updated with $n$-step returns in the forward view, after every $t_{max}$ actions or reaching a terminal state, similar to using minibatches. The gradient update can be seen as $\nabla_{\bm{\theta}'} \log \pi(a_t|s_t; \bm{\theta}')A(s_t, a_t; \bm{\theta}, \bm{\theta}_v)$, where $A(s_t, a_t; \bm{\theta}, \bm{\theta}_v) = \sum_{i=0}^{k-1} \gamma^ir_{t+i} + \gamma^k V(s_{t+k}; \bm{\theta}_v) - V(s_t; \bm{\theta}_v)$ is an estimate of the advantage function, with $k$ upbounded by $t_{max}$.

\citet{Wang2017} proposed a stable and sample efficient actor-critic deep RL model using experience replay, with truncated importance sampling, stochastic dueling network~\citep{Wang-Dueling-2016} as discussed in Section~\ref{dueling}, and trust region policy optimization~\citep{Schulman2015} as discussed in Section~\ref{TRPO}. \citet{Babaeizadeh2017} proposed a hybrid CPU/GPU implementation of A3C.


\subsubsection{Policy Gradient}

REINFORCE~\citep{Williams1992, Sutton2000} is a popular policy gradient method. 
Relatively speaking, Q-learning as discussed in Section~\ref{value} is sample efficient, while policy gradient is stable. 

\subsubsection*{Deterministic Policy Gradient}

Policies are usually stochastic. However, \citet{Silver-DPG-2014} and \citet{Lillicrap2016} proposed deterministic policy gradient (DPG) for efficient estimation of policy gradients.


\citet{Silver-DPG-2014} introduced the deterministic policy gradient (DPG) algorithm for RL problems with continuous action spaces. The deterministic policy gradient is the expected gradient of the action-value function, which integrates over the state space; whereas in the stochastic case, the policy gradient integrates over both state and action spaces. Consequently, the deterministic policy gradient can be estimated more efficiently than the stochastic policy gradient. The authors introduced an off-policy actor-critic algorithm to learn a deterministic target policy from an exploratory behaviour policy, and to ensure unbiased policy gradient with the compatible function approximation for deterministic policy gradients. Empirical results showed its superior to stochastic policy gradients, in particular in high dimensional tasks, on several problems: a high-dimensional bandit; standard benchmark RL tasks of mountain car and pendulum and 2D puddle world with low dimensional action spaces; and controlling an octopus arm with a high-dimensional action space. The experiments were conducted with tile-coding and linear function approximators.

\citet{Lillicrap2016} proposed an actor-critic, model-free, deep deterministic policy gradient (DDPG) algorithm in continuous action spaces, by extending DQN~\citep{Mnih-DQN-2015} and DPG~\citep{Silver-DPG-2014}. With actor-critic as in DPG, DDPG avoids the optimization of action at every time step to obtain a greedy policy as in Q-learning, which will make it infeasible in complex action spaces with large, unconstrained function approximators like deep neural networks.  To make the learning stable and robust, similar to DQN, DDPQ  deploys experience replay and an idea similar to target network, "soft" target, which, rather than copying the weights directly as in DQN, updates the soft target network weights $\bm{\theta}'$ slowly to track the learned networks weights $\bm{\theta}$: $\bm{\theta}' \leftarrow \tau \bm{\theta} + (1-\tau) \bm{\theta}'$, with $\tau \ll 1$. The authors adapted batch normalization to handle the issue that the different components of the observation with different physical units.
As an off-policy algorithm, DDPG learns an actor policy from experiences from an exploration policy by adding noise sampled from a noise process to the actor policy. More than 20 simulated physics tasks of varying difficulty in the MuJoCo environment were solved with the same learning algorithm, network architecture and hyper-parameters, and obtained policies with performance competitive with those found by a planning algorithm with full access to the underlying physical model and its derivatives. DDPG can solve problems with 20 times fewer steps of experience than DQN, although it still needs a large number of training episodes to find solutions, as in most model-free RL methods. It is end-to-end, with raw pixels as input. DDPQ paper also contains links to videos for illustration.

\citet{Hausknecht2016} considers parameterization of action space.


\subsubsection*{Trust Region Policy Optimization}
\label{TRPO}

\citet{Schulman2015} introduced an iterative procedure to monotonically improve policies theoretically, guaranteed by optimizing a surrogate objective function.  
The authors then proposed a practical algorithm, Trust Region Policy Optimization (TRPO), by making several approximations, including, introducing a trust region constraint, defined by the KL divergence between the new policy and the old policy, so that at every point in the state space, the KL divergence is bounded;  approximating the trust region constraint by the average KL divergence constraint; replacing the expectations and Q value in the optimization problem by sample estimates, with two variants: in the single path approach, individual trajectories are sampled; in the vine approach, a rollout set is constructed and multiple actions are performed from each state in the rollout set; and, solving the constrained optimization problem approximately to update the policy's parameter vector.
The authors also unified policy iteration and policy gradient with analysis, and showed that policy iteration, policy gradient, and natural policy gradient~\citep{Kakade2002} are special cases of TRPO. In the experiments, TRPO methods performed well  on simulated robotic tasks of swimming, hopping, and walking, as well as playing Atari games in an end-to-end manner directly from raw images.

\citet{Wu2017TRPO} proposed scalable TRPO with Kronecker-factored approximation to the curvature.

https://blog.openai.com/openai-baselines-ppo/



\subsubsection*{Benchmark Results}

\citet{Duan2016} presented a benchmark for continuous control tasks, including classic tasks like cart-pole, tasks with very large state and action spaces such as 3D humanoid locomotion and tasks with partial observations, and tasks with hierarchical structure, implemented various algorithms, including batch algorithms: REINFORCE, Truncated Natural Policy Gradient (TNPG), Reward-Weighted Regression (RWR), Relative Entropy Policy Search (REPS), Trust Region Policy Optimization (TRPO), Cross Entropy Method (CEM), Covariance Matrix Adaption Evolution Strategy (CMA-ES); online algorithms: Deep Deterministic Policy Gradient (DDPG); and recurrent variants of batch algorithms. The open source is available at:  https://github.com/rllab/rllab.

\citet{Duan2016}  compared various algorithms, and showed that DDPG, TRPO, and Truncated Natural Policy Gradient (TNPG)~\citep{Schulman2015} are effective in training deep neural network policies, yet better algorithms are called for hierarchical tasks.

\citet{Islam2017}

\citet{Tassa2018}

\subsubsection{Combining Policy Gradient with Off-Policy RL}
\label{PGQ}

\citet{ODonoghue2017} proposed to combine policy gradient with off-policy Q-learning (PGQ), to benefit from experience replay. Usually actor-critic methods are on-policy. The authors also showed that action value fitting techniques and actor-critic methods are equivalent, and interpreted regularized policy gradient techniques as advantage function learning algorithms. Empirically, the authors showed that PGQ outperformed DQN and A3C on Atari games.

\citet{Nachum2017Gap} introduced the notion of softmax temporal consistency, to generalize  the hard-max Bellman consistency as in off-policy Q-learning, and in contrast to the average consistency as in on-policy SARSA and actor-critic. The authors established the correspondence and a mutual compatibility property between softmax consistent action values and the optimal policy maximizing entropy regularized expected discounted reward. The authors proposed Path Consistency Learning, attempting to bridge the gap between value and policy based RL, by exploiting multi-step path-wise consistency on traces from both on and off policies.

\citet{Gu2017QProp} proposed Q-Prop to take advantage of the stability of policy gradients and the sample efficiency of off-policy RL. \citet{Schulman2017} showed the equivalence between entropy-regularized Q-learning and policy gradient.

\citet{Gu2017Interpolated}

\subsection{Reward}
\label{reward}

Rewards provide evaluative feedbacks for a RL agent to make decisions. Rewards may be sparse so that it is challenging for learning algorithms, e.g., in computer Go, a reward occurs at the end of a game.  There are unsupervised ways to harness environmental signals, see Section~\ref{unsupervised}. Reward function is a mathematical formulation for rewards. Reward shaping is to modify reward function to facilitate learning while maintaining optimal policy. Reward functions may not be available for some RL problems, which is the focus of this section.

In imitation learning,  an agent learns to perform a task from expert demonstrations, with samples of trajectories from the expert, without reinforcement signal, without additional data from the expert while training; two main approaches for imitation learning are behavioral cloning and inverse reinforcement learning. Behavioral cloning, or apprenticeship learning, or learning from demonstration, is formulated as a supervised learning problem to map state-action pairs from expert trajectories to policy, without learning the reward function~\citep{Ho2016Imitation, Ho2016}. Inverse reinforcement learning (IRL) is the problem of determining a reward function given observations of optimal behaviour~\citep{Ng2000}. \citet{Abbeel2004} approached apprenticeship learning via IRL. 

In the following, we discuss learning from demonstration~\citep{Hester2018}, 
and imitation learning with generative adversarial networks (GANs)~\citep{Ho2016, Stadie2017}. 
We will discuss GANs, a recent unsupervised learning framework, in Section~\ref{GANs}. 

\citet{Su2016} proposed to train dialogue policy jointly with reward model. 
\citet{Christiano2017} proposed to learn reward function by human preferences from comparisons of trajectory segments.
See also \citet{Hadfield-Menell2016, Merel2017,  Wang2017Imitation, vanSeijen2017}.

\citet{Amin2017}

\subsubsection*{Learning from Demonstration}
\label{demonstration}

\citet{Hester2018} proposed Deep Q-learning from Demonstrations (DQfD) to attempt to accelerate learning by leveraging demonstration data, using a combination of temporal difference (TD), supervised, and regularized losses. In DQfQ, reward signal is not available for demonstration data; however, it is available in Q-learning. The supervised large margin classification loss enables the policy derived from the learned value function to imitate the demonstrator; the TD loss enables the validity of value function according to the Bellman equation and its further use for learning with RL; the regularization loss function on network weights and biases prevents overfitting on small demonstration dataset. In the pre-training phase, DQfD trains only on demonstration data, to obtain a policy imitating the demonstrator and a value function for continual RL learning. After that, DQfD self-generates samples, and mixes them with demonstration data according to certain proportion to obtain training data. The authors showed that, on Atari games, DQfD in general has better initial performance, more average rewards, and learns faster than DQN.

In AlphaGo~\citep{Silver-AlphaGo-2016}, to be discussed in Section~\ref{AlphaGo}, the supervised learning policy network is learned from expert moves as learning from demonstration; the results initialize the RL policy network. See also \citet{Kim2014, Perez2017}. See~\citet{Argall2009} for a survey of robot learning from demonstration.

\citet{Vecerik2017}


\subsubsection*{Generative Adversarial Imitation Learning}
\label{ImitationLearing}

With IRL, an agent learns a reward function first, then from which derives an optimal policy.  Many IRL algorithms have high time complexity, with a RL problem in the inner loop.

 \citet{Ho2016} proposed generative adversarial imitation learning algorithm to learn policies directly from data, bypassing the intermediate IRL step. Generative adversarial training was deployed to fit the discriminator, the distribution of states and actions that defines expert behavior, and the generator, the policy.

Generative adversarial imitation learning finds a policy $\pi_{\bm{\theta}}$ so that a discriminator $\mathcal{D}_R$ can not distinguish states following the expert policy $\pi_E$ and states following the imitator policy $\pi_{\bm{\theta}}$, hence forcing $\mathcal{D}_R$ to take 0.5 in all cases and $\pi_{\bm{\theta}}$ not distinguishable from $\pi_E$ in the equillibrium. Such a game is formulated as:

\[\max_{\pi_{\bm{\theta}}} \min_{\mathcal{D}_R} -E_{\pi_{\bm{\theta}}} [\log \mathcal{D}_R(s)] -E_{\pi_E} [\log (1-\mathcal{D}_R(s))]\]  

The authors represented both $\pi_{\bm{\theta}}$ and $\mathcal{D}_R$ as deep neural networks, and found an optimal solution by repeatedly performing gradient updates on each of them.  $\mathcal{D}_R$ can be trained with supervised learning with a data set formed from traces from a current  $\pi_{\bm{\theta}}$ and expert traces. For a fixed $\mathcal{D}_R$, an optimal $\pi_{\bm{\theta}}$ is sought. Hence it is a policy optimization problem, with $-\log \mathcal{D}_R(s)$ as the reward. The authors trained $\pi_{\bm{\theta}}$ by trust region policy optimization~\citep{Schulman2015}. 

\citet{Li2017InfoGAIL}

\subsubsection*{Third Person Imitation Learning}
\label{ThirdPerson}

\citet{Stadie2017} argued that previous works in imitation learning, like \citet{Ho2016} and \citet{Finn2016}, have the limitation of first person demonstrations, and proposed to learn from unsupervised third person demonstration, mimicking human learning by observing other humans achieving goals.

\subsection{Model and Planning}
\label{modelplanning}

A model is an agent's representation of the environment, including the transition model and the reward model. Usually we assume the reward model is known. We discuss how to handle unknown reward models in Section~\ref{reward}.
Model-free RL approaches handle unknown dynamical systems, however, they usually require large number of samples, which may be costly or prohibitive to obtain for real physical systems. Model-based RL approaches learn  value function and/or policy in a data-efficient way, however, they may suffer from the issue of model identification so that the estimated models may not be accurate, and the performance is limited by the estimated model.
Planning constructs a value function or a policy usually with a model, so that planning is usually related to model-based RL methods.

\citet{Chebotar2017} attempted to combine the advantages of both model-free and model-based RL approaches. The authors focused on time-varying linear-Gaussian policies, and integrated a model-based linear quadratic regulator (LQR) algorithm with a model-free path integral policy improvement  algorithm. To generalize the method for arbitrary parameterized policies such as deep neural networks, the authors combined the proposed approach with guided policy search (GPS)~\citep{Levine2016}. The proposed approach does not generate synthetic samples with estimated models to avoid degradation from modelling errors. See recent work on model-based learning, e.g., \citet{Gu2016, Henaff2017, Hester2017texplore, Oh2017VPN, Watter2015}. 

\citet{Tamar2016} introduced Value Iteration Networks (VIN), a fully differentiable CNN planning module to approximate the value iteration algorithm, to learn to plan, e.g, policies in RL. In contrast to conventional planning, VIN is model-free, where reward and transition probability are part of the neural network to be learned, so that it may avoid issues with system identification. VIN can be trained end-to-end with backpropagation. VIN can generalize in a diverse set of tasks: simple gridworlds, Mars Rover Navigation, continuous control and WebNav Challenge for Wikipedia links navigation~\citep{Nogueira2016}. One merit of Value Iteration Network, as well as Dueling Network\citep{Wang-Dueling-2016},  is that they design novel deep neural networks architectures for reinforcement learning problems. See a blog about VIN at https://github.com/karpathy/paper-notes/blob/master/vin.md.

\citet{Silver-Predictron-2016} proposed the predictron to integrate learning and planning into one end-to-end training procedure with raw input in Markov reward process, which can be regarded as Markov decision process without actions. See classical Dyna-Q~\citep{Sutton1990}.

\citet{Weber2017}

\citet{Andrychowicz2017}

\subsection{Exploration}
\label{exploration}

A RL agent usually uses  exploration to reduce its uncertainty about the reward function and transition probabilities of the environment. In tabular cases, this uncertainty can be quantified as confidence intervals or posterior of environment parameters, which are related to the state-action visit counts. With count-based exploration, a RL agent uses visit counts to guide its behaviour to reduce uncertainty. However, count-based methods are not directly useful in large domains. Intrinsic motivation suggests to explore what is surprising, typically in learning process based on change in prediction error. Intrinsic motivation methods do not require Markov property and tabular representation as count-based methods require. \citet{Bellemare2016} proposed pseudo-count, a density model over the state space, to unify count-based exploration and intrinsic motivation, by introducing information gain, to relate to confidence intervals in count-based exploration, and to relate to learning progress in intrinsic motivation. The author established pseudo-count's theoretical advantage over previous intrinsic motivation methods, and validated it with Atari games. 

\citet{Nachum2017} proposed an under-appreciated reward exploration technique to avoid the previous ineffective, undirected exploration strategies of the reward landscape, as in $\epsilon$-greedy and entropy regularization, and to promote directed exploration of the regions, in which the log-probability of an action sequence under the current policy under-estimates the resulting reward. The under-appreciated reward exploration strategy resulted from importance sampling from the optimal policy, and combined a mode seeking and a mean seeking terms to tradeoff exploration and exploitation.  The authors implemented the proposed exploration strategy with minor modifications to REINFORCE, and validated it, for the first time with a RL method,  on several algorithmic tasks. 

\citet{Osband2016} proposed bootstrapped DQN to combine deep exploration with deep neural networks to achieve efficient learning. \citet{Houthooft2016} proposed variational information maximizing exploration for continuous state and action spaces.
\citet{Fortunato2017} proposed NoisyNet for efficient exploration by adding parametric noise added to weights of deep neural networks.
See also \citet{Azar2017, Jiang2016, Ostrovski2017}.

\citet{Tang2017}

\citet{Fu2017}

\subsection{Knowledge}
\label{knowledge}

(This section would be an open-ended discussion.)

Knowledge would be critical for further development of RL. 
Knowledge may be incorporated into RL in various ways, through value, reward, policy, model, exploration strategy, etc. 
During a personal conversation with Rich Sutton, he mentioned that it is still wide open how to incorporate knowledge into RL.

 
 human intelligence, \citet{Lake2016}, 
 developmental start-up software ---
intuitive physics,
intuitive psychology;
learning as rapid model building ---
 compositionality,
 causality;
 learning to learn;
thinking fast ---
 approximate inference in structured models,
 model-based and model-free reinforcement learning
 
consciousness prior, \citet{Bengio2017}
 
ML with knowledge,  \citet{Song2017Knowledge}
 
causality, \citet{Pearl2018}, \citet{Johansson2016}

interpretability, \citet{Zhang2018Interpretability} surveyed visual interpretability for deep learning, \citet{Dong2017}

\citet{George2017} 

\citet{Yang2017KB}






\section{Important Mechanisms}

In this section, we discuss important mechanisms for the development of (deep) reinforcement learning, including attention and memory, unsupervised learning, transfer learning, multi-agent reinforcement learning, hierarchical RL, and learning to learn. We note that we do not discuss in detail some important mechanisms, like Bayesian RL~\citep{Ghavamzadeh2015}, POMDP~\citep{Hausknecht2015}, and semi-supervised RL~\citep{Audiffren2015, Finn2017, Zhu2009}.




\subsection{Attention and Memory}
\label{attention}

Attention is a mechanism to focus on the salient parts. Memory provides data storage for long time, and attention is an approach for memory addressing.

\citet{Grave-DNC-2016} proposed differentiable neural computer (DNC), in which, a neural network can read from and write to an external memory, so that DNC can solve complex, structured problems, which a neural network without read-write memory can not solve. DNC minimizes memory allocation interference and enables long-term storage.  Similar to a conventional computer, in a DNC, the neural network is the controller and the external memory is the random-access memory; and a DNC represents and manipulates complex data structures with the memory. Differently, a DNC learns such representation and manipulation end-to-end with gradient descent from data in a goal-directed manner. When trained with supervised learning, a DNC can solve synthetic question answering problems, for reasoning and inference in natural language; it can solve the shortest path finding problem between two stops in transportation networks and the relationship inference problem in a family tree. When trained with reinforcement learning, a DNC can solve a moving blocks puzzle with changing goals specified by symbol sequences. DNC outperformed normal neural network like LSTM or DNC's precursor Neural Turing Machine \citep{Graves2014}; with harder problems, an LSTM may simply fail. Although these experiments are relatively small-scale, we expect to see further improvements and applications of DNC. See Deepmind's description of DNC at https://deepmind.com/blog/differentiable-neural-computers/. 

\citet{Mnih-attention-2014} applied attention to image classification and object detection. \citet{Xu2015} integrated attention to image captioning. We briefly discuss application of attention in computer vision in Section~\ref{CV}. The attention mechanism is also deployed in NLP, e.g., in \citet{Bahdanau2015, Bahdanau2017}, and with external memory, in differentiable neural computer~\citep{Grave-DNC-2016} as discussed above. Most works follow a soft attention mechanism~\citep{Bahdanau2015}, a weighted addressing scheme to all memory locations. There are endeavours for hard attention~\citep{Gulcehre2016, Liang2017,Luo2016, Xu2015, Zaremba2015}, which is the way conventional computers access memory.

See recent work on attention and/or memory, e.g., \citet{Ba2014, Ba2016, Chen2016-knowledge, Danihelka2016, Duan2017OneShot, Eslami2016, Gregor2015, Jaderberg2015, Kaiser2016, Kadlec2016, Luo2016, Oh2016, Oquab2015, Vaswani2017, Weston2015, Sukhbaatar2015, Yang2015, Zagoruyko2017, Zaremba2015}. See http://distill.pub/2016/augmented-rnns/ and http://www.wildml.com/2016/01/attention-and-memory-in-deep-learning-and-nlp/ for blogs about attention and memory.

\subsection{Unsupervised Learning}
\label{unsupervised}

Unsupervised learning is a way to take advantage of the massive amount of data, and would be a critical mechanism to achieve general artificial intelligence.  Unsupervised learning is categorized into non-probabilistic models, like sparse coding, autoencoders,  k-means etc, and probabilistic (generative) models, where density functions are concerned, either explicitly or implicitly~\citep{Salakhutdinov2016}. Among probabilistic (generative) models with explicit density functions, some are with tractable models, like fully observable belief nets, neural autoregressive distribution estimators, and PixelRNN, etc; some are with non-tractable models, like Botlzmann machines, variational autoencoders,  Helmhotz machines, etc. For probabilistic (generative) models with implicit density functions, we have generative adversarial networks, moment matching networks, etc. 

In the following, we discuss Horde~\citep{Sutton2011}, and unsupervised auxiliary learning~\citep{Jaderberg2017}, two ways to take advantages of possible non-reward training signals in environments. We also discuss generative adversarial networks~\citep{Goodfellow2014}. See also \citet{Le2012}, \citet{Chen2016unpaired}, \citet{Liu2017unsupervised}.

\citet{Artetxe2017}

\subsubsection{Horde}

\citet{Sutton2011} proposed to represent knowledge with general value function, where policy, termination function, reward function, and terminal reward function are parameters. The authors then proposed Horde, a scalable real-time architecture for learning  in parallel general value functions for independent sub-agents from unsupervised sensorimotor interaction, i.e., nonreward signals and observations. Horde can learn to predict the values of many sensors, and  policies to maximize those sensor values, with general value functions, and answer predictive or goal-oriented questions. Horde is off-policy, i.e., it learns in real-time while following some other behaviour policy, and learns with gradient-based temporal difference learning methods, with constant time and memory complexity per time step.

\subsubsection{Unsupervised Auxiliary Learning}

Environments may contain abundant possible training signals, which may help to expedite achieving the main goal of maximizing the accumulative rewards, e.g., pixel changes may imply important events, and auxiliary reward tasks may help to achieve a good representation of rewarding states. This may be even helpful when the extrinsic rewards are rarely observed.


\citet{Jaderberg2017} proposed UNsupervised REinforcement and Auxiliary Learning (UNREAL) to improve learning efficiency by maximizing pseudo-reward functions, besides the usual cumulative reward, while sharing a common representation.  UNREAL is composed of RNN-LSTM base agent, pixel control, reward prediction, and value function replay. The base agent is trained on-policy with A3C~\citep{Mnih-A3C-2016}. Experiences of observations, rewards and actions are stored in a reply buffer, for being used by auxiliary tasks. The auxiliary policies use the base CNN and LSTM, together with a deconvolutional network, to maximize changes in pixel intensity of different regions of the input images. The reward prediction module predicts short-term extrinsic reward in next frame by observing the last three frames, to tackle the issue of reward sparsity. Value function replay further trains the value function.  UNREAL improved A3C's performance on Atari games, and performed well on 3D Labyrinth game. UNREAL has a shared representation among signals, while Horde trains each value function separately with distinct weights. See Deepmind's description of UNREAL at https://deepmind.com/blog/reinforcement-learning-unsupervised-auxiliary-tasks/.

We discuss robotics navigation with similar unsupervised auxiliary learning~\citep{Mirowski2017} in Section~\ref{robotics}. See also~\citet{Lample2017}.

\subsubsection{Generative Adversarial Networks}
\label{GANs}


\citet{Goodfellow2014} proposed generative adversarial nets (GANs) to estimate generative models via an adversarial process by training two models simultaneously, a generative model $G$ to capture the data distribution, and a discriminative model $D$ to estimate the probability that a sample comes from the training data but not the generative model $G$. 

\citet{Goodfellow2014} modelled $G$ and $D$ with multilayer perceptrons: $G(z: \bm{\theta}_g)$ and $D(x: \bm{\theta}_d)$, where $\bm{\theta}_g$ and $\bm{\theta}_d$ are parameters,  $x$ are data points, and $z$ are input noise variables. Define a prior on input noise variable $p_z(z)$. $G$ is a differentiable function and $D(x)$ outputs a scalar as the probability that $x$ comes from the training data rather than $p_g$, the generative distribution we want to learn.   

$D$ will be trained to maximize the probability of assigning labels correctly to samples from both training data and $G$. Simultaneously, $G$ will be trained to minimize such classification accuracy, $\log(1-D(G(z)))$. As a result, $D$ and $G$ form the two-player minimax game as follows:

\[\min_{G} \max_{D} E_{x \sim p_{data}(x)} [\log D(x)]  + E_{z \sim p_{z}(z)} [\log (1- D(G(z)))] \] 

\citet{Goodfellow2014} showed that as $G$ and $D$ are given enough capacity, generative adversarial nets can recover the data generating distribution, and provided a training algorithm with backpropagation by minibatch stochastic gradient descent. 

See \citet{Goodfellow2017} for Ian Goodfellow's summary of his NIPS 2016 Tutorial on GANs. GANs have received much attention and many works have been appearing after the tutorial.

GANs are notoriously hard to train. See \citet{Arjovsky2017WGAN} for Wasserstein GAN (WGAN) as a stable GANs model. \citet{Gulrajani2017} proposed to improve stability of WGAN by penalizing the norm of the gradient of the discriminator with respect to its input,  instead of clipping weights as in \citet{Arjovsky2017WGAN}.  \citet{Mao2016} proposed Least Squares GANs (LSGANs), another stable model. \citet{Berthelot2017} proposed BEGAN to improve WGAN by an equilibrium enforcing model, and set a new milestone in visual quality for image generation. \citet{Bellemare2017Cramer} proposed Cram{\'e}r GAN  to satisfy three machine learning properties of probability divergences: sum invariance, scale sensitivity, and unbiased sample gradients. \citet{Hu2017} unified GANs and Variational Autoencoders (VAEs). 

We discuss imitation learning with GANs in Section~\ref{reward}, including generative adversarial imitation learning, and third person imitation learning. \citet{Finn2016-Connection} established a connection between GANs, inverse RL, and energy-based models. \citet{Pfau2016} established the connection between GANs and actor-critic algorithms. See an answer on Quora, http://bit.ly/2sgtpx8, by Prof Sridhar Mahadevan.



\subsection{Transfer Learning}
\label{transfer}

Transfer learning is about transferring knowledge learned from different domains, possibly with different feature spaces and/or different data distributions~\citep{Taylor09, Pan2010, Weiss2016}. As reviewed in~\citet{Pan2010}, transfer learning can be inductive, transductive, or unsupervised; inductive transfer learning includes self-taught learning and multi-task learning; and transductive transfer learning includes domain adaptation and sample selection bias/covariance shift.

\citet{Bousmalis2017}

https://research.googleblog.com/2017/10/closing-simulation-to-reality-gap-for.html

\citet{Gupta2017} formulated the multi-skill problem for two agents to learn multiple skills, defined the common representation using which to map states and to project the execution of skills, and designed an algorithm for two agents to transfer the informative feature space maximally to transfer new skills, with similarity loss metric, autoencoder, and reinforcement learning.  The authors validated their proposed approach with two simulated robotic manipulation tasks.

See also recent work in transfer learning e.g., \citet{Andreas2017, Dong2015, Ganin2016, Kaiser2017OneModel, Kansky2017, Long2015, Long2016, Maurer2016, Mo2016, Parisotto2016, Papernot2017, Perez2017, Rajendran2017, WhyeTeh2017, Yosinski2014}. See~\citet{Ruder2017} for an overview about multi-task learning.
See NIPS 2015 Transfer and Multi-Task Learning: Trends and New Perspectives Workshop.

\citet{Long2017}

\citet{Killian2017}

\citet{Barreto2017}

\citet{McCann2017}


\subsection{Multi-Agent Reinforcement Learning}
\label{MARL}

Multi-agent RL (MARL) is the integration of multi-agent systems~\citep{Shoham2009, Stone2000} with RL, thus it is at the intersection of game theory~\citep{Leyton-Brown2008} and RL/AI communities.  Besides issues in RL like convergence and curse-of-dimensionality, there are new issues like multiple equilibria, and even fundamental issues like what is the question for multi-agent learning, whether convergence to an equilibrium is an appropriate goal, etc. Consequently, multi-agent learning is challenging both technically and conceptually, and demands clear understanding of the problem to be solved, the criteria for evaluation, and coherent research agendas~\citep{Shoham2007}.

Multi-agent systems have many applications, e.g., as we will discuss, games in Section~\ref{games}, robotics in Section~\ref{robotics}, Smart Grid in Section~\ref{smartgrid}, Intelligent Transportation Systems in Section~\ref{ITS}, and compute systems in Section~\ref{systems}.

\citet{Busoniu2008} surveyed works in multi-agent RL. There are several recent works, 
about new deep MARL algorithms~\citep{Foerster2018MAPG, Foerster2017, Lowe2017, Omidshafiei2017}, 
new communication mechanisms in MARL~\citep{Foerster2016,  Sukhbaatar2016}, and
sequential social dilemmas with MARL~\citep{Leibo2017}.


\citet{Bansal2017}

\citet{Al-Shedivat2017meta}

\citet{Ghavamzadeh2006}

\citet{Foerster2017opponent}

\citet{Perolat2017}

\citet{Lanctot2017}

\citet{Hadfield-Menell2016}

\citet{Hadfield-Menell2017}

\citet{MahdiElMhamdi2017}

\citet{Lowe2017}

\citet{Hoshen2017}

\subsection{Hierarchical Reinforcement Learning}

\label{hierarchical}

Hierarchical RL is a way to learn, plan, and represent knowledge with spatio-temporal abstraction at multiple levels. Hierarchical RL is an approach for issues of sparse rewards and/or long horizons~\citep{Sutton1999, Dietterich2000, Barto2003}.

\citet{Vezhnevets2016} proposed strategic attentive writer (STRAW), a deep recurrent neural network architecture, for learning high-level temporally abstracted macro-actions in an end-to-end manner based on observations from the environment. Macro-actions are sequences of actions commonly occurring.  STRAW builds a multi-step action plan, updated periodically based on observing rewards, and learns for how long to commit to the plan by following it without replanning. STRAW learns to discover macro-actions automatically from data, in contrast to the manual approach in previous work. \citet{Vezhnevets2016} validated STRAW on next character prediction in text, 2D maze navigation, and Atari games.

\citet{Kulkarni2016} proposed hierarchical-DQN (h-DQN) by organizing goal-driven intrinsically motivated deep RL modules hierarchically to work at different time-scales.  h-DQN integrates a top level action value function and a lower level action value function; the former learns a policy over intrinsic sub-goals, or options~\citep{Sutton1999}; the latter learns a policy over raw actions to satisfy given sub-goals. 
In a hard Atari game, Montezuma's Revenge, h-DQN outperformed  previous methods, including DQN and A3C.

\citet{Florensa2017} proposed to pre-train a large span of skills using Stochastic Neural Networks with an information-theoretic regularizer, then on top of these skills, to train high-level policies for downstream tasks.  Pre-training is based on a proxy reward signal, which is a form of intrinsic motivation to explore agent's own capabilities; its design requires minimal domain knowledge about the downstream tasks.  Their method combined hierarchical methods with intrinsic motivation, and the pre-training follows an unsupervised way.

\citet{Tessler2017} proposed a hierarchical deep RL network architecture for lifelong learning. Reusable skills, or sub-goals, are learned to transfer knowledge to new tasks. The authors tested their approach on the game of Minecraft. 

See also \citet{Bacon2017}, \citet{Kompella2017}, \citet{Machado2017}, \citet{Peng2017Dialogue}, \citet{Schaul2015}, \citet{Sharma2017}, \citet{Vezhnevets2017}, \citet{Yao2014}. See a survey on hierarchical RL~\citep{Barto2003}.

\citet{Harutyunyan2018}



\subsection{Learning to Learn}
\label{learning2learn}

Learning to learn, also know as meta-learning, is about learning to adapt rapidly to new tasks. It is related to transfer learning, multi-task learning, representation learning, and one/few/zero-shot learning. We can also see hyper-parameter learning and neural architecture design as learning to learn. It is a core ingredient to achieve strong AI~\citep{Lake2016}.

hypermarameter tuning, e.g., \citet{Jaderberg2017PBT}

\citet{Sutton1992}


\subsubsection{Learning to Learn/Optimize}

\citet{Li2017} proposed to automate unconstrained continuous optimization algorithms with guided policy search~\citep{Levine2016} by representing a particular optimization algorithm as a policy, and convergence rate as reward. 
See also \citet{Andrychowicz2016}. 

\citet{Duan2016RL2} and \citet{Wang2016LearnRL} proposed to learn a flexible RNN model to handle a family of RL tasks, to improve sample efficiency, learn new tasks in a few samples, and benefit from prior knowledge. 

combinatorial optimization, e.g., \citet{Vinyals2015}, \citet{Bello2016}, \citet{Dai2017graph}


\citet{Xu2017NTP}

\citet{Smith2017multi}

\citet{Li2017OptNN}

\subsubsection{Zero/One/Few-Shot Learning}
\label{one-shot-learning}

\citet{Lake2015} proposed an one-shot concept learning model, for handwritten characters in particular, with probabilistic program induction.
\citet{Koch2015} proposed siamese neural networks with metric learning for one-shot image recognition.
\citet{Vinyals2016} designed matching networks for one-shot classification.
\citet{Duan2017OneShot} proposed a model for one-shot imitation learning with attention for robotics.
\citet{Ravi2017} proposed a meta-learning model for few shot learning. 
\citet{Johnson2016} presented zero-shot translation for Google's multilingual neural machine translation system.
\citet{Kaiser2017} designed a large scale memory module for life-long one-shot learning to remember rare events.
\citet{Kansky2017} proposed Schema Networks for zero-shot transfer with a generative causal model of intuitive physics.
\citet{Snell2017} proposed prototypical networks for few/zero-shot classification by learning a metric space to compute distances to prototype representations of each class.

\subsubsection{Neural Architecture Design}

Neural networks architecture design is a notorious, nontrivial engineering issue.  Neural architecture search provides a promising avenue to explore.

\citet{Zoph2017} proposed the neural architecture search to generate neural networks architectures with an RNN trained by RL, in particular, REINFORCE, searching from scratch in variable-length architecture space, to maximize the expected accuracy of the generated architectures on a validation set. In the RL formulation, a controller generates hyperparameters as a sequence of tokens,  which are actions chosen from hyperparameters spaces; each gradient update to the policy parameters corresponds to training one generated network to convergence; an accuracy on a validation set is the reward signal. The neural architecture search can generate convolutional layers, with skip connections or branching layers, and recurrent cell architecture. The authors designed a parameter server approach to speed up training. Comparing with state-of-the-art methods, the proposed approach achieved competitive results for an image classification task with CIFAR-10 dataset; and better results for a language modeling task with Penn Treebank. 

\citet{Zoph2017Transfer}  proposed to transfer the architectural building block learned with the neural architecture search~\citep{Zoph2017} on small dataset to large dataset for scalable image recognition. \citet{Baker2017} proposed a meta-learning approach, using Q-learning with $\epsilon$-greedy exploration and experience replay, to generate CNN architectures automatically for a given learning task. \citet{Zhong2017} proposed to construct network blocks to reduce the search space of network design, trained by Q-learning. See also~\citet{Bello2017}.

There are recent works exploring new neural architectures. \citet{Kaiser2017OneModel} proposed to train a single model, MultiModel, which is composed of convolutional layers, an attention mechanism, and sparsely-gated layers, to learn multiple tasks from various domains, including image classification, image captioning and machine translation. \citet{Vaswani2017} proposed a new achichitecture for translation that replaces CNN and RNN with attention and positional encoding. \citet{Wang-Dueling-2016} proposed the dueling network architecture to estimate state value function and associated advantage function, to combine them to estimate action value function for faster convergence. \citet{Tamar2016} introduced Value Iteration Networks, a fully differentiable CNN planning module to approximate the value iteration algorithm, to learn to plan. \citet{Silver-Predictron-2016} proposed the predictron to integrate learning and planning into one end-to-end training procedure with raw input in Markov reward process.

\citet{Liu2017H}

\citet{Liu2017arch}


\section{Applications}

\begin{figure}
\includegraphics[width=1.0\linewidth]{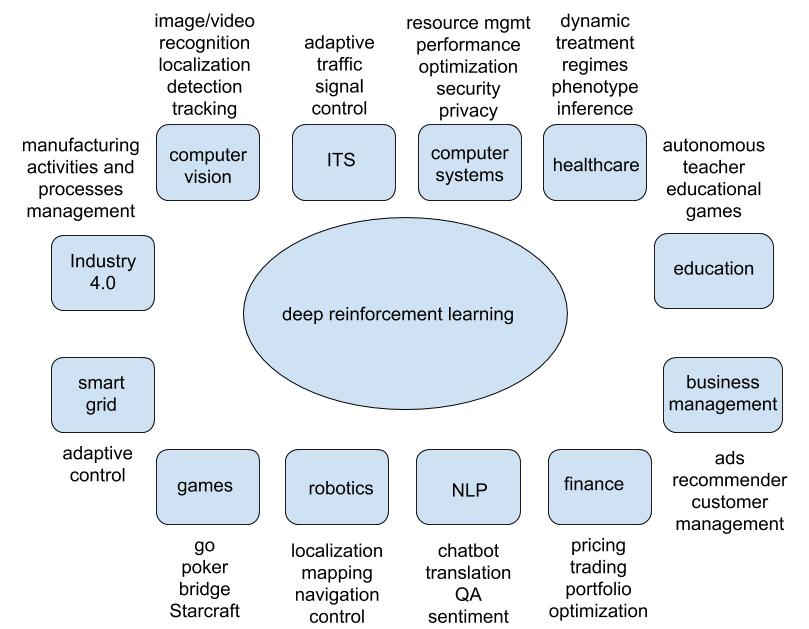}
\caption{Deep RL Applications}
\label{apps}
\end{figure}

Reinforcement learning has a wide range of applications.  
We discuss games in Section~\ref{games} and robotics in Section~\ref{robotics}, two classical RL application areas. 
Games will still be important testbeds for RL/AI. Robotics will be critical in the era of AI.
Next we discuss natural language processing in Section~\ref{NLP}, which enjoys wide and deep applications of RL recently. 
Computer vision follows in Section~\ref{CV}, in which, there are efforts for integration of vision and language.
Combinatorial optimization including neural architecture design in Section~\ref{combinatorial} is an exciting application of RL.
In Section~\ref{business}, we discuss business management,  like ads, recommendation, customer management, and marketing.
We discuss finance in Section~\ref{fin}. Business and finance have natural problems for RL.
We discuss healthcare in Section~\ref{healthcare}, which receives much attention recently, esp. after the success of deep learning. 
We discuss Industry 4.0 in Section~\ref{Industry40}. Many countries have made plans to integrate AI with manufacturing.
We discuss smart grid  in Section~\ref{smartgrid},
intelligent transportation systems  in Section~\ref{ITS},
and computer systems in Section~\ref{systems}. There are optimization and control problems in these areas, and many of them are concerned with networking and graphs. 
These application areas may overlap with each other, e.g., a robot may need skills for many of the application areas. 
We present deep RL applications briefly in Figure~\ref{apps}.

RL is usually for sequential decision making problems. However, some problems, seemingly non-sequential on surface, like machine translation and neural network architecture design, have been approached by RL. RL applications abound; and creativity would be the boundary.

Reinforcement learning is widely used in operations research~\citep{Powell11}, e.g., supply chain, inventory management, resource management, etc; we do not list it as an application area --- it is implicitly a component in application areas like intelligent transportation system and Industry 4.0. We do not list smart city, an important application area of AI, as it includes several application areas here: healthcare, intelligent transportation system, smart grid, etc. We do not discuss some interesting applications, like music generation~\citep{Briot2017, Jaques-2017}, and retrosynthesis~\citep{Segler2017}.  See previous work on lists of RL applications at: http://bit.ly/2pDEs1Q, and http://bit.ly/2rjsmaz. We may only touch the surface of some application areas. 
 It is desirable to do a deeper analysis of all application areas listed in the following, which we leave as a future work.

\subsection{Games}
\label{games}

Games provide excellent testbeds for RL/AI algorithms. We discuss Deep Q-Network (DQN) in Section~\ref{DQN} and its extensions, all of which experimented with Atari games. We discuss \citet{Mnih-A3C-2016} in Section~\ref{A3C}, \citet{Jaderberg2017} in Section~\ref{unsupervised}, and \citet{Mirowski2017} in Section~\ref{robotics}, and they used Labyrinth as the testbed. See~\citet{Yannakakis2018} for a book on artificial intelligence and games.
We discuss multi-agent RL  in Section~\ref{MARL}, which is at the intersection of game theory and RL/AI.

Backgammon and computer Go are perfect information board games. In Section~\ref{boardgames}, we discuss briefly Backgammon, and focus on computer Go, in particular, AlphaGo.  Variants of card games, including majiang/mahjong, are imperfect information board games, which we discuss in Section~\ref{imperfectgames}, and focus on Texas Hold'em Poker.  In video games, information may be perfect or imperfect, and game theory may be deployed or not. We discuss video games in Section~\ref{videogames}. We will see more achievements in imperfect information games and video games, and their applications.

\subsubsection{Perfect Information Board Games}
\label{boardgames}

Board games like Backgammon, Go, chess, checker and Othello, are classical testbeds for RL/AI algorithms. In such games, players reveal prefect information. \citet{Tesauro1994} approached Backgammon by using neural networks to approximate value function learned with TD learning, and achieved human level performance.  We focus on computer Go, in particular, AlphaGo~\citep{Silver-AlphaGo-2016, Silver-AlphaGo-2017},  for its significance. 

\subsubsection*{Computer Go}
\label{AlphaGo}


The challenge of solving Computer Go comes from not only the gigantic search space of size $250^{150}$, an astronomical number, but also the hardness of position evaluation~\citep{Muller2002}, which was successfully used in solving many other games, like Backgammon and chess.

AlphaGo~\citep{Silver-AlphaGo-2016}, a computer Go program, won the human European Go champion, 5 games to 0, in October 2015, and became the first computer Go program to won a human professional Go player without handicaps on a full-sized 19 $\times$ 19 board. Soon after that in March 2016, AlphaGo defeated Lee Sedol, an 18-time world champion Go player, 4 games to 1, making headline news worldwide. This set a landmark in AI.  AlphaGo defeated Ke Jie 3:0 in May 2017. AlphaGo Zero~\citep{Silver-AlphaGo-2017} further improved previous versions by learning a superhuman computer Go program without human knowledge.

\subsubsection*{AlphaGo: Training pipeline and MCTS}

We discuss briefly how AlphaGo works based on \citet{Silver-AlphaGo-2016}  and \citet{Sutton2018}. See \citet{Sutton2018} for a  detailed and intuitive description of AlphaGo. See Deepmind's description of AlphaGo at goo.gl/lZoQ1d.

AlphaGo was built with techniques of deep convolutional neural networks,  supervised learning, reinforcement learning, and Monte Carlo tree search (MCTS)~\citep{Browne2012, Gelly2007, Gelly12}. AlphaGo is composed of two phases: neural network training pipeline and MCTS. The training pipeline phase includes training a supervised learning (SL) policy network from expert moves, a fast rollout policy,  a RL policy network, and a RL value network.  

The SL policy network has convolutional layers, ReLU nonlinearities, and an output softmax layer representing probability distribution over legal moves. The inputs to the CNN are 19 $\times$ 19 $\times$ 48 image stacks, where 19 is the dimension of a Go board and 48 is the number of features. State-action pairs are sampled from expert moves to train the network with stochastic gradient ascent to maximize the likelihood of the move selected in a given state. The fast rollout policy uses a linear softmax with small pattern features. 

The RL policy network improves SL policy network, with the same network architecture, and the weights of SL policy network as initial weights, and policy gradient for training. The reward function is +1 for winning and -1 for losing in the terminal states, and 0 otherwise. Games are played between the current policy network and a random, previous iteration of the policy network, to stabilize the learning and to avoid overfitting. Weights are updated by stochastic gradient ascent to maximize the expected outcome.

The RL value network still has the same network architecture as SL policy network, except the output is a single scalar predicting the value of a position. The value network is learned in a Monte Carlo policy evaluation approach. To tackle the overfitting problem caused by strongly correlated successive positions in games, data are generated by self-play between the RL policy network and itself until game termination.  The weights are trained by regression on state-outcome pairs, using stochastic gradient descent to minimize the mean squared error between the prediction and the corresponding outcome. 

In MCTS phase, AlphaGo selects moves by lookahead search. It builds a partial game tree starting from the current state, in the following stages: 1) select a promising node to explore further, 2) expand a leaf node guided by the SL policy network and collected statistics, 3) evaluate a leaf node with a mixture of the RL value network and the rollout policy, 4) backup evaluations to update the action values.  A move is then selected.    

\subsubsection*{AlphaGo Zero}

AlphaGo Zero can be understood as an approximation policy iteration, incorporating MCTS inside the training loop to perform both policy improvement and policy evaluation.
MCTS may be regarded as a policy improvement operator. It outputs move probabilities stronger than raw probabilities of the neural network.
Self-play with search may be regarded as a policy evaluation operator. It uses MCTS to select moves, and game winners as samples of value function.
Then the policy iteration procedure updates the neural network's weights to match the move probabilities and value more closely with the improved search probabilities and self-play winner, and conduct self-play with updated neural network weights in the next iteration to make the search stronger.

The features of AlphaGo Zero~\citep{Silver-AlphaGo-2017}, comparing with AlphaGo~\citep{Silver-AlphaGo-2016}, are: 
1) it learns from random play, with self-play reinforcement learning, without human data or supervision;
2) it uses black and white stones from the board as input, without any manual feature engineering;
3) it uses a single neural network to represent both policy and value, rather than separate policy network and value network; and
4) it utilizes the neural network for position evaluation and move sampling for MCTS, and it does not perform Monte Carlo rollouts. 
AlphaGo Zero deploys several recent achievements in neural networks:  residual convolutional neural networks (ResNets), batch normalization, and rectifier nonlinearities.

AlphaGo Zero has three main components in its self-play training pipeline executed in parallel asynchronously: 1) optimize neural network weights from recent self-play data continually; 2) evaluate players continually; 3) use the strongest player to generate new self-play data.  

When AlphaGo Zero playing a game against an opponent, MCTS searches from the current state, with the trained neural network weights, to generate move probabilities, and then selects a move.

We present a brief, conceptual pseudo code  in Algorithm~\ref{AlphaGoZero} for training in AlphaGo Zero, conducive for easier understanding. Refer to the original paper~\citep{Silver-AlphaGo-2017} for details.

\begin{algorithm}
\SetAlgoNoLine
\textbf{Input: } the raw board representation of the position, its history, and the colour to play as 19 $\times$ 19 images; game rules; a game scoring function; invariance of game rules under rotation and reflection, and invariance to colour transposition except for komi\\
\textbf{Output: } policy (move probabilities) $p$, value $v$\\
\vspace{3mm}
initialize neural network weights $\theta_0$ randomly\\
//AlphaGo Zero follows a policy iteration procedure\\
\For{each iteration $i$}{
        \vspace{3mm}
        // termination conditions: \\
        // 1. both players pass \\
        // 2. the search value drops below a resignation threshold \\
        // 3. the game exceeds a maximum length \\
        \vspace{3mm}
        initialize $s_0$\\
        
	\For{each step $t$, until termination at step $T$}{
		\vspace{3mm}
		
		// MCTS can be viewed as a policy improvement operator \\
		// search algorithm: asynchronous policy and value MCTS algorithm (APV-MCTS)\\
	        // execute an MCTS search $\pi_{t} = \alpha_{\theta_{i-1}}(s_t)$ with previous neural network $f_{\theta_{i-1}}$\\
	        // each edge $(s, a)$ in the search tree stores a prior probability $P(s, a)$, a visit count $N(s, a)$, and an action value $Q(s, a)$\\
		\While{computational resource remains} {
		select: each simulation traverses the tree by selecting the edge with maximum upper confidence bound $Q(s, a) + U(s, a)$,
		where $U(s, a) \propto P(s, a) / (1 + N(s, a))$ \\
		expand and evaluate: the leaf node is expanded and the associated position $s$ is evaluated by the neural network, $(P(s, \cdot), V(s)) = f_{\theta_i}(s)$; the vector of $P$ values are stored in the outgoing edges from $s$\\
		backup: each edge $(s, a)$ traversed in the simulation is updated to increment its visit count $N(s, a)$, and to update its action value to the mean evaluation over these simulations, $Q(s,a) = 1/N(s, a) \sum_{s'|s, a \rightarrow s'} V(s')$, where $s'|s, a \rightarrow s'$ indicates that a simulation eventually reached $s'$ after taking move $a$ from position $s$\\
		}
	
		 // self-play with search  can be viewed as a policy evaluation operator: select each move with the improved MCTS-based policy, uses the game winner as a sample of the value\\ 
		play: once the search is complete, search probabilities $\pi \propto N^{1/\tau}$ are returned, where $N$ is the visit count of each move from root and $\tau$ is a parameter controlling temperature;      
	        play a move by sampling the search probabilities $\pi_{t}$, transition to next state $s_{t+1}$\\     
	}
	\vspace{3mm}
	score the game to give a final reward $r_{T} \in \{-1, +1\}$\\
	\For{each step $t$ in the last game}{
	        $z_t \leftarrow \pm r_{T}$, the game winner from the perspective of the current player\\
	        store data as $(s_t, \pi_{t}, z_t)$\\     
	}
	sample data $(s, \pi, z)$ uniformly among all time-steps of the last iteration(s) of self-play\\
	\vspace{3mm}
	//train neural network weights $\theta_i$\\
	//optimizing loss function $l$ performs both policy evaluation, via $(z-v)^2$, and policy improvement, via $- \pi^{T} \log p$, in a single step\\
	adjust the neural network $(p, v) = f_{\theta_i}(s)$: \\
	to minimize the error between the predicted value $v$ and the self-play winner $z$, and \\
	to maximize similarity of neural network move probabilities $p$ to search probabilities $\pi$\\
	
	specifically, adjust the parameters $\theta$ by gradient descent on a loss function \\
	$(p, v) = f_{\theta_i}(s)$ and $l = (z-v)^2 - \pi^{T} \log p + c \| \theta_i \|^2 $ \\
	$l$ sums over the mean-squared error and cross-entropy losses, respectively \\
	$c$ is a parameter controlling the level of $L2$ weight regularization to prevent overfitting\\
	\vspace{3mm}
	evaluate the checkpoint every 1000 training steps to decide if replacing the current best player (neural network weights) for generating next batch of self-play games\\ 
}
\caption{AlphaGo Zero training pseudo code, based on~\citet{Silver-AlphaGo-2017}}
\label{AlphaGoZero}
\end{algorithm}

\citet{Silver-AlphaZero-2017}

\subsubsection*{Discussions}

AlphaGo Zero is a reinforcement learning algorithm. It is neither supervised learning nor unsupervised learning. 
The game score is a reward signal, not a supervision label. 
Optimizing the loss function $l$ is supervised learning.
However, it performs policy evaluation and policy improvement, as one iteration in policy iteration.

AlphaGo Zero is not only a heuristic search algorithm. AlphaGo Zero is a policy iteration procedure, in which, heuristic search, in particular, MCTS, plays a critical role, but within the scheme of reinforcement learning policy iteration, as illustrated in the pseudo code in Algorithm~\ref{AlphaGoZero}. MCTS can be viewed as a policy improvement operator.


AlphaGo attains a superhuman level. It may confirm that professionals have developed effective  strategies. However, it does not need to mimic professional plays. Thus it does not need to predict their moves correctly.

The inputs to AlphaGo Zero include the raw board representation of the position, its history, and the colour to play as 19 $\times$ 19 images; game rules; a game scoring function; invariance of game rules under rotation and reflection, and invariance to colour transposition except for komi. An additional and critical input is  solid research and development experiences.

AlphaGo Zero utilized 64 GPU workers (each maybe with multiple GPUs) and 19 CPU parameter servers (each with multiple CPUs) for training, around 2000 TPUs for data generation, and 4 TPUs for game playing. The computation cost is too formidable for replicating AlphaGo Zero.

AlphaGo requires huge amount of data for training, so it is still a big data issue. However, the data can be generated by self play, with a perfect model or precise game rules.

Due to the perfect model or precise game rules for computer Go, AlphaGo algorithms have their limitations. For example, in healthcare, robotics and self driving problems, it is usually hard to collect a large amount of data, and it is hard or impossible to have a close enough or even perfect model. As such, it is nontrivial to directly apply AlphaGo algorithms to such applications. 

On the other hand, AlphaGo algorithms, especially, the underlying techniques, namely, deep learning, reinforcement learning, and Monte Carlo tree search, have many applications. \citet{Silver-AlphaGo-2016} and~\citet{Silver-AlphaGo-2017} recommended the following applications: 
general game-playing (in particular, video games), classical planning, partially observed planning, scheduling, constraint satisfaction,  robotics, industrial control, and online recommendation systems.
AlphaGo Zero blog mentioned the following structured problems: protein folding, reducing energy consumption, and searching for revolutionary new materials.\footnote{Andrej Karpathy posted a blog titled "AlphaGo, in context", after AlphaGo defeated Ke Jie in May 2017. He characterized properties of Computer Go as: fully deterministic, fully observable, discrete action space, accessible perfect simulator, relatively short episode/game, clear and fast evaluation conducive for many trail-and-errors, and huge datasets of human play games, to illustrate the narrowness of AlphaGo. It is true that computer Go has limitations in the problem setting and thus potential applications, and is far from artificial general intelligence. However, we see the success of AlphaGo as the triumph of AI, in particular, AlphaGo's underlying techniques, i.e., learning from demonstration (as supervised learning), deep learning, reinforcement learning, and Monte Carlo tree search; these techniques are present in many recent achievements in AI. As a whole technique, AlphaGo will probably shed lights on classical AI areas, like planning, scheduling, and constraint satisfaction~\citep{Silver-AlphaGo-2016}, and new areas for AI, like retrosynthesis~\citep{Segler2017}. Reportedly, the success of AlphaGo's conquering titanic search space inspired quantum physicists to solve the quantum many-body problem~\citep{Carleo2017}.}

AlphaGo has made tremendous progress, and set a landmark in AI. However, we are still far away from attaining artificial general intelligence (AGI).

It is interesting to see how strong a raw deep neural network in AlphaGo can become, and how soon a very strong computer Go program would be available on a mobile phone.

\subsubsection{Imperfect Information Board Games}
\label{imperfectgames}

Imperfect information games, or game theory in general, have many applications, e.g., security and medical decision support.  It is interesting to see more progress of deep RL in such applications, and the full version of Texas Hold'em.

\citet{Heinrich2016} proposed Neural Fictitious Self-Play (NFSP) to combine fictitious self-play with deep RL to learn approximate Nash equilibria for games of imperfect information in a scalable end-to-end approach without prior domain knowledge. NFSP was evaluated on two-player zero-sum games. In Leduc poker, NFSP approached a Nash equilibrium, while common RL methods diverged. In Limit Texas Hold'em, a real-world scale imperfect-information game, NFSP performed similarly  to state-of-the-art, superhuman algorithms which are based on significant domain expertise. 

Heads-up Limit Hold'em Poker was essentially solved~\citep{Bowling2015} with counterfactual regret minimization (CFR), which is  an iterative method to approximate a Nash equilibrium of an extensive-form game with repeated self-play between two regret-minimizing algorithms. 

\subsubsection*{DeepStack}

Recently, significant progress has been made for Heads-up No-Limit Hold'em Poker~\citep{Moravcik2017}, the DeepStack computer program defeated professional poker players for the first time. DeepStack utilized the recursive reasoning of CFR to handle information asymmetry, focusing computation on specific situations arising when making decisions and use of value functions trained automatically, with little domain knowledge or human expert games, without  abstraction and offline computation of complete strategies  as before.  





\subsubsection{Video Games}
\label{videogames}

Video games would be great testbeds for artificial general intelligence.

\citet{Wu2017} deployed A3C with CNN to train an agent in a partially observable 3D environment, Doom, from recent four raw frames and game variables, to predict next action and value function, following the curriculum learning~\citep{Bengio-2009-Curriculum} approach of starting with simple tasks and gradually transition to harder ones.  It is nontrivial to apply A3C to such 3D games directly, partly due to sparse and long term reward. The authors won the champion in Track 1 of ViZDoom Competition by a large margin, and plan the following future work: a map from an unknown environment, localization, a global plan to act, and visualization of the reasoning process.

\citet{Dosovitskiy2017} approached the problem of sensorimotor control in immersive environments with supervised learning, and won the Full Deathmatch track of the Visual Doom AI Competition.  We list it here since it is usually a RL problem, yet it was solved with supervised learning. \citet{Lample2017} also discussed how to tackle Doom. 

\citet{Peng2017} proposed a multiagent actor-critic framework, with a bidirectionally-coordinated network to form coordination among multiple agents in a team, deploying the concept of dynamic grouping and parameter sharing for better scalability.  
The authors used StarCraft as the testbed.
Without human demonstration or labelled data as supervision, the proposed approach learned strategies for coordination similar to the level of experienced human players, like move without collision, hit and run, cover attack, and focus fire without overkill.
\citet{Usunier2016, Justesen2017} also studied StarCraft.

\citet{Oh2016} and \citet{Tessler2017} studied Minecraft, 
\citet{Chen2017, Firoiu2017} studied Super Smash Bros,
and \citet{Kansky2017} proposed Schema Networks and empirically studied variants of Breakout in Atari games. 

See \citet{Justesen2017Survey} for a survey about applying deep (reinforcement) learning to video games.
See \citet{Ontanon2013} for a survey about Starcraft. Check AIIDE and CIG Starcraft AI Competitions, 
and its history at https://www.cs.mun.ca/\textasciitilde dchurchill/starcraftaicomp/history.shtml.
See \citet{Lin2017} for StarCraft Dataset.

\subsection{Robotics}
\label{robotics}


Robotics is a classical area for reinforcement learning. 
See~\citet{Kober2013} for a survey of RL in robotics, \citet{Deisenroth2013} for a survey on policy search for robotics, and
\citet{Argall2009} for a survey of robot learning from demonstration. 
See the journal Science Robotics.
It is interesting to note that from NIPS 2016 invited talk, Boston Dynamics robots did not use machine learning.

In the following, we discuss guided policy search~\citep{Levine2016} and learn to navigate~\citep{Mirowski2017}.
See more recent robotics papers, e.g., \citet{Chebotar2016, Chebotar2017, Duan2017OneShot, Finn-robotics-2016, Gu-robotics-2016, Lee2017, Levine2016-grasp, Mahler2017, Perez2017, Popov2017, Yahya2016, Zhu2017}. 

We recommend Pieter Abbeel's NIPS 2017 Keynote Speech, Deep Learning for Robotics, slides at, https://www.dropbox.com/s/fdw7q8mx3x4wr0c/


\subsubsection{Guided Policy Search}
\label{GPS}

\citet{Levine2016} proposed to train the perception and control systems jointly end-to-end, to map raw image observations directly to torques at the robot's motors. The authors introduced guided policy search (GPS) to train policies represented as CNN, by transforming policy search into supervised learning to achieve data efficiency, with training data provided by a trajectory-centric RL method operating under unknown dynamics. GPS alternates between trajectory-centric RL and supervised learning, to obtain the training data coming from the policy's own state distribution, to address the issue that supervised learning usually does not achieve good, long-horizon performance. GPS utilizes pre-training to reduce the amount of experience data to train visuomotor policies. Good performance was achieved on a range of real-world manipulation tasks requiring localization, visual tracking, and handling complex contact dynamics, and simulated comparisons with previous policy search methods. As the authors mentioned, "this is the first method that can train deep visuomotor policies for complex, high-dimensional manipulation skills with direct torque control". 

\subsubsection{Learn to Navigate}
\label{navigate}

\citet{Mirowski2017} obtained the navigation ability by solving a RL problem maximizing cumulative reward and  jointly considering un/self-supervised tasks to improve data efficiency and task performance.  The authors addressed the sparse reward issues by augmenting the loss with two auxiliary tasks, 1) unsupervised reconstruction of a low-dimensional depth map for representation learning to aid obstacle avoidance and short-term trajectory planning; 2) self-supervised loop closure classification task within a local trajectory. The authors incorporated a stacked LSTM to use memory at different time scales for dynamic elements in the environments. The proposed agent learn to navigate in complex 3D mazes end-to-end from raw sensory input, and performed similarly to human level, even when start/goal locations change frequently.

In this approach, navigation is a by-product of the goal-directed RL optimization problem, in contrast to conventional approaches such as Simultaneous Localisation and Mapping (SLAM), where explicit position inference and mapping are used for navigation. This may have the chance to replace the popular SLAM, which usually requires manual processing.

\subsection{Natural Language Processing}
\label{NLP}

In the following we talk about natural language processing (NLP), dialogue systems in Section~\ref{chatbot}, machine translation in Section~\ref{translation},  and text generation in Section~\ref{textgeneration}.  There are many interesting issues in NLP, and we list some in the following.
\begin{itemize}
\item language tree-structure learning, e.g., \citet{Socher2011, Socher2013, Yogatama2017}
\item semantic parsing, e.g., \citet{Liang2017NSM}
\item question answering, e.g., \citet{Celikyilmaz2017}, \citet{Shen2017}, \citet{Trischler2016}, \citet{Xiong2017}, and \citet{Wang2017RNet}, \citet{Choi2017}
\item summarization, e.g., \citet{Paulus2017, Zhang2017sentence}
\item sentiment analysis~\citep{Liu2012, Zhang2018sentiment}, e.g., \citet{Radford2017}
\item information retrieval~\citep{Manning2008}, e.g., \citet{Zhang2016}, and \citet{Mitra2017}
\item information extraction, e.g., \citet{Narasimhan2016}
\item automatic query reformulation, e.g., \citet{Nogueira2017}
\item language to executable program, e.g., \citet{Guu2017}
\item knowledge graph reasoning, e.g., \citet{Xiong2017DeepPath}
\item text games, e.g., \citet{WangSida2016}, \citet{He2016-textgame}, and \citet{Narasimhan2015}
\end{itemize}

Deep learning has been permeating into many subareas in NLP, and helping make significant progress. The above is a partial list.
It appears that NLP is still a field, more about synergy than competition, for deep learning vs. non-deep learning algorithms, and for approaches based on no domain knowledge (end-to-end) vs linguistics knowledge.
Some non-deep learning algorithms are effective and perform well, e.g., word2vec~\citep{Mikolov2013, Mikolov2017} and fastText~\citep{Joulin2017}, and many works that study syntax and semantics of languages,
see a recent example in semantic role labeling~\citep{He2017SRL}. 
Some deep learning approaches to NLP problems incorporate explicitly or implicitly linguistics knowledge, e.g., \citet{Socher2011, Socher2013, Yogatama2017}. See an article by Christopher D. Manning, titled "Last Words: Computational Linguistics and Deep Learning, A look at the importance of Natural Language Processing", at http://mitp.nautil.us/article/170/last-words-computational-linguistics-and-deep-learning. 

\citet{Melis2017}

\subsubsection{Dialogue Systems}
\label{chatbot}

In dialogue systems, conversational agents, or chatbots, human and computer interacts with natural language. We intentionally remove "spoken" before "dialogue systems" to accommodate both spoken and written language user interface (UI). 
\citet{Jurafsky2017} categorize dialogue systems as task-oriented dialog agents and chatbots; the former are set up to have short conversations to help complete particular tasks; the latter are set up to mimic human-human interactions with extended conversations, sometimes with entertainment value.
As in~\citet{Deng2017AIFrontiers}, there are four categories: social chatbots, infobots (interactive question answering), task completion bots (task-oriented or goal-oriented) and personal assistant bots. We have seen generation one dialogue systems: symbolic rule/template based, and generation two: data driven with (shallow) learning. We are now experiencing generation three: data driven with deep learning, and reinforcement learning usually play an important role.  A dialogue system usually include the following modules: (spoken) language understanding, dialogue manager (dialogue state tracker and dialogue policy learning), and a natural language generation~\citep{Young2013}. In task-oriented systems, there is usually a knowledge base to query. A deep learning approach, as usual, attempts to make the learning of the system parameters end-to-end. See~\citet{Deng2017AIFrontiers} for more details.
See a survey paper on applying machine learning to speech recognition~\citep{Deng2013}.



\cite{Li2017TC} presented an end-to-end task-completion neural dialogue system with parameters learned by supervised and reinforcement learning. The proposed framework includes a user simulator~\citep{LiXiujun2016} and a neural dialogue system. The user simulator consists of user agenda modelling and natural language generation. The neural dialogue system is composed of language understanding and dialogue management (dialogue state tracking and policy learning). The authors deployed RL to train dialogue management end-to-end, representing the dialogue policy as a deep Q-network~\citep{Mnih-DQN-2015}, with the tricks of a target network and a customized experience replay, and using a rule-based agent to warm-start the system with supervised learning. The source code is available at http://github.com/MiuLab/TC-Bot.

\citet{Dhingra2017} proposed KB-InfoBot, a goal-oriented dialogue system for multi-turn information access. KB-InfoBot is trained end-to-end using RL from user feedback with differentiable operations, including those for accessing external knowledge database (KB). In previous work, e.g., \cite{Li2017TC} and \citet{WenTH2017}, a dialogue system accesses real world knowledge from KB by symbolic, SQL-like operations, which is non-differentiable and disables the dialogue system from fully end-to-end trainable. KB-InfoBot achieved the differentiability by inducing a soft posterior distribution over the KB entries to indicate which ones the user is interested in.  The authors designed a modified version of the episodic REINFORCE algorithm to explore and learn both the policy to select dialogue acts and the posterior over the  KB entries for correct retrievals.The authors deployed imitation learning from rule-based belief trackers and policy to warm up the system.

\citet{Su2016} proposed an on-line learning framework to train the dialogue policy jointly with the reward model via active learning with a Gaussian process model, to tackle the issue that it is unreliable and costly to use explicit user feedback as the reward signal. The authors showed empirically that the proposed framework reduced manual data annotations significantly and mitigated noisy user feedback in dialogue policy learning. 


\citet{LiJiwei2016} proposed to use deep RL to generate dialogues to model future reward for better informativity, coherence, and ease of answering, to attempt to address the issues in the sequence to sequence models based on \citet{Sutskever2014}: the myopia and misalignment of maximizing the probability of generating a response given the previous dialogue turn, and the infinite loop of repetitive responses. The authors designed a reward function to reflect the above desirable properties, and deployed policy gradient to optimize the long term reward.  It would be interesting to investigate the reward model with the approach in~\citet{Su2016} or with inverse RL and imitation learning as discussed in Section~\ref{reward}, although \citet{Su2016} mentioned that such methods are costly, and humans may not act optimally. 

Some recent papers follow: \citet{ElAsri2016}, \citet{Bordes2017}, \citet{Chen2016}, \citet{Eric2017}, \citet{Fatemi2016}, \citet{Kandasamy2017}, \citet{Lewis2017}, \citet{LiJiwei2016-fast},  \citet{LiJiwei2017human}, \citet{LiJiwei2017Q}, \citet{Lipton2016},  \citet{Mesnil2015}, \citet{Mo2016}, \citet{Peng2017Dialogue}, \citet{Saon2016}, \citet{Serban2017}, \citet{Shah2016}, \citet{She2017}, \citet{Su2016-continuous}, \citet{Weiss2017}, \citet{WenTH2015}, \citet{WenTH2017}, \citet{Williams2016}, \citet{Williams2017}, \citet{XiongW2017}, \citet{XiongW2017MS}, \citet{Yang2016}, \citet{Zhang2017Speak}, \citet{Zhang2017Speech}, \citet{Zhao2016}, \citet{Zhou2017Emotional}. 
See \citet{Serban2015} for a survey of corpora for building dialogue systems.

See NIPS 2016 Workshop on End-to-end Learning for Speech and Audio Processing, and NIPS 2015 Workshop on Machine Learning for Spoken Language Understanding and Interactions.





\subsubsection{Machine Translation}
\label{translation}

Neural machine translation~\citep{Kalchbrenner2013, Cho2014, Sutskever2014, Bahdanau2015} utilizes end-to-end deep learning for machine translation, and becomes dominant, against the traditional statistical machine translation techniques.  The neural machine translation approach usually first encodes a variable-length source sentence,  and then decodes it to a variable-length target sentence. \citet{Cho2014} and \citet{Sutskever2014} used two RNNs to encode a sentence to a fix-length vector and then decode the vector into a target sentence.  \citet{Bahdanau2015} introduced the soft-attention technique to learn to jointly align and translate. 

\citet{He2016} proposed dual learning mechanism to tackle the data hunger issue in machine translation, inspired by the observation that the information feedback between the primal, translation from language A to language B, and the dual, translation from B to A, can help improve both translation models, with a policy gradient method, using the language model likelihood as the reward signal.  Experiments showed that, with only 10\% bilingual data for warm start and monolingual data, the dual learning approach performed comparably with previous neural machine translation methods with full bilingual data in English to French tasks. The dual learning mechanism may have extensions to many tasks, if the task has a dual form, e.g.,  speech recognition and text to speech, image caption and image generation, question answering and question generation, search and keyword extraction, etc.

See \citet{Wu2016, Johnson2016} for Google's Neural Machine Translation System; \citet{Gehring2017} for convolutional sequence to sequence learning for fast neural machine translation; \citet{Klein2017opennmt} for OpenNMT, an open source neural machine translation system;  \citet{Cheng2016} for semi-supervised learning for neural machine translation, and \citet{Wu2017AdversarialNMT} for adversarial neural machine translation.  See \citet{Vaswani2017} for a new approach for translation that replaces CNN and RNN with attention and positional encoding. 
See~\citet{Zhang2017NMT} for an open source toolkit for neural machine translation.
See~\citet{Monroe2017} for a gentle introduction to translation.

\citet{Artetxe2017}




\subsubsection{Text Generation}
\label{textgeneration}

Text generation is the basis for many NLP problems, like conversational response generation, machine translation, abstractive summarization, etc.

Text generation models are usually based on $n$-gram, feed-forward neural networks, or recurrent neural networks, trained to predict next word given the previous ground truth words as inputs; then in testing, the trained models are used to generate a sequence word by word, using the generated words as inputs. The errors will accumulate on the way, causing the exposure bias issue. Moreover, these models are trained with word level losses, e.g., cross entropy, to maximize the probability of next word; however, the models are evaluated on a different metrics like BLEU.

\citet{Ranzato2016} proposed Mixed Incremental Cross-Entropy Reinforce (MIXER) for sequence prediction, with incremental learning and a loss function combining both REINFORCE and cross-entropy. MIXER is a sequence level training algorithm, aligning training and testing objective, such as BLEU, rather than predicting the next word as in previous works. 

\citet{Bahdanau2017}  proposed an actor-critic algorithm for sequence prediction, attempting to further improve \citet{Ranzato2016}. The authors utilized a critic network to predict the value of a token, i.e., the expected score following the sequence prediction policy, defined by an actor network, trained by the predicted value of tokens. Some techniques are deployed to improve performance: SARSA rather than Monter-Carlo method to lessen the variance in estimating value functions; target network for stability; sampling prediction from a delayed actor whose weights are updated more slowly than the actor to be trained, to avoid the feedback loop when actor and critic need to be trained based on the output of each other; reward shaping to avoid the issue of sparse training signal.  

\citet{Yu2017} proposed SeqGAN,  sequence generative adversarial nets with policy gradient, integrating the adversarial scheme in \citet{Goodfellow2014}. \citet{LiJiwei2017future} proposed to improve sequence generation by considering the knowledge about the future.

\subsection{Computer Vision}
\label{CV}


Computer vision is about how computers gain understanding from digital images or videos. 
In the following, after presenting background in computer vision,  we discuss recognition, motion analysis, scene understanding, integration with NLP, and visual control. 

Reinforcement learning would be an important ingredient for interactive perception~\citep{Bohg2017}, where perception and interaction with the environment would be helpful to each other, in tasks like object segmentation, articulation model estimation, object dynamics learning and haptic property estimation, object recognition or categorization, multimodal object model learning, object pose estimation, grasp planning, and manipulation skill learning.

More topics about applying deep RL to computer vision: 
\begin{itemize}
\item \citet{Liu2017F} for semantic parsing of large-scale 3D point clouds;  
\item \citet{Kaba2017} for view planning, which is a set cover problem; 
\item \citet{Cao2017} for face hallucination, i.e., generating a high-resolution face image from a low-resolution input image;
\item \citet{Brunner2018} for learning to read maps;
\item \citet{Bhatti2016} for SLAM-augmented DQN.
\end{itemize}

\subsubsection{Background}

Todo: 
AlexNet~\citep{Krizhevsky2012}, 
ResNets~\citep{He2016-ResNets} 
DenseNets~\citep{Huang2017DenseNet},
Fast R-CNN~\citep{Girshick2015}, Faster R-CNN~\citet{Ren2015}, Mask R-CNN~\citet{He2017Mask},
\citet{Shrivastava2017},
VAEs (variational autoencoder)~\citep{Kingma2014VAE}

Todo:
GANs~\citep{Goodfellow2014, Goodfellow2017};
CycleGAN~\citep{Zhu2017CycleGAN}, DualGAN~\citep{Yi2017DualGAN};
See \citet{Arjovsky2017WGAN} for Wasserstein GAN (WGAN) as a stable GANs model. \citet{Gulrajani2017} proposed to improve stability of WGAN by penalizing the norm of the gradient of the discriminator with respect to its input,  instead of clipping weights as in \citet{Arjovsky2017WGAN}.  \citet{Mao2016} proposed Least Squares GANs (LSGANs), another stable model.

Connection with RL: \citet{Finn2016-Connection} established a connection between GANs, inverse RL, and energy-based models. \citet{Pfau2016} established the connection between GANs and actor-critic algorithms. \citet{Ho2016} and \citet{Li2017InfoGAIL} studied the connection between GANs and imitation learning.

autoencoder~\citep{Hinton2006} 

For disentangled factor learning,
\citet{Kulkarni2015DC-IGN} proposed DC-IGN, the Deep Convolution Inverse Graphics Network, which follows a semi-supervised way; 
and \citet{Chen2016InfoGAN} proposed InfoGAN, an information-theoretic extension to the Generative Adversarial Network, which follows an unsupervised way.
\citet{Zhou2015} showed that object detectors emerge from learning to recognize scenes,  without supervised labels for objects.

\citet{Higgins2017betaVAE} proposed $\beta$-VAE to automatically discover interpretable, disentangled, factorised, latent representations from raw images in an unsupervised way. 
The hyperparameter $\beta$ balances latent channel capacity and independence constraints with reconstruction accuracy.
When $\beta = 1$,  $\beta$-VAE is the same as the original VAEs.

\citet{Eslami2016} proposed the framework of Attend-Infer-Repeat for efficient inference in structured image models to reason about objects explicitly. The authors deployed a recurrent neural network to design an iterative process for inference, by attending to one object at a time, and for each image, learning an appropriate number of inference steps. The authors showed that, in an unsupervised way, the proposed approach can learn generative models to identify multiple objects for both 2D and 3D problems.

\citet{Zhang2018Interpretability} surveyed visual interpretability for deep learning.

\subsubsection{Recognition}

RL can improve efficiency for image classification by focusing only on salient parts.
For visual object localization and detection,  RL can improve efficiency over approaches  with exhaustive spatial hypothesis search and sliding windows, and strikes a balance between sampling more regions for better accuracy and stopping the search when sufficient confidence is obtained about the target's location.  

\citet{Mnih-attention-2014} introduced the recurrent attention model (RAM) to focus on selected sequence of regions or locations from an image or video for image classification and object detection. The authors used REINFORCE to train the model, to overcome the issue that the model is non-differentiable, and experimented on an image classification task and a dynamic visual control problem.  

\citet{Caicedo2015} proposed an active detection model for object localization with DQN, by deforming a bounding box with transformation actions to determine the most specific location for target objects. 
\citet{Jie2016} proposed a tree-structure RL approach to search for objects sequentially, considering both the current observation and previous search paths, by maximizing the long-term reward associated with localization accuracy over all objects with DQN. 
\citet{Mathe2016} proposed to use policy search for visual object detection.
\citet{Kong2017} deployed collaborative multi-agent RL with  inter-agent communication for joint object search.
\citet{Welleck2017} proposed a hierarchical visual architecture with an attention mechanism for multi-label image classification.
\citet{Rao2017} proposed an attention-aware deep RL method for video face recognition.

\citet{Krull2017} for 6D object pose estimation

\subsubsection{Motion Analysis}

In tracking, an agent needs to follow a moving object. 
\citet{Supancic2017} proposed online decision-making process for tracking, formulated it as a partially observable decision-making process (POMDP), and learned policies with deep RL algorithms,
to decide where to look for the object, when to reinitialize, and when to update the appearance model for the object,
where image frames may be ambiguous and computational budget may be constrained.
\citet{Yun2017} also studied visual tracking with deep RL.

\citet{Rhinehart2017}  proposed Discovering Agent Rewards for K-futures Online (DARKO) to model and forecast first-person camera wearer's long-term goals, together with states, transitions, and rewards from streaming data, with online inverse reinforcement learning.

\subsubsection{Scene Understanding}

\citet{Wu2017De-rendering} studied the problem of scene understanding, and attempted to obtain a compact, expressive, and interpretable representation to encode scene information like objects, their categories, poses, positions, etc, in a semi-supervised way. In contrast to encoder-decoder based neural architectures as in previous works, \citet{Wu2017De-rendering} proposed to replace the decoder with a deterministic rendering function, to map a structured and disentangled scene description, scene XML, to an image; consequently, the encoder transforms an image to the scene XML  by inverting the rendering operation, a.k.a., de-rendering. The authors deployed a variant of REINFORCE algorithm to overcome the non-differentiability issue of graphics rendering engines.

\citet{Wu2017De-animation} proposed a paradigm with three major components, a convolutional perception module, a physics engine, and a graphics engine, to understand physical scenes without human annotations. The perception module recovers a physical world representation by inverting the graphics engine, inferring the physical object state for each segment proposal in input and combining them. The generative physics and graphics engines then run forward with the world representation to reconstruct the visual data. 
The authors showed results on both neural, differentiable and more mature but non-differentiable physics engines.


There are recent works about physics learning, e.g.,~\cite{Agrawal2016Poke, Battaglia2016,  Denil2017, Watters2017, Wu2015Galileo}.

\subsubsection{Integration with NLP}

Some are integrating computer vision with natural language processing. \citet{Xu2015} integrated attention to image captioning, trained the hard version attention with REINFORCE, and showed the effectiveness of attention on Flickr8k, Flickr30k, and MS COCO datasets. 
\citet{Rennie2017} introduced self-critical sequence training, using the output of test-time inference algorithm as the baseline in REINFORCE to normalize the rewards it experiences, for image captioning. 
See also \citet{Liu2016}, \citet{Lu2016}, and \citet{Ren2017} for image captioning. \citet{Strub2017} proposed end-to-end optimization with deep RL for goal-driven and visually grounded dialogue systems for GuessWhat?! game. \citet{Das2017} proposed to learn cooperative Visual Dialog agents with deep RL. See also~\citet{Kottur2017}. See \citet{Pasunuru2017} for video captioning. See \citet{Liang2017X} for visual relationship and attribute detection.

\subsubsection{Visual Control}

Visual control is about deriving a policy from visual inputs, e.g., in games~\citep{Mnih-DQN-2015, Silver-AlphaGo-2016, Silver-AlphaGo-2017, Oh2015, Wu2017, Dosovitskiy2017, Lample2017, Jaderberg2017}, robotics~\citep{Finn-robotics-2016, Gupta2017CMP, Lee2017, Levine2016, Mirowski2017, Zhu2017}, and self-driving vehicles~\citep{Bojarski2016, Bojarski2017, Zhou2017Apple}. \footnote{Although we include visual control here, it is not very clear if we should categorize the following type of problems, e.g., DQN~\citep{Mnih-DQN-2015} and AlphaGo~\citep{Silver-AlphaGo-2016, Silver-AlphaGo-2017},  into computer vision: pixels (DQN) or problem setting (Go board status) as the input, some deep neural networks as the architecture, and an end-to-end gradient descent/ascent algorithm as the optimization method to find a policy, without any further knowledge of computer vision. Or we may see this as part of the synergy of computer vision and reinforcement learning.}



\subsection{Business Management}
\label{business}

Reinforcement learning has many applications in business management, like ads, recommendation, customer management, and marketing. 

\citet{Li2010} formulated personalized news articles recommendation as a contextual bandit problem, to learn an algorithm to select articles sequentially for users based on contextual information of the user and articles, such as historical activities of the user and descriptive information and categories of content, and to take user-click feedback to adapt article selection policy to maximize total user clicks in the long run. 

\citet{Theocharous2015} formulated a personalized ads recommendation systems as a RL problem to maximize life-time value (LTV) with theoretical guarantees. This is in contrast to a myopic solution with supervised learning or contextual bandit formulation, usually with the performance metric of click through rate (CTR). As the models are hard to learn, the authors deployed a model-free approach to compute a lower-bound on the expected return of a policy to address the off-policy evaluation problem, i.e., how to evaluate a RL policy without deployment. 

\citet{Li2016-hybrid} also attempted to maximize lifetime value of customers. \citet{Silver2013} proposed concurrent reinforcement learning for the customer interaction problem. See \citet{Sutton2018} for a  detailed and intuitive description of some topics discussed here under the section title of Personalized Web Services. 


\subsection{Finance}
\label{fin}

RL is a natural solution to some finance and economics problems~\citep{Hull06,Luenberger97}, like option pricing~\citep{Longstaff01, Tsitsiklis01, Li2009Option}, and multi-period portfolio optimization~\citep{Brandt05, Neuneier1997}, where value function based RL methods were used. \citet{Moody01} proposed to utilize policy search to learn to trade; \citet{Deng2016} extended it with deep neural networks. Deep (reinforcement) learning would provide better solutions in some issues in risk management~\citep{Hull06,Yu2009}. The market efficiency hypothesis is fundamental in finance. However, there are well-known behavioral biases in human decision-making under uncertainty, in particular, prospect theory~\citep{Prashanth2016}. A reconciliation is the adaptive markets hypothesis~\citep{Lo04}, which may be approached by reinforcement learning. 

It is nontrivial for finance and economics academia to accept blackbox methods like neural networks; \citet{Heaton2016} may be regarded as an exception. However, there is a lecture in AFA 2017 annual meeting: Machine Learning and Prediction in Economics and Finance. We may also be aware that  financial firms would probably hold state-of-the-art research/application results.

FinTech has been attracting attention, especially after the notion of big data. FinTech employs machine learning techniques to deal with issues like fraud detection~\citep{Phua2010},  consumer credit risk~\citep{Khandani2010}, etc.
 
 
\subsection{Healthcare}
\label{healthcare}

There are many opportunities and challenges in healthcare for machine learning \citep{Miotto2017, Saria2014}.
Personalized medicine is getting popular in healthcare. It  systematically optimizes the patient's health care, in particular, for chronic conditions and cancers using individual patient information, potentially from electronic health/medical record (EHR/EMR).  Dynamic treatment regimes (DTRs) or adaptive treatment strategies are sequential decision making problems. Some issues in DTRs are not in standard RL. \citet{Shortreed2011} tackled the missing data problem,  and designed methods to quantify the evidence of the learned optimal policy. \citet{Goldberg2012} proposed methods for censored data (patients may drop out during the trial) and flexible number of stages. See \citet{Chakraborty2014} for a recent survey, and \citet{Kosorok2015} for an edited book about recent progress in DTRs. Currently Q-learning is the RL method in DTRs. 
\citet{Ling2017} applied deep RL to the problem of inferring patient phenotypes.

Some recent workshops at the intersection of machine learning and healthcare are: NIPS 2016 Workshop on Machine Learning for Health (http://www.nipsml4hc.ws) and NIPS 2015 Workshop on Machine Learning in Healthcare (https://sites.google.com/site/nipsmlhc15/). See ICML 2017 Tutorial on Deep Learning for Health Care Applications: Challenges and Solutions (https://sites.google.com/view/icml2017-deep-health-tutorial/home). 
 
\subsection{Education}
\label{education}

\citet{Mandel2014}

\citet{Liu2014Education}
 
\subsection{Industry 4.0}
\label{Industry40}

The ear of Industry 4.0 is approaching, e.g., see~\citet{ODonovan2015}, and \citet{Preuveneers2017}. Reinforcement learning in particular, artificial intelligence in general, will be critical enabling techniques for many aspects of Industry 4.0, e.g., predictive maintenance, real-time diagnostics, and management of manufacturing activities and processes. Robots will prevail in Industry 4.0, and we discuss robotics in Section~\ref{robotics}. 

\citet{Liu2016CoRobot, Liu2017CoRobot} studied how to make robots and people to collaborate to achieve both flexibility and productivity in production lines. See a blog titled Towards Intelligent Industrial Co-robots, at 
http://bair.berkeley.edu/blog/2017/12/12/corobots/

\citet{Hein2017} designed a benchmark for the RL community to attempt to bridge the gap between academic research and real industrial problems. Its open source based on OpenAI Gym is available at https://github.com/siemens/industrialbenchmark.

\citet{Surana2016} proposed to apply  guided policy search~\citep{Levine2016} as discussed in Section~\ref{GPS} to optimize trajectory policy of cold spray nozzle dynamics, to handle complex trajectories traversing by robotic agents. The authors generated cold spray surface simulation profiles to train the model.


\subsection{Smart Grid}
\label{smartgrid}

A smart grid is a power grid utilizing modern information technologies to create an intelligent electricity delivery network for electricity generation, transmission, distribution, consumption, and control~\citep{Fang2012}. An important aspect is adaptive control~\citep{Anderson2011}. \citet{Glavic2017} reviewed application of RL for electric power system decision and control. Here we briefly discuss demand response~\citep{Wen2015, Ruelens2016}.

Demand response systems motivate users to dynamically adapt electrical demands in response to changes in grid signals, like electricity price, temperature, and weather, etc. With suitable electricity prices, load of peak consumption may be rescheduled/lessened, to improve efficiency, reduce costs, and reduce risks.  \citet{Wen2015} proposed to design a fully automated energy management system with model-free reinforcement learning, so that it doesn't need to specify a disutility function to model users' dissatisfaction with job rescheduling. The authors decomposed the RL formulation over devices, so that the computational complexity grows linearly with the number of devices, and conducted simulations using Q-learning.   \citet{Ruelens2016} tackled the demand response problem with batch RL. \citet{Wen2015} took the exogenous prices as states, and \citet{Ruelens2016} utilized the average as feature extractor to construct states.




\subsection{Intelligent Transportation Systems}
\label{ITS}

Intelligent transportation systems~\citep{Bazzan2014} apply advanced information technologies for tackling issues in transport networks, like congestion, safety, efficiency, etc., to make transport networks, vehicles and users smart. 

An important issue in intelligent transportation systems is adaptive traffic signal control. \citet{El-Tantawy2013} proposed to model the adaptive traffic signal control problem as a multiple player stochastic game, and  solve it with the approach of multi-agent RL~\citep{Shoham2007, Busoniu2008}. Multi-agent RL integrates single agent RL with game theory, facing challenges of stability, nonstationarity, and curse of dimensionality. \citet{El-Tantawy2013} approached the issue of coordination by considering agents at neighbouring intersections. The authors validated their proposed approach with simulations, and real traffic data from the City of Toronto. \citet{El-Tantawy2013} didn't explore function approximation. See also \citet{vanDerPol2017} for a recent work, and \citet{Mannion2016} for an experimental review, about applying RL to adaptive traffic signal control. 

Self-driving vehicle is also a topic of  intelligent transportation systems. See \citet{Bojarski2016}, \citet{Bojarski2017}, \citet{Zhou2017Apple}. 

See NIPS 2017, 2016 Workshop on Machine Learning for Intelligent Transportation Systems. 
Check for a special issue of IEEE Transactions on Neural Networks and Learning Systems on Deep Reinforcement Learning and Adaptive Dynamic Programming, tentative publication date December 2017.




\subsection{Computer Systems}
\label{systems}

Computer systems are indispensable in our daily life and work, e.g., mobile phones, computers, and cloud computing. Control and optimization problems abound in computer systems, e,g., \citet{Mestres2016} proposed knowledge-defined networks, \citet{Gavrilovska2013} reviewed learning and reasoning techniques in cognitive radio networks, and \citet{Haykin2005} discussed issues in cognitive radio, like channel state prediction and resource allocation.  We also note that Internet of Things (IoT)\citep{Xu2014} and wireless sensor networks~\citep{Alsheikh2014} play an important role in Industry 4.0 as discussed in Section~\ref{Industry40}, in Smart Grid as discussed in Section~\ref{smartgrid}, and in Intelligent Transportation Systems as discussed in Section~\ref{ITS}. 

\subsubsection{Resource Allocation}

\citet{Mao2016RM} studied resource management in systems and networking with deep RL. The authors proposed to tackle multi-resource cluster scheduling with policy gradient, in an online manner with dynamic job arrivals, optimizing various objectives like average job slowdown or completion time.
The authors validated their proposed approach with simulation.

\citet{Liu2017} proposed a hierarchical framework to tackle resource allocation and power management in cloud computing with deep RL. The authors decomposed the problem as a global tier for virtual machines resource allocation and a local tier for servers power management. The authors validated their proposed approach with actual Google cluster traces. Such hierarchical framework/decomposition approach was to reduce state/action space, and to enable distributed operation of power management. 

Google deployed machine learning for data centre power management, reducing energy consumption by 40\%, https://deepmind.com/blog/deepmind-ai-reduces-google-data-centre-cooling-bill-40/. Optimizing memory control is discussed in \citet{Sutton2018}.

\subsubsection{Performance Optimization}

\citet{Mirhoseini2017} proposed to optimize device placement for Tensorflow computational graphs with RL. 
The authors deployed a seuqence-to-sequence model to predict how to place subsets of operations in a Tensorflow graph on available devices, using the execution time of the predicted placement as reward signal for REINFORCE algorithm.
The proposed method found placements of Tensorflow operations on devices for Inception-V3, recurrent neural language model  and neural machine translation, yielding shorter execution time than those placements designed by human experts. 
Computation burden is one concern for a RL approach to search directly in the solution space of a combinatorial problem.
We discuss combinatorial optimization in Section~\ref{combinatorial}.

\subsubsection{Security \& Privacy}

adversarial attacks, e.g., \citet{Huang2017}, 

http://rll.berkeley.edu/adversarial/

\citet{Papernot2016cleverhans}

\citet{Abadi2016}

\citet{Balle2016}

\citet{Delle2014}

~\citet{Carlini2017SP}

defenders~\citet{Goodfellow2015, Carlini2017, Madry2017, Tramer2017}

\citet{Anderson2017} 
https://github.com/endgameinc/gym-malware

\citet{Evtimov2017}
See a blog titled Physical Adversarial Examples Against Deep Neural Networks at http://bair.berkeley.edu/blog/2017/12/30/yolo-attack/, which contains a brief survey of attack and defence algorithms.
Check ACM Conference on Computer and Communications Security (CCS 2016) tutorial on Adversarial Data Mining: Big Data Meets Cyber Security, https://www.sigsac.org/ccs/CCS2016/tutorials/index.html



\section{More Topics}
\label{todo}

We list more interesting and/or important topics we have not discussed in this overview as below, hoping it would provide pointers for those who may be interested in studying them further. Some topics/papers may not contain RL yet. However, we believe these are interesting and/or important directions for RL in the sense of either theory or application. It would be definitely more desirable if we could finish reviewing these, however, we leave it as future work.

\begin{itemize}
\item understanding deep learning, \citet{Daniely2016, LiJiwei2016Understanding, Karpathy2016, Kawaguchi2017, Koh2017, Neyshabur2017, Shalev-Shwartz2017, Shwartz-Ziv2017, Zhang2017}
\item interpretability, e.g.,  \citet{Al-Shedivat2017, Doshi-Velez2017, Harrison2017, Lei2016, Lipton2016Mythos, Miller2017, Ribeiro2016, Park2016}
\begin{itemize}
\item NIPS 2017 Interpretable Machine Learning Symposium
\item ICML 2017 Tutorial on Interpretable Machine Learning
\item NIPS 2016 Workshop on Interpretable ML for Complex Systems
\item ICML Workshop on Human Interpretability in Machine Learning 2017, 2016
\end{itemize}
\item usable machine learning, \citet{Bailis2017}
\item[] $\circ$ Cloud AutoML: Making AI accessible to every business, \newline https://www.blog.google/topics/google-cloud/cloud-automl-making-ai-accessible-every-business/
\item expressivity, \citet{Raghu2016}
\item testing, \citet{Pei2017} 
\item deep learning efficiency, e.g., \citet{Han2016}, \citet{Spring2017}, \citet{Sze2017}
\item deep learning compression

\item optimization, e.g., \citet{Wilson2017}, \citet{Czarnecki2017}
\item normalization, \citet{Klambauer2017}, \citet{vanHasselt2016-adaptive}
\item curriculum learning, \citet{Graves2017}, \citet{Held2017}, \citet{Matiisen2017}
\item professor forcing, \citet{Lamb2016}



\item new Q-value operators, \citet{Asadi2017}, \citet{Haarnoja2017}
\item large action space, e.g., \cite{Dulac-Arnold2016, He2016-actionspace}
\item Predictive State Representation, \citet{Downey2017}, \citet{Venkatraman2017}
\item safe RL, e.g., \citet{Berkenkamp2017}
\item agent modelling, e.g., \citet{Albrechta2018} 
\item semi-supervised learning, e.g., \citet{Audiffren2015, Cheng2016, Dai2017, Finn2017, Kingma2014,Papernot2017,YangZ2017, Zhu2009} 
\item neural episodic control, \citet{Pritzel2017}
\item continual learning, \citet{Chen2016Lifelong, Kirkpatrick2017, Lopez-Paz2017} 
\begin{itemize}
\item Satinder Singh, Steps towards continual learning, tutorial at Deep Learning and Reinforcement Learning Summer School 2017
\end{itemize}
\item symbolic learning, \citet{Evans2017,  Liang2017, Parisotto2017} 
\item pathNet, \citet{Fernando2017} 
\item evolution strategies, \citet{Petroski2017}, \citet{Salimans2017} , \citet{Lehman2017} 
\item capsules, \citet{Sabour2017}
\item DeepForest, \citet{Zhou2017, Feng2017}
\item deep probabilistic programming, \citet{Tran2017}
\item active learning, e.g., \citet{Fang2017} 
\item deep learning games, \citet{Schuurmans2016}
\item program learning, e.g., \citet{Balog2017, Cai2017, Denil2017agent, Parisotto2017, Reed2016}
\item relational reasoning, e.g., \citet{Santoro2017}, \citet{Watters2017}
\item proving, e.g., \citet{Loos2017, Rocktaschel2017}
\item education, e.g., \citet{}
\item music generation, e.g., \citet{Jaques-2017}
\item retrosynthesis, e.g., \citet{Segler2017}
\item quantum RL, e.g., \citet{Crawford2016}
\begin{itemize}
\item NIPS 2015 Workshop on Quantum Machine Learning
\end{itemize}
\end{itemize}



\section{Resources}
\label{resources}

We list a collection of deep RL resources including books, surveys, reports, online courses,  tutorials, conferences, journals and workshops, blogs, testbed, and open source algorithm implementations. This by no means is complete.

It is essential to have a good understanding of reinforcement learning, before having a good understanding of deep reinforcement learning. We recommend to start with the textbook by Sutton and Barto~\citep{Sutton2018}, the RL courses by Rich Sutton and by David Silver as the first two items in the Courses subsection below.
 
In the current information/social media age, we are overwhelmed by information, e.g., from Twitter, arXiv, Google+, etc. The skill to efficiently select the best information becomes essential. 
The Wild Week in AI (http://www.wildml.com) is an excellent series of weekly summary blogs. 
In an ear of AI, we expect to see an AI agent to do such tasks like intelligently searching and summarizing relevant news, blogs, research papers, etc.

\subsection{Books}

\begin{itemize}
\item the definitive and intuitive reinforcement learning book by  Richard S. Sutton and Andrew G. Barto~\citep{Sutton2018} 
\item deep learning books~\citep{Deng2014, Goodfellow2016}
\end{itemize}

\subsection{More Books}

\begin{itemize}
\item theoretical RL books~\citep{Bertsekas12, Bertsekas96,  Szepesvari2010}
\item an operations research oriented RL book~\citep{Powell11} 
\item an edited RL book~\citep{Wiering2012}
\item Markov decision processes~\citep{Puterman05}
\item machine learning~\citep{Bishop2011,  Hastie2009, Haykin2008, James2013, Kuhn2013, Murphy2012, Provost2013, Simeone2017, Zhou2016}
\item artificial intelligence~\citep{Russell2009}
\item natural language processing (NLP)~\citep{Deng2017, Goldberg2017, Jurafsky2017}
\item semi-supervised learning~\citep{Zhu2009}
\item game theory~\citep{Leyton-Brown2008}
\end{itemize}

\subsection{Surveys and Reports}

\begin{itemize}
\item reinforcement learning~\citep{Littman2015, Kaelbling1996, Geramifard2013, Grondman2012, Roijers2013}; deep reinforcement learning~\citep{Arulkumaran2017} \footnote{Our overview is much more comprehensive, and was online much earlier, than this brief survey.}
\item deep learning~\citep{LeCun2015, Schmidhuber2015-DL, Bengio2009, Wang2017Origin}
\item efficient processing of deep neural networks~\citep{Sze2017}
\item machine learning~\citep{Jordan2015}
\item practical machine learning advices~\citep{Domingos2012, Smith2017, Zinkevich2017}
\item natural language processing (NLP)~\citep{Hirschberg2015,Cho2015, Young2017}
\item spoken dialogue systems~\citep{Deng2013, Hinton2012, He2013, Young2013}
\item robotics~\citep{Kober2013}
\item transfer learning~\citep{Taylor09, Pan2010, Weiss2016}
\item Bayesian RL~\citep{Ghavamzadeh2015}
\item AI safety~\citep{Amodei2016, Garcia2015}
\item Monte Carlo tree search (MCTS)~\citep{Browne2012, Gelly12}
\end{itemize}

\subsection{Courses}
\begin{itemize}
\item Richard Sutton, Reinforcement Learning, 2016, slides, assignments, reading materials, etc. \newline http://www.incompleteideas.net/sutton/609\%20dropbox/
\item David Silver, Reinforcement Learning, 2015, slides (goo.gl/UqaxlO), video-lectures (goo.gl/7BVRkT)
\item Sergey Levine, John Schulman and Chelsea Finn, CS 294: Deep Reinforcement Learning, Spring 2017, http://rll.berkeley.edu/deeprlcourse/
\item Katerina Fragkiadaki, Ruslan Satakhutdinov, Deep Reinforcement Learning and Control, Spring 2017, https://katefvision.github.io
\item Emma Brunskill, CS234: Reinforcement Learning, http://web.stanford.edu/class/cs234/
\item Charles Isbell, Michael Littman and Pushkar Kolhe, Udacity: Machine Learning: Reinforcement Learning, goo.gl/eyvLfg
\item  David Donoho, Hatef Monajemi, and Vardan Papyan, Theories of Deep Learning (Stanford STATS 385), https://stats385.github.io
\item Nando de Freitas, Deep Learning Lectures, https://www.youtube.com/user/ProfNandoDF
\item Fei-Fei Li, Andrej Karpathy and Justin Johnson, CS231n: Convolutional Neural Networks for Visual Recognition, http://cs231n.stanford.edu
\item Richard Socher, CS224d: Deep Learning for Natural Language Processing, \newline http://cs224d.stanford.edu
\item Brendan Shillingford, Yannis Assael, Chris Dyer, Oxford Deep NLP 2017 course, \newline https://github.com/oxford-cs-deepnlp-2017
\item Pieter Abbeel, Advanced Robotics, Fall 2015, https://people.eecs.berkeley.edu/~pabbeel/cs287-fa15/
\item Emo Todorov, Intelligent control through learning and optimization, \newline http://homes.cs.washington.edu/~todorov/courses/amath579/index.html
\item Abdeslam Boularias, Robot Learning Seminar, http://www.abdeslam.net/robotlearningseminar
\item MIT 6.S094: Deep Learning for Self-Driving Cars, http://selfdrivingcars.mit.edu
\item Jeremy Howard, Practical Deep Learning For Coders, http://course.fast.ai
\item Andrew Ng, Deep Learning Specialization \newline https://www.coursera.org/specializations/deep-learning
\end{itemize}

\subsection{Tutorials}
\begin{itemize}
\item Rich Sutton, Introduction to Reinforcement Learning with Function Approximation, https://www.microsoft.com/en-us/research/video/tutorial-introduction-to-reinforcement-learning-with-function-approximation/ 
\item Deep Reinforcement Learning
\begin{itemize}
    \item David Silver, ICML 2016
    \item David Silver, 2nd Multidisciplinary Conference on Reinforcement Learning and Decision Making (RLDM), Edmonton, Alberta, Canada, 2015; \newline http://videolectures.net/rldm2015\_silver\_reinforcement\_learning/
    \item John Schulman,  Deep Learning School, 2016
    \item  Pieter Abbeel, Deep Learning Summer School, 2016; \newline http://videolectures.net/deeplearning2016\_abbeel\_deep\_reinforcement/
    \item Pieter Abbeel and John Schulman, Deep Reinforcement Learning Through Policy Optimization, NIPS 2016
    \item Sergey Levine and Chelsea Finn, Deep Reinforcement Learning, Decision Making, and Control, ICML 2017
\end{itemize}

\item John Schulman, The Nuts and Bolts of Deep Reinforcement Learning Research, Deep Reinforcement Learning Workshop, NIPS 2016 
\item Joelle Pineau, Introduction to Reinforcement Learning, Deep Learning Summer School, 2016; http://videolectures.net/deeplearning2016\_pineau\_reinforcement\_learning/
\item Andrew Ng, Nuts and Bolts of Building Applications using Deep Learning, NIPS 2016
\item Deep Learning Summer School, 2016, 2015
\item Deep Learning and Reinforcement Learning Summer Schools, 2017
\item Simons Institute Interactive Learning Workshop, 2017
\item Simons Institute Representation Learning Workshop, 2017
\item Simons Institute Computational Challenges in Machine Learning Workshop, 2017
\end{itemize}

\subsection{Conferences, Journals and Workshops}
\begin{itemize}
\item NIPS: Neural Information Processing Systems 
\item ICML: International Conference on Machine Learning
\item ICLR: International Conference on Learning Representation
\item RLDM: Multidisciplinary Conference on Reinforcement Learning and Decision Making
\item EWRL: European Workshop on Reinforcement Learning
\item AAAI, IJCAI, ACL, EMNLP, SIGDIAL, ICRA, IROS,  KDD, SIGIR, CVPR, etc.
\item Nature Machine Intelligence, Science Robotics, JMLR, MLJ, AIJ, JAIR, PAMI, etc
\item Nature May 2015, Science July 2015, survey papers on machine learning/AI
\item Science, July 7, 2017 issue, The Cyberscientist, a special issue about AI
\item Deep Reinforcement Learning Workshop, NIPS 2016, 2015; IJCAI 2016
\item Deep Learning Workshop, ICML 2016
\item http://distill.pub
\end{itemize}

\subsection{Blogs}
\begin{itemize}
\item Deepmind Blog, https://deepmind.com/blog/
\item[] $\circ$
\item[] $\circ$
\item Google Research Blog, https://research.googleblog.com, goo.gl/ok88b7
\item[] $\circ$ The Google Brain Team --- Looking Back on 2017, goo.gl/1G7jnb, goo.gl/uCWDLr
\item[] $\circ$ The Google Brain Team --- Looking Back on 2016,  
\item Berkeley AI Research Blog, http://bair.berkeley.edu/blog/
\item OpenAI Blog, https://blog.openai.com

\item Marc Bellemare, Classic and modern reinforcement learning, \newline http://www.marcgbellemare.info/blog/

\item Denny Britz, The Wild Week in AI, a weekly AI \& deep learning newsletter,  www.wildml.com, esp. goo.gl/MyrwDC
\item Andrej Karpathy, karpathy.github.io, esp. goo.gl/1hkKrb
\item Junling Hu, Reinforcement learning explained - learning to act based on long-term payoffs \newline https://www.oreilly.com/ideas/reinforcement-learning-explained
\item Li Deng, How deep reinforcement learning can help chatbots\newline https://venturebeat.com/2016/08/01/how-deep-reinforcement-learning-can-help-chatbots/
\item Reinforcement Learning, https://www.technologyreview.com/s/603501/10-breakthrough-technologies-2017-reinforcement-learning/
\item Deep Learning, https://www.technologyreview.com/s/513696/deep-learning/
\end{itemize}

\subsection{Testbeds}
\label{testbed}


\begin{itemize}

\item The Arcade Learning Environment (ALE) \citep{Bellemare2013, Machado2017ALE} is a framework composed of Atari 2600 games to develop and evaluate AI agents.

\item Ray RLlib: A Composable and Scalable Reinforcement Learning Library~\citep{Liang2017RayRLLib},  http://ray.readthedocs.io/en/latest/rllib.html

\item OpenAI Gym (https://gym.openai.com) is a toolkit for the development of RL algorithms, consisting of environments, e.g., Atari games and simulated robots, and a site for the comparison and reproduction of results. 

\item OpenAI Universe (https://universe.openai.com) is used to turn any program into a Gym environment. Universe has already integrated many environments, including Atari games, flash games, browser tasks like Mini World of Bits and real-world browser tasks. Recently, GTA V was added to Universe for self-driving vehicle simulation.

\item DeepMind Control Suite,~\citet{Tassa2018}

\item DeepMind released a first-person 3D game platform DeepMind Lab~\citep{Beattie2016}.  Deepmind and Blizzard will collaborate to release the Starcraft II AI research environment (goo.gl/Ptiwfg).

\item Psychlab: A Psychology Laboratory for Deep Reinforcement Learning Agents~\citep{Leibo2018Psychlab}

\item FAIR TorchCraft~\citep{Synnaeve2016} is a library for Real-Time Strategy (RTS) games such as StarCraft: Brood War.

\item Deepmind PySC2 - StarCraft II Learning Environment, https://github.com/deepmind/pysc2

\item David Churchill, CommandCenter: StarCraft 2 AI Bot, \newline https://github.com/davechurchill/commandcenter

\item ParlAI is a framework for dialogue research, implemented in Python, open-sourced by Facebook. https://github.com/facebookresearch/ParlAI

\item ELF, an extensive, lightweight and flexible platform for RL research~\citep{Tian2017ELF}

\item Project Malmo (https://github.com/Microsoft/malmo), from Microsoft, is an AI research and experimentation platform built on top of Minecraft. 

\item Twitter open-sourced torch-twrl, a framework for RL development.

\item ViZDoom is a Doom-based AI research platform for visual RL~\citep{Kempka2016}.

\item Baidu Apollo Project, self-driving open-source, http://apollo.auto

\item TORCS is a car racing simulator \citep{TORCS}. 

\item MuJoCo, Multi-Joint dynamics with Contact, is a physics engine, http://www.mujoco.org.


\item \citet{Nogueira2016} presented WebNav Challenge for Wikipedia links navigation.

\item RLGlue~\citep{Tanner2009} is a language-independent software for RL experiments. It may need extensions to accommodate progress in deep learning.

\item RLPy~\citep{Geramifard2015} is a value-function-based reinforcement learning framework for education and research.

\end{itemize}




\subsection{Algorithm Implementations}


We collect implementations of algorithms, either classical ones as in a textbook like ~\citet{Sutton2018} or in recent papers.

\begin{itemize}
\item Shangtong Zhang, Python code to accompany Sutton \& Barto's RL  book and David Silver's RL course, https://github.com/ShangtongZhang/reinforcement-learning-an-introduction 
\item Learning Reinforcement Learning (with Code, Exercises and Solutions), \newline http://www.wildml.com/2016/10/learning-reinforcement-learning/
\item OpenAI Baselines: high-quality implementations of reinforcement learning algorithms, https://github.com/openai/baselines

\item TensorFlow implementation of Deep Reinforcement Learning papers, \newline https://github.com/carpedm20/deep-rl-tensorflow 
\item Deep reinforcement learning for Keras, https://github.com/matthiasplappert/keras-rl

\item Code Implementations for NIPS 2016 papers, http://bit.ly/2hSaOyx 

\item Benchmark results of various policy optimization algorithms~\citep{Duan2016}, \newline https://github.com/rllab/rllab

\item  Tensor2Tensor (T2T)~\citep{Vaswani2017, Kaiser2017OneModel, Kaiser2017}


\item DQN~\citep{Mnih-DQN-2015}, https://sites.google.com/a/deepmind.com/dqn/
\item Tensorflow implementation of DQN~\citep{Mnih-DQN-2015}, \newline https://github.com/devsisters/DQN-tensorflow
\item Deep Q Learning with Keras and Gym, https://keon.io/deep-q-learning/

\item Deep Exploration via Bootstrapped DQN~\citep{Osband2016}, a Torch implementation, https://github.com/iassael/torch-bootstrapped-dqn 

\item DarkForest, the Facebook Go engine (Github), https://github.com/facebookresearch/darkforestGo 
\item Using Keras and Deep Q-Network to Play FlappyBird, \newline https://yanpanlau.github.io/2016/07/10/FlappyBird-Keras.html 
\item Deep Deterministic Policy Gradients~\citep{Lillicrap2016} in TensorFlow, \newline http://pemami4911.github.io/blog/2016/08/21/ddpg-rl.html
\item Deep Deterministic Policy Gradient~\citep{Lillicrap2016} to play TORCS, \newline https://yanpanlau.github.io/2016/10/11/Torcs-Keras.html 
\item Reinforcement learning with unsupervised auxiliary tasks~\citep{Jaderberg2017}, \newline https://github.com/miyosuda/unreal
\item Learning to communicate with deep multi-agent reinforcement learning, \newline https://github.com/iassael/learning-to-communicate 
\item Deep Reinforcement Learning: Playing a Racing Game - Byte Tank,  http://bit.ly/2pVIP4i 
\item Differentiable Neural Computer (DNC)~\citep{Grave-DNC-2016}, \newline https://github.com/deepmind/dnc

\item Playing FPS Games with Deep Reinforcement Learning~\citep{Lample2017}, \newline https://github.com/glample/Arnold

\item Learning to Learn~\citep{Reed2016} in TensorFlow, \newline https://github.com/deepmind/learning-to-learn 
\item Value Iteration Networks~\citep{Tamar2016} in Tensorflow, \newline https://github.com/TheAbhiKumar/tensorflow-value-iteration-networks 
\item Tensorflow implementation of the Predictron~\citep{Silver-Predictron-2016}, \newline https://github.com/zhongwen/predictron 
\item Meta Reinforcement Learning~\citep{Wang2016LearnRL} in Tensorflow, \newline https://github.com/awjuliani/Meta-RL 
\item Generative adversarial imitation learning~\citep{Ho2016}, containing an implementation of Trust Region Policy Optimization (TRPO)~\citep{Schulman2015}, https://github.com/openai/imitation 
\item Starter code for evolution strategies~\citep{Salimans2017}, \newline https://github.com/openai/evolution-strategies-starter 
\item Transfer learning~\citep{Long2015, Long2016}, https://github.com/thuml/transfer-caffe
\item DeepForest~\citep{Zhou2017}, http://lamda.nju.edu.cn/files/gcforest.zip



\end{itemize}

\section{Brief Summary}
\label{summary}



We list some RL issues and corresponding proposed approaches covered in this overview, as well as some classical work.
One direction of future work is to further refine this section, especially for issues and solutions in applications. 

\begin{itemize}
\item issue: prediction, policy evaluation
\item[] proposed approaches:
\begin{itemize}
  \item temporal difference (TD) learning~\citep{Sutton1988}
\end{itemize}

\item issue: control, finding optimal policy (classical work)
\item[] proposed approaches:
\begin{itemize}
  \item Q-learning~\citep{Watkins1992}
  \item policy gradient~\citep{Williams1992}
  \begin{itemize}
    \item[]  $\diamond$ reduce variance of gradient estimate: baseline, advantage function~\citep{Williams1992, Sutton2000}
  \end{itemize}
  \item actor-critic~\citep{Barto1983}
  \item SARSA~\citep{Sutton2018}
\end{itemize}

\item issue: the deadly triad: instability and divergence when combining off-policy, function approximation, and bootstrapping
\item[] proposed approaches:
\begin{itemize}
  \item DQN with experience replay~\citep{Lin1992} and target network~\citep{Mnih-DQN-2015}
  \begin{itemize}
    \item[]  $\diamond$ overestimate problem in Q-learning: double DQN~\citep{vanHasselt2016} 
    \item[]  $\diamond$ prioritized experience replay~\citep{Schaul2016} 
    \item[]  $\diamond$ better exploration strategy~\citep{Osband2016}
    \item[]  $\diamond$ optimality tightening to accelerate DQN~\citep{He2017}
    \item[]  $\diamond$ reduce variability and instability with averaged-DQN~\citep{Anschel2017}
  \end{itemize}
  \item dueling architecture~\citep{Wang-Dueling-2016} 
  \item asynchronous methods~\citep{Mnih-A3C-2016}
  \item trust region policy optimization~\citep{Schulman2015}
  \item distributed proximal policy optimization~\citep{Heess2017DPPO}
  \item combine policy gradient and Q-learning~\citep{ODonoghue2017, Nachum2017Gap, Gu2017QProp, Schulman2017}
  \item GTD~\citep{Sutton2009GTD-ICML, Sutton2009GTD-NIPS, Mahmood2014}
  \item Emphatic-TD~\citep{Sutton2016}
\end{itemize}

\item issue: train perception and control jointly end-to-end
\item[] proposed approaches:
\begin{itemize}
  \item guided policy search~\citep{Levine2016}
\end{itemize}

\item issue: data/sample efficiency
\item[] proposed approaches:
\begin{itemize}
  \item Q-learning, actor-critic
  \item model-based policy search, e.g., PILCO~\citet{Deisenroth2011}
  \item actor-critic with experience replay~\citep{Wang2017}
  \item PGQ, policy gradient and Q-learning~\citep{ODonoghue2017}
  \item Q-Prop, policy gradient with off-policy critic~\citep{Gu2017QProp}
  \item return-based off-policy control, Retrace~\citep{Remi2016}, Reactor~\citep{Gruslys2017}
  \item learning to learn, e.g., \citet{Duan2016RL2,Wang2016LearnRL, Lake2015}
\end{itemize}

\item issue: reward function not available 
\item[] proposed approaches:
\begin{itemize}
  \item imitation learning
  \item inverse RL~\citep{Ng2000}
  \item learn from demonstration~\citep{Hester2018}
  \item imitation learning with GANs~\citep{Ho2016, Stadie2017} 
  \item train dialogue policy jointly with reward model~\citep{Su2016}
\end{itemize}

\item issue: exploration-exploitation tradeoff
\item[] proposed approaches:
\begin{itemize}
  \item unify count-based exploration and intrinsic motivation~\citep{Bellemare2016}
  \item under-appreciated reward exploration~\citep{Nachum2017}
  \item deep exploration via bootstrapped DQN~\citep{Osband2016} 
  \item variational information maximizing exploration~\citep{Houthooft2016}
\end{itemize}

\item issue: model-based learning
\item[] proposed approaches:
\begin{itemize}
  \item Dyna-Q~\citep{Sutton1990}
  \item combine model-free and model-based RL~\citep{Chebotar2017}
\end{itemize}

\item issue: model-free planning
\item[] proposed approaches:
\begin{itemize}
  \item value iteration networks~\citep{Tamar2016}
  \item predictron~\citep{Silver-Predictron-2016}
\end{itemize}

\item issue: focus on salient parts
\item[] proposed approaches: attention
\begin{itemize}
  \item object detection~\citep{Mnih-attention-2014}
  \item neural machine translation~\citep{Bahdanau2015}
  \item image captioning~\citep{Xu2015}
  \item replace CNN and RNN with attention in sequence modelling~\citep{Vaswani2017}
\end{itemize}

\item issue: data storage over long time, separating from computation
\item[] proposed approaches: memory
\begin{itemize}
  \item differentiable neural computer (DNC) with external memory~\citep{Grave-DNC-2016}   
\end{itemize}

\item issue: benefit from non-reward training signals in environments 
\item[] proposed approaches: unsupervised Learning
\begin{itemize}
  \item Horde~\citep{Sutton2011}
  \item unsupervised reinforcement and auxiliary learning~\citep{Jaderberg2017}
  \item learn to navigate with unsupervised auxiliary learning~\citep{Mirowski2017}
  \item generative adversarial networks (GANs)~\citep{Goodfellow2014}
\end{itemize}

\item issue: learn knowledge from different domains
\item[] proposed approaches: transfer Learning~\citep{Taylor09, Pan2010, Weiss2016}
\begin{itemize}
  \item learn invariant features to transfer skills~\citep{Gupta2017}
\end{itemize}

\item issue: benefit from both labelled and unlabelled data
\item[] proposed approaches: semi-supervised learning~\citep{Zhu2009}
\begin{itemize}
  \item learn with MDPs both with and without reward functions~\citep{Finn2017}
  \item learn with expert's trajectories and those may not from experts~\citep{Audiffren2015}
\end{itemize}

\item issue:  learn, plan, and represent knowledge with spatio-temporal abstraction at multiple levels
\item[] proposed approaches: hierarchical RL~\citep{Barto2003}
\begin{itemize}
  \item options~\citep{Sutton1999}, MAXQ~\citep{Dietterich2000}
  \item strategic attentive writer to learn macro-actions~\citep{Vezhnevets2016}
  \item integrate temporal abstraction with intrinsic motivation~\citep{Kulkarni2016}
  \item stochastic neural networks for hierarchical RL~\citep{Florensa2017}
  \item lifelong learning with hierarchical RL~\citep{Tessler2017} 
\end{itemize}

\item issue: adapt rapidly to new tasks
\item[] proposed approaches: learning to learn
\begin{itemize}
  \item learn to optimize~\citep{Li2017}
  \item learn a flexible RNN model to handle a family of RL tasks~\citep{Duan2016RL2,Wang2016LearnRL}
  \item one/few/zero-shot learning~\citep{Duan2017OneShot, Johnson2016, Kaiser2017, Koch2015, Lake2015,  Ravi2017, Vinyals2016}
\end{itemize}

\item issue: gigantic search space
\item[] proposed approaches:
\begin{itemize}
  \item integrate supervised learning, reinforcement learning, and Monte-Carlo tree search as in AlphaGo~\citep{Silver-AlphaGo-2016}
\end{itemize}

\item issue: neural networks architecture design 
\item[] proposed approaches:
\begin{itemize}
  \item neural architecture search \citep{Bello2017, Baker2017, Zoph2017}
  \item new architectures, e.g., \citet{Kaiser2017OneModel}, \citet{Silver-Predictron-2016}, \citet{Tamar2016}, \citet{Vaswani2017}, \citet{Wang-Dueling-2016}
\end{itemize}

\end{itemize}

\section{Discussions}
\label{discussion}

It is both the best and the worst of times for the field of deep RL, for the same reason: it has been growing so fast and so enormously. We have been witnessing breakthroughs, exciting new methods and applications, and we expect to see much more and much faster. As a consequence, this overview is incomplete, in the sense of both depth and width. However, we attempt to summarize important achievements and discuss potential directions and applications in this amazing field. 

In this overview, we summarize six core elements -- 
value function,
policy,
reward,
model and planning,
exploration,
and knowledge; 
six important mechanisms -- 
attention and memory, 
unsupervised learning,
transfer learning,
multi-agent RL,
hierarchical RL, 
and learning to learn; 
and twelve applications --
games,
robotics,
natural language processing, 
computer vision,
business management,
finance,
healthcare, 
education,
Industry 4.0,
smart grid,
intelligent transportation systems,
and computer systems.
We also discuss background of machine learning, deep learning, and reinforcement learning, 
and list a collection of RL resources.

We have seen breakthroughs about deep RL, including deep Q-network~\citep{Mnih-DQN-2015} and AlphaGo~\citep{Silver-AlphaGo-2016}. There have been many extensions to, improvements  for and applications of deep Q-network~\citep{Mnih-DQN-2015}. 


Novel architectures and applications using deep RL were recognized in top tier conferences as best papers in 2016: 
dueling network architectures~\citep{Wang-Dueling-2016} at ICML, 
spoken dialogue systems~\citep{Su2016} at ACL (student), 
information extraction~\citep{Narasimhan2016} at EMNLP, and,    
value iteration networks~\citep{Tamar2016} at NIPS.
\citet{Gelly2007} was the recipient of Test of Time Award at ICML 2017.
In 2017, the following were recognized as best papers,
\citet{Kottur2017}  at EMNLP (short), and, 
\citet{Bacon2017}  at AAAI (student). 
Exciting achievements abound: 
differentiable neural computer~\citep{Grave-DNC-2016}, 
unsupervised reinforcement and auxiliary learning~\citep{Jaderberg2017}, 
asynchronous methods~\citep{Mnih-A3C-2016}, 
dual learning for machine translation~\citep{He2016}, 
guided policy search~\citep{Levine2016}, 
generative adversarial imitation learning~\citep{Ho2016}, 
and neural architecture design~\citep{Zoph2017}, etc. 
Creativity would push the frontiers of deep RL further with respect to core elements, mechanisms, and applications.

State of the Art Control of Atari Games Using Shallow RL was accepted at AAMAS. It was also nominated for the Best Paper Award

Value function is central to reinforcement learning, e.g., in deep Q-network and its many extentions. Policy optimization approaches have been gaining traction, in many, diverse applications, e.g., robotics, neural architecture design, spoken dialogue systems, machine translation, attention, and learning to learn, and this list is boundless. New learning mechanisms have emerged, e.g., using transfer/unsupervised/semi-supervised learning to improve the quality and speed of learning, and more new mechanisms will be emerging. This is the renaissance of reinforcement learning~\citep{Krakovsky2016}. In fact, reinforcement learning and deep learning have been making steady progress even during the last AI winter.  




A popular criticism about deep learning is that it is a blackbox, or even the "alchemy" as a comment during NIPS 2017 Test of Time Award~\citep{Rahimi2007} speech,  so it is not clear how it works. This should not be the reason not to accept deep learning; rather, having a better understanding of how deep learning works is helpful for deep learning and general machine learning community. There are works in this direction as well as for interpretability of  deep learning as we list in Section~\ref{todo}.

It is essential to consider issues of learning models, like stability, convergence, accuracy, data efficiency, scalability, speed, simplicity, interpretability, robustness, and safety, etc. 
It is important to investigate comments/criticisms, e.g., from conginitive science, like intuitive physics, intuitive psychology, causal model, compositionality, learning to learn, and act in real time~\citep{Lake2016}, for stronger AI.  
It is interesting to check Deepmind's commentary~\citep{Botvinick2017} about one additional ingredient, autonomy, 
so that agents can build and exploit their own internal models, with minimal human manual engineering, 
 and investigate the connection between neuroscience and RL/AI~\citep{Hassabis2017}.
See also Peter Norvig's perspective at http://bit.ly/2qpehcd. 
See~\citet{Stoica2017} for systems challenges for AI.

Nature in May 2015 and Science in July 2015 featured survey papers on machine learning/AI.  
Science Robotics launched in 2016. 
Science has a special issue on July 7, 2017 about AI on The Cyberscientist. 
Nature Machine Intelligence will launch in January 2019.
The coverage of AI by premier journals like Nature and Science and the launch of Science Robotics  and Nature Machine Intelligence illustrate the apparent importance of AI. 
It is interesting to mention that NIPS 2017 main conference was sold out only two weeks after opening for registration.  

It is worthwhile to envision deep RL considering  perspectives of government, academia and industry on AI, e.g., Artificial Intelligence, Automation, and the economy, Executive Office of the President, USA;  Artificial Intelligence and Life in 2030 - One Hundred Year Study on Artificial Intelligence: Report of the 2015-2016 Study Panel, Stanford University~\citep{Stone2016}; and AI, Machine Learning and Data Fuel the Future of Productivity by The Goldman Sachs Group, Inc., etc. See also the recent AI Frontiers Conference, https://www.aifrontiers.com.

Deep learning was among MIT Technology Review 10 Breakthrough Technologies in 2013. We have been witnessing the dramatic development of deep learning in both academia and industry in the last few years. Reinforcement learning was among MIT Technology Review 10 Breakthrough Technologies in 2017. Deep learning has made many achievements, has "conquered" speech recognition,  computer vision, and now NLP, is more mature and well-accepted, and has been validated by products and market. In contrast, RL has lots of (potential, promising) applications, yet few products so far, may still need better algorithms, may still need  products and market validation. However, it is probably the right time to nurture, educate and lead the market. We will see both deep learning and reinforcement learning prospering in the coming years and beyond. Prediction is very difficult, especially about the future. However, we expect that 2018 for reinforcement learning would be 2010 for deep learning.


Deep learning, in this third wave of AI, will have deeper influences, as we have already seen from its many achievements. Reinforcement learning, as a more general learning and decision making paradigm, will deeply influence deep learning, machine learning, and artificial intelligence in general.  Deepmind, conducting leading research in deep reinforcement learning, recently opened its first ever international AI research office in Alberta, Canada, co-locating with the major research center for reinforcement learning led by Rich Sutton. It is interesting to mention that when Professor Rich Sutton started working in the University of Alberta in 2003, he named his lab RLAI: Reinforcement Learning and Artificial Intelligence.


\section*{Ackowledgement}

We appreciate comments from Baochun Bai, Kan Deng, Hai Fang, Hua He, Junling Hu, Ruitong Huang, Aravind Lakshminarayanan, Jinke Li, Lihong Li,  Bhairav Mehta, Dale Schuurmans, David Silver, Rich Sutton, Csaba Szepesv{\'a}ri, Arash Tavakoli, Cameron Upright, Yi Wan, Qing Yu, Yaoliang Yu, attendants of various seminars and webinars, in particular, a seminar at MIT on AlphaGo: Key Techniques and Applications, and an AI seminar at the University of Alberta on Deep Reinforcement Learning: An Overview. Any  remaining issues and errors are our own. 




\bibliography{DeepRL}

\begin{thebibliography}{}

\bibitem[\protect\astroncite{}{}]{}


\bibitem[\protect\astroncite{Abadi et~al.}{2016}]{Abadi2016}
Abadi, M., Chu, A., Goodfellow, I., McMahan, H.~B., Mironov, I., Talwar, K.,
  and Zhang, L. (2016).
\newblock Deep learning with differential privacy.
\newblock In {\em ACM Conference on Computer and Communications Security (ACM
  CCS)}.

\bibitem[\protect\astroncite{Abbeel and Ng}{2004}]{Abbeel2004}
Abbeel, P. and Ng, A.~Y. (2004).
\newblock Apprenticeship learning via inverse reinforcement learning.
\newblock In {\em the International Conference on Machine Learning (ICML)}.

\bibitem[\protect\astroncite{Agrawal et~al.}{2016}]{Agrawal2016Poke}
Agrawal, P., Nair, A., Abbeel, P., Malik, J., and Levine, S. (2016).
\newblock Learning to poke by poking: Experiential learning of intuitive
  physics.
\newblock In {\em the Annual Conference on Neural Information Processing
  Systems (NIPS)}.

\bibitem[\protect\astroncite{{Al-Shedivat} et~al.}{2017a}]{Al-Shedivat2017meta}
{Al-Shedivat}, M., {Bansal}, T., {Burda}, Y., {Sutskever}, I., {Mordatch}, I.,
  and {Abbeel}, P. (2017a).
\newblock {Continuous Adaptation via Meta-Learning in Nonstationary and
  Competitive Environments}.
\newblock {\em ArXiv e-prints}.

\bibitem[\protect\astroncite{{Al-Shedivat} et~al.}{2017b}]{Al-Shedivat2017}
{Al-Shedivat}, M., {Dubey}, A., and {Xing}, E.~P. (2017b).
\newblock {Contextual Explanation Networks}.
\newblock {\em ArXiv e-prints}.

\bibitem[\protect\astroncite{Albrechta and Stone}{2018}]{Albrechta2018}
Albrechta, S.~V. and Stone, P. (2018).
\newblock Autonomous agents modelling other agents: A comprehensive survey and
  open problems.
\newblock {\em Artificial Intelligence}.

\bibitem[\protect\astroncite{Alsheikh et~al.}{2014}]{Alsheikh2014}
Alsheikh, M.~A., Lin, S., Niyato, D., and Tan, H.-P. (2014).
\newblock Machine learning in wireless sensor networks: Algorithms, strategies,
  and applications.
\newblock {\em IEEE Communications Surveys \& Tutorials}, 16(4):1996--2018.

\bibitem[\protect\astroncite{Amin et~al.}{2017}]{Amin2017}
Amin, K., Jiang, N., and Singh, S. (2017).
\newblock Repeated inverse reinforcement learning.
\newblock In {\em the Annual Conference on Neural Information Processing
  Systems (NIPS)}.

\bibitem[\protect\astroncite{Amodei et~al.}{2016}]{Amodei2016}
Amodei, D., Olah, C., Steinhardt, J., Christiano, P., Schulman, J., and
  Man{\'e}, D. (2016).
\newblock {Concrete Problems in AI Safety}.
\newblock {\em ArXiv e-prints}.

\bibitem[\protect\astroncite{Anderson et~al.}{2017}]{Anderson2017}
Anderson, H.~S., Kharkar, A., Filar, B., and Roth, P. (2017).
\newblock Evading machine learning malware detection.
\newblock In {\em Black Hat USA}.

\bibitem[\protect\astroncite{Anderson et~al.}{2011}]{Anderson2011}
Anderson, R.~N., Boulanger, A., Powell, W.~B., and Scott, W. (2011).
\newblock Adaptive stochastic control for the smart grid.
\newblock {\em Proceedings of the IEEE}, 99(6):1098--1115.

\bibitem[\protect\astroncite{Andreas et~al.}{2017}]{Andreas2017}
Andreas, J., Klein, D., and Levine, S. (2017).
\newblock Modular multitask reinforcement learning with policy sketches.
\newblock In {\em the International Conference on Machine Learning (ICML)}.

\bibitem[\protect\astroncite{Andrychowicz et~al.}{2016}]{Andrychowicz2016}
Andrychowicz, M., Denil, M., Colmenarejo, S.~G., Hoffman, M.~W., Pfau, D.,
  Schaul, T., Shillingford, B., and de~Freitas, N. (2016).
\newblock Learning to learn by gradient descent by gradient descent.
\newblock In {\em the Annual Conference on Neural Information Processing
  Systems (NIPS)}.

\bibitem[\protect\astroncite{{Andrychowicz} et~al.}{2017}]{Andrychowicz2017}
{Andrychowicz}, M., {Wolski}, F., {Ray}, A., {Schneider}, J., {Fong}, R.,
  {Welinder}, P., {McGrew}, B., {Tobin}, J., {Abbeel}, P., and {Zaremba}, W.
  (2017).
\newblock Hindsight experience replay.
\newblock In {\em the Annual Conference on Neural Information Processing
  Systems (NIPS)}.

\bibitem[\protect\astroncite{Anschel et~al.}{2017}]{Anschel2017}
Anschel, O., Baram, N., and Shimkin, N. (2017).
\newblock {Averaged-DQN}: Variance reduction and stabilization for deep
  reinforcement learning.
\newblock In {\em the International Conference on Machine Learning (ICML)}.

\bibitem[\protect\astroncite{Argall et~al.}{2009}]{Argall2009}
Argall, B.~D., Chernova, S., Veloso, M., and Browning, B. (2009).
\newblock A survey of robot learning from demonstration.
\newblock {\em Robotics and Autonomous Systems}, 57(5):469--483.

\bibitem[\protect\astroncite{{Arjovsky} et~al.}{2017}]{Arjovsky2017WGAN}
{Arjovsky}, M., {Chintala}, S., and {Bottou}, L. (2017).
\newblock {Wasserstein GAN}.
\newblock {\em ArXiv e-prints}.

\bibitem[\protect\astroncite{{Artetxe} et~al.}{2017}]{Artetxe2017}
{Artetxe}, M., {Labaka}, G., {Agirre}, E., and {Cho}, K. (2017).
\newblock {Unsupervised Neural Machine Translation}.
\newblock {\em ArXiv e-prints}.

\bibitem[\protect\astroncite{{Arulkumaran} et~al.}{2017}]{Arulkumaran2017}
{Arulkumaran}, K., {Deisenroth}, M.~P., {Brundage}, M., and {Bharath}, A.~A.
  (2017).
\newblock {A Brief Survey of Deep Reinforcement Learning}.
\newblock {\em ArXiv e-prints}.

\bibitem[\protect\astroncite{Asri et~al.}{2016}]{ElAsri2016}
Asri, L.~E., He, J., and Suleman, K. (2016).
\newblock A sequence-to-sequence model for user simulation in spoken dialogue
  systems.
\newblock In {\em Annual Meeting of the International Speech Communication
  Association (INTERSPEECH)}.

\bibitem[\protect\astroncite{Audiffren et~al.}{2015}]{Audiffren2015}
Audiffren, J., Valko, M., Lazaric, A., and Ghavamzadeh, M. (2015).
\newblock Maximum entropy semi-supervised inverse reinforcement learning.
\newblock In {\em the International Joint Conference on Artificial Intelligence
  (IJCAI)}.

\bibitem[\protect\astroncite{Azar et~al.}{2017}]{Azar2017}
Azar, M.~G., Osband, I., and Munos, R. (2017).
\newblock Minimax regret bounds for reinforcement learning.
\newblock In {\em the International Conference on Machine Learning (ICML)}.

\bibitem[\protect\astroncite{Ba et~al.}{2016}]{Ba2016}
Ba, J., Hinton, G.~E., Mnih, V., Leibo, J.~Z., and Ionescu, C. (2016).
\newblock Using fast weights to attend to the recent past.
\newblock In {\em the Annual Conference on Neural Information Processing
  Systems (NIPS)}.

\bibitem[\protect\astroncite{Ba et~al.}{2014}]{Ba2014}
Ba, J., Mnih, V., and Kavukcuoglu, K. (2014).
\newblock Multiple object recognition with visual attention.
\newblock In {\em the International Conference on Learning Representations
  (ICLR)}.

\bibitem[\protect\astroncite{Babaeizadeh et~al.}{2017}]{Babaeizadeh2017}
Babaeizadeh, M., Frosio, I., Tyree, S., Clemons, J., and Kautz, J. (2017).
\newblock Reinforcement learning through asynchronous advantage actor-critic on
  a gpu.
\newblock In {\em the International Conference on Learning Representations
  (ICLR)}.

\bibitem[\protect\astroncite{Bacon et~al.}{2017}]{Bacon2017}
Bacon, P.-L., Harb, J., and Precup, D. (2017).
\newblock The option-critic architecture.
\newblock In {\em the AAAI Conference on Artificial Intelligence (AAAI)}.

\bibitem[\protect\astroncite{Bahdanau et~al.}{2017}]{Bahdanau2017}
Bahdanau, D., Brakel, P., Xu, K., Goyal, A., Lowe, R., Pineau, J., Courville,
  A., and Bengio, Y. (2017).
\newblock An actor-critic algorithm for sequence prediction.
\newblock In {\em the International Conference on Learning Representations
  (ICLR)}.

\bibitem[\protect\astroncite{Bahdanau et~al.}{2015}]{Bahdanau2015}
Bahdanau, D., Cho, K., and Bengio, Y. (2015).
\newblock Neural machine translation by jointly learning to align and
  translate.
\newblock In {\em the International Conference on Learning Representations
  (ICLR)}.

\bibitem[\protect\astroncite{{Bailis} et~al.}{2017}]{Bailis2017}
{Bailis}, P., {Olukoton}, K., {Re}, C., and {Zaharia}, M. (2017).
\newblock {Infrastructure for Usable Machine Learning: The Stanford DAWN
  Project}.
\newblock {\em ArXiv e-prints}.

\bibitem[\protect\astroncite{Baird}{1995}]{Baird1995}
Baird, L. (1995).
\newblock Residual algorithms: Reinforcement learning with function
  approximation.
\newblock In {\em the International Conference on Machine Learning (ICML)}.

\bibitem[\protect\astroncite{Baker et~al.}{2017}]{Baker2017}
Baker, B., Gupta, O., Naik, N., and Raskar, R. (2017).
\newblock Designing neural network architectures using reinforcement learning.
\newblock In {\em the International Conference on Learning Representations
  (ICLR)}.

\bibitem[\protect\astroncite{Balle et~al.}{2016}]{Balle2016}
Balle, B., Gomrokchi, M., and Precup, D. (2016).
\newblock Differentially private policy evaluation.
\newblock In {\em the International Conference on Machine Learning (ICML)}.

\bibitem[\protect\astroncite{Balog et~al.}{2017}]{Balog2017}
Balog, M., Gaunt, A.~L., Brockschmidt, M., Nowozin, S., and Tarlow, D. (2017).
\newblock Deepcoder: Learning to write programs.
\newblock In {\em the International Conference on Learning Representations
  (ICLR)}.

\bibitem[\protect\astroncite{{Bansal} et~al.}{2017}]{Bansal2017}
{Bansal}, T., {Pachocki}, J., {Sidor}, S., {Sutskever}, I., and {Mordatch}, I.
  (2017).
\newblock {Emergent Complexity via Multi-Agent Competition}.
\newblock {\em ArXiv e-prints}.

\bibitem[\protect\astroncite{{Barreto} et~al.}{2017}]{Barreto2017}
{Barreto}, A., {Munos}, R., {Schaul}, T., and {Silver}, D. (2017).
\newblock Successor features for transfer in reinforcement learning.
\newblock In {\em the Annual Conference on Neural Information Processing
  Systems (NIPS)}.

\bibitem[\protect\astroncite{Barto and Mahadevan}{2003}]{Barto2003}
Barto, A.~G. and Mahadevan, S. (2003).
\newblock Recent advances in hierarchical reinforcement learning.
\newblock {\em Discrete Event Dynamic Systems}, 13(4):341--379.

\bibitem[\protect\astroncite{Barto et~al.}{1983}]{Barto1983}
Barto, A.~G., Sutton, R.~S., and Anderson, C.~W. (1983).
\newblock Neuronlike elements that can solve difficult learning control
  problems.
\newblock {\em IEEE Transactions on Systems, Man, and Cybernetics},
  13:835--846.

\bibitem[\protect\astroncite{Battaglia et~al.}{2016}]{Battaglia2016}
Battaglia, P.~W., Pascanu, R., Lai, M., Rezende, D., and Kavukcuoglu, K.
  (2016).
\newblock Interaction networks for learning about objects, relations and
  physics.
\newblock In {\em the Annual Conference on Neural Information Processing
  Systems (NIPS)}.

\bibitem[\protect\astroncite{Bazzan and Kl{\"u}gl}{2014}]{Bazzan2014}
Bazzan, A.~L. and Kl{\"u}gl, F. (2014).
\newblock {\em Introduction to Intelligent Systems in Traffic and
  Transportation}.
\newblock Morgan \& Claypool.

\bibitem[\protect\astroncite{Beattie et~al.}{2016}]{Beattie2016}
Beattie, C., Leibo, J.~Z., Teplyashin, D., Ward, T., Wainwright, M.,
  K{\"u}ttler, H., Lefrancq, A., Green, S., Vald{\'e}s, V., Sadik, A.,
  Schrittwieser, J., Anderson, K., York, S., Cant, M., Cain, A., Bolton, A.,
  Gaffney, S., King, H., Hassabis, D., Legg, S., and Petersen, S. (2016).
\newblock {DeepMind Lab}.
\newblock {\em ArXiv e-prints}.

\bibitem[\protect\astroncite{Bellemare
  et~al.}{2017}]{Bellemare2017Distributional}
Bellemare, M.~G., Dabney, W., and Munos, R. (2017).
\newblock A distributional perspective on reinforcement learning.
\newblock In {\em the International Conference on Machine Learning (ICML)}.

\bibitem[\protect\astroncite{{Bellemare} et~al.}{2017}]{Bellemare2017Cramer}
{Bellemare}, M.~G., {Danihelka}, I., {Dabney}, W., {Mohamed}, S.,
  {Lakshminarayanan}, B., {Hoyer}, S., and {Munos}, R. (2017).
\newblock {The Cramer Distance as a Solution to Biased Wasserstein Gradients}.
\newblock {\em ArXiv e-prints}.

\bibitem[\protect\astroncite{Bellemare et~al.}{2013}]{Bellemare2013}
Bellemare, M.~G., Naddaf, Y., Veness, J., and Bowling, M. (2013).
\newblock The arcade learning environment: {A}n evaluation platform for general
  agents.
\newblock {\em Journal of Artificial Intelligence Research}, 47:253--279.

\bibitem[\protect\astroncite{Bellemare et~al.}{2016}]{Bellemare2016}
Bellemare, M.~G., Schaul, T., Srinivasan, S., Saxton, D., Ostrovski, G., and
  Munos, R. (2016).
\newblock Unifying count-based exploration and intrinsic motivation.
\newblock In {\em the Annual Conference on Neural Information Processing
  Systems (NIPS)}.

\bibitem[\protect\astroncite{{Bello} et~al.}{2016}]{Bello2016}
{Bello}, I., {Pham}, H., {Le}, Q.~V., {Norouzi}, M., and {Bengio}, S. (2016).
\newblock {Neural Combinatorial Optimization with Reinforcement Learning}.
\newblock {\em ArXiv e-prints}.

\bibitem[\protect\astroncite{Bello et~al.}{2017}]{Bello2017}
Bello, I., Zoph, B., Vasudevan, V., and Le, Q.~V. (2017).
\newblock Neural optimizer search with reinforcement learning.
\newblock In {\em the International Conference on Machine Learning (ICML)}.

\bibitem[\protect\astroncite{Bengio}{2009}]{Bengio2009}
Bengio, Y. (2009).
\newblock Learning deep architectures for ai.
\newblock {\em Foundations and trends\textregistered in Machine Learning},
  2(1):1--127.

\bibitem[\protect\astroncite{{Bengio}}{2017}]{Bengio2017}
{Bengio}, Y. (2017).
\newblock {The Consciousness Prior}.
\newblock {\em ArXiv e-prints}.

\bibitem[\protect\astroncite{Bengio et~al.}{2009}]{Bengio-2009-Curriculum}
Bengio, Y., Louradour, J., Collobert, R., and Weston, J. (2009).
\newblock Curriculum learning.
\newblock In {\em the International Conference on Machine Learning (ICML)}.

\bibitem[\protect\astroncite{Berkenkamp et~al.}{2017}]{Berkenkamp2017}
Berkenkamp, F., Turchetta, M., Schoellig, A.~P., and Krause, A. (2017).
\newblock Safe model-based reinforcement learning with stability guarantees.
\newblock In {\em the Annual Conference on Neural Information Processing
  Systems (NIPS)}.

\bibitem[\protect\astroncite{Bernhard~Wymann et~al.}{2014}]{TORCS}
Bernhard~Wymann, E.~E., Guionneau, C., Dimitrakakis, C., and R{\'e}mi~Coulom,
  A.~S. (2014).
\newblock {TORCS}, {T}he {O}pen {R}acing {C}ar {S}imulator.
\newblock "http://www.torcs.org".

\bibitem[\protect\astroncite{{Berthelot} et~al.}{2017}]{Berthelot2017}
{Berthelot}, D., {Schumm}, T., and {Metz}, L. (2017).
\newblock {BEGAN: Boundary Equilibrium Generative Adversarial Networks}.
\newblock {\em ArXiv e-prints}.

\bibitem[\protect\astroncite{Bertsekas}{2012}]{Bertsekas12}
Bertsekas, D.~P. (2012).
\newblock {\em Dynamic programming and optimal control (Vol. II, 4th Edition:
  Approximate Dynamic Programming)}.
\newblock Athena Scientific, Massachusetts, USA.

\bibitem[\protect\astroncite{Bertsekas and Tsitsiklis}{1996}]{Bertsekas96}
Bertsekas, D.~P. and Tsitsiklis, J.~N. (1996).
\newblock {\em Neuro-Dynamic Programming}.
\newblock Athena Scientific.

\bibitem[\protect\astroncite{{Bhatti} et~al.}{2016}]{Bhatti2016}
{Bhatti}, S., {Desmaison}, A., {Miksik}, O., {Nardelli}, N., {Siddharth}, N.,
  and {Torr}, P.~H.~S. (2016).
\newblock {Playing Doom with {SLAM}-Augmented Deep Reinforcement Learning}.
\newblock {\em ArXiv e-prints}.

\bibitem[\protect\astroncite{Bishop}{2011}]{Bishop2011}
Bishop, C. (2011).
\newblock {\em Pattern Recognition and Machine Learning}.
\newblock Springer.

\bibitem[\protect\astroncite{Blei and Smyth}{2017}]{Blei2017}
Blei, D.~M. and Smyth, P. (2017).
\newblock Science and data science.
\newblock {\em PNAS}, 114(33):8689--8692.

\bibitem[\protect\astroncite{Bohg et~al.}{2017}]{Bohg2017}
Bohg, J., Hausman, K., Sankaran, B., Brock, O., Kragic, D., Schaal, S., and
  Sukhatme, G.~S. (2017).
\newblock Interactive perception: Leveraging action in perception and
  perception in action.
\newblock {\em IEEE Transactions on Robotics}, 33(6):1273--1291.

\bibitem[\protect\astroncite{Bojarski et~al.}{2016}]{Bojarski2016}
Bojarski, M., Testa, D.~D., Dworakowski, D., Firner, B., Flepp, B., Goyal, P.,
  Jackel, L.~D., Monfort, M., Muller, U., Zhang, J., Zhang, X., Zhao, J., and
  Zieba, K. (2016).
\newblock {End to End Learning for Self-Driving Cars}.
\newblock {\em ArXiv e-prints}.

\bibitem[\protect\astroncite{{Bojarski} et~al.}{2017}]{Bojarski2017}
{Bojarski}, M., {Yeres}, P., {Choromanska}, A., {Choromanski}, K., {Firner},
  B., {Jackel}, L., and {Muller}, U. (2017).
\newblock {Explaining How a Deep Neural Network Trained with End-to-End
  Learning Steers a Car}.
\newblock {\em ArXiv e-prints}.

\bibitem[\protect\astroncite{Bordes et~al.}{2017}]{Bordes2017}
Bordes, A., Boureau, Y.-L., and Weston, J. (2017).
\newblock Learning end-to-end goal-oriented dialog.
\newblock In {\em the International Conference on Learning Representations
  (ICLR)}.

\bibitem[\protect\astroncite{Botvinick et~al.}{2017}]{Botvinick2017}
Botvinick, M., Barrett, D. G.~T., Battaglia, P., de~Freitas, N., Kumaran, D.,
  Leibo, J.~Z., Lillicrap, T., Modayil, J., Mohamed, S., Rabinowitz, N.~C.,
  Rezende, D.~J., Santoro, A., Schaul, T., Summerfield, C., Wayne, G., Weber,
  T., Wierstra, D., Legg, S., and Hassabis, D. (2017).
\newblock Building machines that learn and think for themselves.
\newblock {\em Behavioral and Brain Sciences}, 40.

\bibitem[\protect\astroncite{{Bousmalis} et~al.}{2017}]{Bousmalis2017}
{Bousmalis}, K., {Irpan}, A., {Wohlhart}, P., {Bai}, Y., {Kelcey}, M.,
  {Kalakrishnan}, M., {Downs}, L., {Ibarz}, J., {Pastor}, P., {Konolige}, K.,
  {Levine}, S., and {Vanhoucke}, V. (2017).
\newblock {Using Simulation and Domain Adaptation to Improve Efficiency of Deep
  Robotic Grasping}.
\newblock {\em ArXiv e-prints}.

\bibitem[\protect\astroncite{Bowling et~al.}{2015}]{Bowling2015}
Bowling, M., Burch, N., Johanson, M., and Tammelin, O. (2015).
\newblock Heads-up limit hold'em poker is solved.
\newblock {\em Science}, 347(6218):145--149.

\bibitem[\protect\astroncite{Boyd and Vandenberghe}{2004}]{Boyd04}
Boyd, S. and Vandenberghe, L. (2004).
\newblock {\em Convex Optimization}.
\newblock Cambridge University Press.

\bibitem[\protect\astroncite{Bradtke and Barto}{1996}]{Bradtke96}
Bradtke, S.~J. and Barto, A.~G. (1996).
\newblock Linear least-squares algorithms for temporal difference learning.
\newblock {\em Machine Learning}, 22(1-3):33--57.

\bibitem[\protect\astroncite{Brandt et~al.}{2005}]{Brandt05}
Brandt, M.~W., Goyal, A., Santa-Clara, P., and Stroud, J.~R. (2005).
\newblock A simulation approach to dynamic portfolio choice with an application
  to learning about return predictability.
\newblock {\em The Review of Financial Studies}, 18(3):831--873.

\bibitem[\protect\astroncite{{Briot} et~al.}{2017}]{Briot2017}
{Briot}, J.-P., {Hadjeres}, G., and {Pachet}, F. (2017).
\newblock {Deep Learning Techniques for Music Generation - A Survey}.
\newblock {\em ArXiv e-prints}.

\bibitem[\protect\astroncite{Browne et~al.}{2012}]{Browne2012}
Browne, C., Powley, E., Whitehouse, D., Lucas, S., Cowling, P.~I., Rohlfshagen,
  P., Tavener, S., Perez, D., Samothrakis, S., and Colton, S. (2012).
\newblock A survey of {Monte Carlo} tree search methods.
\newblock {\em IEEE Transactions on Computational Intelligence and AI in
  Games}, 4(1):1--43.

\bibitem[\protect\astroncite{Brunner et~al.}{2018}]{Brunner2018}
Brunner, G., Richter, O., Wang, Y., and Wattenhofer, R. (2018).
\newblock Teaching a machine to read maps with deep reinforcement learning.
\newblock In {\em the AAAI Conference on Artificial Intelligence (AAAI)}.

\bibitem[\protect\astroncite{Busoniu et~al.}{2008}]{Busoniu2008}
Busoniu, L., Babuska, R., and Schutter, B.~D. (2008).
\newblock A comprehensive survey of multiagent reinforcement learning.
\newblock {\em IEEE Transactions on Systems, Man, and Cybernetics - Part C:
  Applications and Reviews}, 38(2).

\bibitem[\protect\astroncite{Cai et~al.}{2017}]{Cai2017}
Cai, J., Shin, R., and Song, D. (2017).
\newblock Making neural programming architectures generalize via recursion.
\newblock In {\em the International Conference on Learning Representations
  (ICLR)}.

\bibitem[\protect\astroncite{Caicedo and Lazebnik}{2015}]{Caicedo2015}
Caicedo, J.~C. and Lazebnik, S. (2015).
\newblock Active object localization with deep reinforcement learning.
\newblock In {\em the IEEE International Conference on Computer Vision (ICCV)}.

\bibitem[\protect\astroncite{Cao et~al.}{2017}]{Cao2017}
Cao, Q., Lin, L., Shi, Y., Liang, X., and Li, G. (2017).
\newblock Attention-aware face hallucination via deep reinforcement learning.
\newblock In {\em the IEEE Conference on Computer Vision and Pattern
  Recognition (CVPR)}.

\bibitem[\protect\astroncite{Carleo and Troyer}{2017}]{Carleo2017}
Carleo, G. and Troyer, M. (2017).
\newblock Solving the quantum many-body problem with artificial neural
  networks.
\newblock {\em Science}, 355(6325):602--606.

\bibitem[\protect\astroncite{Carlini and Wagner}{2017a}]{Carlini2017}
Carlini, N. and Wagner, D. (2017a).
\newblock Adversarial examples are not easily detected: Bypassing ten detection
  methods.
\newblock In {\em ACM CCS 2017 Workshop on Artificial Intelligence and
  Security}.

\bibitem[\protect\astroncite{Carlini and Wagner}{2017b}]{Carlini2017SP}
Carlini, N. and Wagner, D. (2017b).
\newblock Towards evaluating the robustness of neural networks.
\newblock In {\em IEEE Symposium on Security and Privacy}.

\bibitem[\protect\astroncite{{Celikyilmaz} et~al.}{2017}]{Celikyilmaz2017}
{Celikyilmaz}, A., {Deng}, L., {Li}, L., and {Wang}, C. (2017).
\newblock {Scaffolding Networks: Incremental Learning and Teaching Through
  Questioning}.
\newblock {\em ArXiv e-prints}.

\bibitem[\protect\astroncite{Chakraborty and Murphy}{2014}]{Chakraborty2014}
Chakraborty, B. and Murphy, S.~A. (2014).
\newblock Dynamic treatment regimes.
\newblock {\em Annual Review of Statistics and Its Application}, 1:447--464.

\bibitem[\protect\astroncite{Chebotar et~al.}{2017}]{Chebotar2017}
Chebotar, Y., Hausman, K., Zhang, M., Sukhatme, G., Schaal, S., and Levine, S.
  (2017).
\newblock Combining model-based and model-free updates for trajectory-centric
  reinforcement learning.
\newblock In {\em the International Conference on Machine Learning (ICML)}.

\bibitem[\protect\astroncite{Chebotar et~al.}{2016}]{Chebotar2016}
Chebotar, Y., Kalakrishnan, M., Yahya, A., Li, A., Schaal, S., and Levine, S.
  (2016).
\newblock Path integral guided policy search.
\newblock {\em ArXiv e-prints}.

\bibitem[\protect\astroncite{{Chen} et~al.}{2016}]{Chen2016unpaired}
{Chen}, J., {Huang}, P.-S., {He}, X., {Gao}, J., and {Deng}, L. (2016).
\newblock {Unsupervised Learning of Predictors from Unpaired Input-Output
  Samples}.
\newblock {\em ArXiv e-prints}.

\bibitem[\protect\astroncite{Chen et~al.}{2016a}]{Chen2016InfoGAN}
Chen, X., Duan, Y., Houthooft, R., Schulman, J., Sutskever, I., and Abbeel, P.
  (2016a).
\newblock {InfoGAN}: Interpretable representation learning by information
  maximizing generative adversarial nets.
\newblock In {\em the Annual Conference on Neural Information Processing
  Systems (NIPS)}.

\bibitem[\protect\astroncite{Chen et~al.}{2016b}]{Chen2016-knowledge}
Chen, Y.-N., Hakkani-Tur, D., Tur, G., Celikyilmaz, A., Gao, J., and Deng, L.
  (2016b).
\newblock {Knowledge as a Teacher: Knowledge-Guided Structural Attention
  Networks}.
\newblock {\em ArXiv e-prints}.

\bibitem[\protect\astroncite{Chen et~al.}{2016c}]{Chen2016}
Chen, Y.-N.~V., Hakkani-T{\"u}r, D., Tur, G., Gao, J., and Deng, L. (2016c).
\newblock End-to-end memory networks with knowledge carryover for multi-turn
  spoken language understanding.
\newblock In {\em Annual Meeting of the International Speech Communication
  Association (INTERSPEECH)}.

\bibitem[\protect\astroncite{Chen and Liu}{2016}]{Chen2016Lifelong}
Chen, Z. and Liu, B. (2016).
\newblock {\em Lifelong Machine Learning}.
\newblock Morgan \& Claypool Publishers.

\bibitem[\protect\astroncite{{Chen} and {Yi}}{2017}]{Chen2017}
{Chen}, Z. and {Yi}, D. (2017).
\newblock {The Game Imitation: Deep Supervised Convolutional Networks for Quick
  Video Game AI}.
\newblock {\em ArXiv e-prints}.

\bibitem[\protect\astroncite{{Cheng} et~al.}{2016}]{Cheng2016}
{Cheng}, Y., {Xu}, W., {He}, Z., {He}, W., {Wu}, H., {Sun}, M., and {Liu}, Y.
  (2016).
\newblock Semi-supervised learning for neural machine translation.
\newblock In {\em the Association for Computational Linguistics annual meeting
  (ACL)}.

\bibitem[\protect\astroncite{Cho}{2015}]{Cho2015}
Cho, K. (2015).
\newblock {Natural Language Understanding with Distributed Representation}.
\newblock {\em ArXiv e-prints}.

\bibitem[\protect\astroncite{Cho et~al.}{2014}]{Cho2014}
Cho, K., van Merrienboer, B., Gulcehre, C., Bougares, F., Schwenk, H., and
  Bengio, Y. (2014).
\newblock Learning phrase representations using {RNN} encoder-decoder for
  statistical machine translation.
\newblock In {\em Conference on Empirical Methods in Natural Language
  Processing (EMNLP)}.

\bibitem[\protect\astroncite{Choi et~al.}{2017}]{Choi2017}
Choi, E., Hewlett, D., Polosukhin, I., Lacoste, A., Uszkoreit, J., and Berant,
  J. (2017).
\newblock Coarse-to-fine question answering for long documents.
\newblock In {\em the Association for Computational Linguistics annual meeting
  (ACL)}.

\bibitem[\protect\astroncite{Christiano et~al.}{2017}]{Christiano2017}
Christiano, P., Leike, J., Brown, T.~B., Martic, M., Legg, S., and Amodei, D.
  (2017).
\newblock Deep reinforcement learning from human preferences.
\newblock In {\em the Annual Conference on Neural Information Processing
  Systems (NIPS)}.

\bibitem[\protect\astroncite{Chung et~al.}{2014}]{Chung2014}
Chung, J., Gulcehre, C., Cho, K., and Bengio, Y. (2014).
\newblock Empirical evaluation of gated recurrent neural networks on sequence
  modeling.
\newblock In {\em NIPS 2014 Deep Learning and Representation Learning
  Workshop}.

\bibitem[\protect\astroncite{Crawford et~al.}{2016}]{Crawford2016}
Crawford, D., Levit, A., Ghadermarzy, N., Oberoi, J.~S., and Ronagh, P. (2016).
\newblock {Reinforcement Learning Using Quantum Boltzmann Machines}.
\newblock {\em ArXiv e-prints}.

\bibitem[\protect\astroncite{{Czarnecki} et~al.}{2017}]{Czarnecki2017}
{Czarnecki}, W.~M., {{\'S}wirszcz}, G., {Jaderberg}, M., {Osindero}, S.,
  {Vinyals}, O., and {Kavukcuoglu}, K. (2017).
\newblock {Understanding Synthetic Gradients and Decoupled Neural Interfaces}.
\newblock {\em ArXiv e-prints}.

\bibitem[\protect\astroncite{Dai et~al.}{2017}]{Dai2017graph}
Dai, H., Khalil, E.~B., Zhang, Y., Dilkina, B., and Song, L. (2017).
\newblock Learning combinatorial optimization algorithms over graphs.
\newblock In {\em the Annual Conference on Neural Information Processing
  Systems (NIPS)}.

\bibitem[\protect\astroncite{{Dai} et~al.}{2017}]{Dai2017}
{Dai}, Z., {Yang}, Z., {Yang}, F., {Cohen}, W.~W., and {Salakhutdinov}, R.
  (2017).
\newblock {Good Semi-supervised Learning that Requires a Bad GAN}.
\newblock {\em ArXiv e-prints}.

\bibitem[\protect\astroncite{Daniely et~al.}{2016}]{Daniely2016}
Daniely, A., Frostig, R., and Singer, Y. (2016).
\newblock Toward deeper understanding of neural networks: The power of
  initialization and a dual view on expressivity.
\newblock In {\em the Annual Conference on Neural Information Processing
  Systems (NIPS)}.

\bibitem[\protect\astroncite{Danihelka et~al.}{2016}]{Danihelka2016}
Danihelka, I., Wayne, G., Uria, B., Kalchbrenner, N., and Graves, A. (2016).
\newblock Associative long short-term memory.
\newblock In {\em the International Conference on Machine Learning (ICML)}.

\bibitem[\protect\astroncite{{Das} et~al.}{2017}]{Das2017}
{Das}, A., {Kottur}, S., {Moura}, J.~M.~F., {Lee}, S., and {Batra}, D. (2017).
\newblock Learning cooperative visual dialog agents with deep reinforcement
  learning.
\newblock In {\em the IEEE International Conference on Computer Vision (ICCV)}.

\bibitem[\protect\astroncite{{De Asis} et~al.}{2018}]{DeAsis2018}
{De Asis}, K., {Hernandez-Garcia}, J.~F., {Zacharias Holland}, G., and
  {Sutton}, R.~S. (2018).
\newblock Multi-step reinforcement learning: A unifying algorithm.
\newblock In {\em the AAAI Conference on Artificial Intelligence (AAAI)}.

\bibitem[\protect\astroncite{Deisenroth et~al.}{2013}]{Deisenroth2013}
Deisenroth, M.~P., Neumann, G., and Peters, J. (2013).
\newblock A survey on policy search for robotics.
\newblock {\em Foundations and Trend in Robotics}, 2:1--142.

\bibitem[\protect\astroncite{Deisenroth and Rasmussen}{2011}]{Deisenroth2011}
Deisenroth, M.~P. and Rasmussen, C.~E. (2011).
\newblock {PILCO}: A model-based and data-efficient approach to policy search.
\newblock In {\em the International Conference on Machine Learning (ICML)}.

\bibitem[\protect\astroncite{{Delle Fave} et~al.}{2014}]{Delle2014}
{Delle Fave}, F.~M., Jiang, A.~X., Yin, Z., Zhang, C., Tambe, M., Kraus, S.,
  and Sullivan, J.~P. (2014).
\newblock Game-theoretic security patrolling with dynamic execution uncertainty
  and a case study on a real transit system.
\newblock 50:321--367.

\bibitem[\protect\astroncite{Deng}{2017}]{Deng2017AIFrontiers}
Deng, L. (2017).
\newblock Three generations of spoken dialogue systems (bots), talk at {AI
  Frontiers Conference}.
\newblock
  \url{https://www.slideshare.net/AIFrontiers/li-deng-three-generations-of-spoken-dialogue-systems-bots}.

\bibitem[\protect\astroncite{Deng and Dong}{2014}]{Deng2014}
Deng, L. and Dong, Y. (2014).
\newblock {\em Deep Learning: Methods and Applications}.
\newblock Now Publishers Inc.

\bibitem[\protect\astroncite{Deng and Li}{2013}]{Deng2013}
Deng, L. and Li, X. (2013).
\newblock Machine learning paradigms for speech recognition: An overview.
\newblock {\em IEEE Transactions on Audio, Speech, and Language Processing},
  21(5):1060--1089.

\bibitem[\protect\astroncite{Deng and Liu}{2017}]{Deng2017}
Deng, L. and Liu, Y. (2017).
\newblock {\em Deep Learning in Natural Language Processing (edited book,
  scheduled August 2017)}.
\newblock Springer.

\bibitem[\protect\astroncite{Deng et~al.}{2016}]{Deng2016}
Deng, Y., Bao, F., Kong, Y., Ren, Z., and Dai, Q. (2016).
\newblock Deep direct reinforcement learning for financial signal
  representation and trading.
\newblock {\em IEEE Transactions on Neural Networks and Learning Systems}.

\bibitem[\protect\astroncite{Denil et~al.}{2017}]{Denil2017}
Denil, M., Agrawal, P., Kulkarni, T.~D., Erez, T., Battaglia, P., and
  de~Freitas, N. (2017).
\newblock Learning to perform physics experiments via deep reinforcement
  learning.
\newblock In {\em the International Conference on Learning Representations
  (ICLR)}.

\bibitem[\protect\astroncite{{Denil} et~al.}{2017}]{Denil2017agent}
{Denil}, M., {G{\'o}mez Colmenarejo}, S., {Cabi}, S., {Saxton}, D., and {de
  Freitas}, N. (2017).
\newblock {Programmable Agents}.
\newblock {\em ArXiv e-prints}.

\bibitem[\protect\astroncite{Devrim~Kaba et~al.}{2017}]{Kaba2017}
Devrim~Kaba, M., Gokhan~Uzunbas, M., and Nam~Lim, S. (2017).
\newblock A reinforcement learning approach to the view planning problem.
\newblock In {\em the IEEE Conference on Computer Vision and Pattern
  Recognition (CVPR)}.

\bibitem[\protect\astroncite{Dhingra et~al.}{2017}]{Dhingra2017}
Dhingra, B., Li, L., Li, X., Gao, J., Chen, Y.-N., Ahmed, F., and Deng, L.
  (2017).
\newblock End-to-end reinforcement learning of dialogue agents for information
  access.
\newblock In {\em the Association for Computational Linguistics annual meeting
  (ACL)}.

\bibitem[\protect\astroncite{Diederik P~Kingma}{2014}]{Kingma2014VAE}
Diederik P~Kingma, M.~W. (2014).
\newblock Auto-encoding variational bayes.
\newblock In {\em the International Conference on Learning Representations
  (ICLR)}.

\bibitem[\protect\astroncite{Dietterich}{2000}]{Dietterich2000}
Dietterich, T.~G. (2000).
\newblock Hierarchical reinforcement learning with the {MAXQ} value function
  decomposition.
\newblock {\em Journal of Artificial Intelligence Research}, 13(1):227--303.

\bibitem[\protect\astroncite{Domingos}{2012}]{Domingos2012}
Domingos, P. (2012).
\newblock A few useful things to know about machine learning.
\newblock {\em Communications of the ACM}, 55(10):78--87.

\bibitem[\protect\astroncite{Dong et~al.}{2015}]{Dong2015}
Dong, D., Wu, H., He, W., Yu, D., and Wang, H. (2015).
\newblock Multi-task learning for multiple language translation.
\newblock In {\em the Association for Computational Linguistics annual meeting
  (ACL)}.

\bibitem[\protect\astroncite{Dong et~al.}{2017}]{Dong2017}
Dong, Y., Su, H., Zhu, J., and Zhang, B. (2017).
\newblock Improving interpretability of deep neural networks with semantic
  information.
\newblock In {\em the IEEE Conference on Computer Vision and Pattern
  Recognition (CVPR)}.

\bibitem[\protect\astroncite{{Doshi-Velez} and {Kim}}{2017}]{Doshi-Velez2017}
{Doshi-Velez}, F. and {Kim}, B. (2017).
\newblock {Towards A Rigorous Science of Interpretable Machine Learning}.
\newblock {\em ArXiv e-prints}.

\bibitem[\protect\astroncite{Dosovitskiy and Koltun}{2017}]{Dosovitskiy2017}
Dosovitskiy, A. and Koltun, V. (2017).
\newblock Learning to act by predicting the future.
\newblock In {\em the International Conference on Learning Representations
  (ICLR)}.

\bibitem[\protect\astroncite{Downey et~al.}{2017}]{Downey2017}
Downey, C., Hefny, A., Li, B., Boots, B., and Gordon, G. (2017).
\newblock Predictive state recurrent neural networks.
\newblock In {\em the Annual Conference on Neural Information Processing
  Systems (NIPS)}.

\bibitem[\protect\astroncite{Du et~al.}{2017}]{Du2017PE}
Du, S.~S., Chen, J., Li, L., Xiao, L., and Zhou, D. (2017).
\newblock Stochastic variance reduction methods for policy evaluation.
\newblock In {\em the International Conference on Machine Learning (ICML)}.

\bibitem[\protect\astroncite{{Duan} et~al.}{2017}]{Duan2017OneShot}
{Duan}, Y., {Andrychowicz}, M., {Stadie}, B.~C., {Ho}, J., {Schneider}, J.,
  {Sutskever}, I., {Abbeel}, P., and {Zaremba}, W. (2017).
\newblock One-shot imitation learning.
\newblock In {\em the Annual Conference on Neural Information Processing
  Systems (NIPS)}.

\bibitem[\protect\astroncite{Duan et~al.}{2016}]{Duan2016}
Duan, Y., Chen, X., Houthooft, R., Schulman, J., and Abbeel, P. (2016).
\newblock Benchmarking deep reinforcement learning for continuous control.
\newblock In {\em the International Conference on Machine Learning (ICML)}.

\bibitem[\protect\astroncite{{Duan} et~al.}{2016}]{Duan2016RL2}
{Duan}, Y., {Schulman}, J., {Chen}, X., {Bartlett}, P.~L., {Sutskever}, I., and
  {Abbeel}, P. (2016).
\newblock {RL$^2$: Fast Reinforcement Learning via Slow Reinforcement
  Learning}.
\newblock {\em ArXiv e-prints}.

\bibitem[\protect\astroncite{{Dulac-Arnold} et~al.}{2015}]{Dulac-Arnold2016}
{Dulac-Arnold}, G., {Evans}, R., {van Hasselt}, H., {Sunehag}, P., {Lillicrap},
  T., {Hunt}, J., {Mann}, T., {Weber}, T., {Degris}, T., and {Coppin}, B.
  (2015).
\newblock {Deep Reinforcement Learning in Large Discrete Action Spaces}.
\newblock {\em ArXiv e-prints}.

\bibitem[\protect\astroncite{El-Tantawy et~al.}{2013}]{El-Tantawy2013}
El-Tantawy, S., Abdulhai, B., and Abdelgawad, H. (2013).
\newblock Multiagent reinforcement learning for integrated network of adaptive
  traffic signal controllers (marlin-atsc): methodology and large-scale
  application on downtown toronto.
\newblock {\em IEEE Transactions on Intelligent Transportation Systems},
  14(3):1140--1150.

\bibitem[\protect\astroncite{{Eric} and {Manning}}{2017}]{Eric2017}
{Eric}, M. and {Manning}, C.~D. (2017).
\newblock {A Copy-Augmented Sequence-to-Sequence Architecture Gives Good
  Performance on Task-Oriented Dialogue}.
\newblock {\em ArXiv e-prints}.

\bibitem[\protect\astroncite{Ernst et~al.}{2005}]{Ernst2005}
Ernst, D., Geurts, P., and Wehenkel, L. (2005).
\newblock Tree-based batch mode reinforcement learning.
\newblock {\em The Journal of Machine Learning Research}, 6:503--556.

\bibitem[\protect\astroncite{Eslami et~al.}{2016}]{Eslami2016}
Eslami, S. M.~A., Heess, N., Weber, T., Tassa, Y., Szepesv{\'a}ri, D.,
  Kavukcuoglu, K., and Hinton, G.~E. (2016).
\newblock Attend, infer, repeat: Fast scene understanding with generative
  models.
\newblock In {\em the Annual Conference on Neural Information Processing
  Systems (NIPS)}.

\bibitem[\protect\astroncite{{Evans} and {Grefenstette}}{2017}]{Evans2017}
{Evans}, R. and {Grefenstette}, E. (2017).
\newblock {Learning Explanatory Rules from Noisy Data}.
\newblock {\em ArXiv e-prints}.

\bibitem[\protect\astroncite{{Evtimov} et~al.}{2017}]{Evtimov2017}
{Evtimov}, I., {Eykholt}, K., {Fernandes}, E., {Kohno}, T., {Li}, B.,
  {Prakash}, A., {Rahmati}, A., and {Song}, D. (2017).
\newblock {Robust Physical-World Attacks on Deep Learning Models}.
\newblock {\em ArXiv e-prints}.

\bibitem[\protect\astroncite{Fang et~al.}{2017}]{Fang2017}
Fang, M., Li, Y., and Cohn, T. (2017).
\newblock Learning how to active learn: A deep reinforcement learning approach.
\newblock In {\em Conference on Empirical Methods in Natural Language
  Processing (EMNLP)}.

\bibitem[\protect\astroncite{Fang et~al.}{2012}]{Fang2012}
Fang, X., Misra, S., Xue, G., and Yang, D. (2012).
\newblock Smart grid - the new and improved power grid: A survey.
\newblock {\em IEEE Communications Surveys Tutorials}, 14(4):944--980.

\bibitem[\protect\astroncite{Fatemi et~al.}{2016}]{Fatemi2016}
Fatemi, M., Asri, L.~E., Schulz, H., He, J., and Suleman, K. (2016).
\newblock Policy networks with two-stage training for dialogue systems.
\newblock In {\em the Annual SIGdial Meeting on Discourse and Dialogue
  (SIGDIAL)}.

\bibitem[\protect\astroncite{{Feng} and {Zhou}}{2017}]{Feng2017}
{Feng}, J. and {Zhou}, Z.-H. (2017).
\newblock {AutoEncoder by Forest}.
\newblock {\em ArXiv e-prints}.

\bibitem[\protect\astroncite{{Fernando} et~al.}{2017}]{Fernando2017}
{Fernando}, C., {Banarse}, D., {Blundell}, C., {Zwols}, Y., {Ha}, D., {Rusu},
  A.~A., {Pritzel}, A., and {Wierstra}, D. (2017).
\newblock {PathNet: Evolution Channels Gradient Descent in Super Neural
  Networks}.
\newblock {\em ArXiv e-prints}.

\bibitem[\protect\astroncite{Finn et~al.}{2016a}]{Finn2016-Connection}
Finn, C., Christiano, P., Abbeel, P., and Levine, S. (2016a).
\newblock A connection between {GAN}s, inverse reinforcement learning, and
  energy-based models.
\newblock In {\em NIPS 2016 Workshop on Adversarial Training}.

\bibitem[\protect\astroncite{Finn and Levine}{2016}]{Finn-robotics-2016}
Finn, C. and Levine, S. (2016).
\newblock Deep visual foresight for planning robot motion.
\newblock In {\em IEEE International Conference on Robotics and Automation
  (ICRA)}.

\bibitem[\protect\astroncite{Finn et~al.}{2016b}]{Finn2016}
Finn, C., Levine, S., and Abbeel, P. (2016b).
\newblock Guided cost learning: Deep inverse optimal control via policy
  optimization.
\newblock In {\em the International Conference on Machine Learning (ICML)}.

\bibitem[\protect\astroncite{Finn et~al.}{2017}]{Finn2017}
Finn, C., Yu, T., Fu, J., Abbeel, P., and Levine, S. (2017).
\newblock Generalizing skills with semi-supervised reinforcement learning.
\newblock In {\em the International Conference on Learning Representations
  (ICLR)}.

\bibitem[\protect\astroncite{{Firoiu} et~al.}{2017}]{Firoiu2017}
{Firoiu}, V., {Whitney}, W.~F., and {Tenenbaum}, J.~B. (2017).
\newblock {Beating the World's Best at Super Smash Bros. with Deep
  Reinforcement Learning}.
\newblock {\em ArXiv e-prints}.

\bibitem[\protect\astroncite{Florensa et~al.}{2017}]{Florensa2017}
Florensa, C., Duan, Y., and Abbeel, P. (2017).
\newblock Stochastic neural networks for hierarchical reinforcement learning.
\newblock In {\em the International Conference on Learning Representations
  (ICLR)}.

\bibitem[\protect\astroncite{Foerster et~al.}{2016}]{Foerster2016}
Foerster, J., Assael, Y.~M., de~Freitas, N., and Whiteson, S. (2016).
\newblock Learning to communicate with deep multi-agent reinforcement learning.
\newblock In {\em the Annual Conference on Neural Information Processing
  Systems (NIPS)}.

\bibitem[\protect\astroncite{{Foerster} et~al.}{2018}]{Foerster2018MAPG}
{Foerster}, J., {Farquhar}, G., {Afouras}, T., {Nardelli}, N., and {Whiteson},
  S. (2018).
\newblock Counterfactual multi-agent policy gradients.
\newblock In {\em the AAAI Conference on Artificial Intelligence (AAAI)}.

\bibitem[\protect\astroncite{Foerster et~al.}{2017}]{Foerster2017}
Foerster, J., Nardelli, N., Farquhar, G., Torr, P. H.~S., Kohli, P., and
  Whiteson, S. (2017).
\newblock Stabilising experience replay for deep multi-agent reinforcement
  learning.
\newblock In {\em the International Conference on Machine Learning (ICML)}.

\bibitem[\protect\astroncite{{Foerster} et~al.}{2017}]{Foerster2017opponent}
{Foerster}, J.~N., {Chen}, R.~Y., {Al-Shedivat}, M., {Whiteson}, S., {Abbeel},
  P., and {Mordatch}, I. (2017).
\newblock {Learning with Opponent-Learning Awareness}.
\newblock {\em ArXiv e-prints}.

\bibitem[\protect\astroncite{{Fortunato} et~al.}{2017}]{Fortunato2017}
{Fortunato}, M., {Gheshlaghi Azar}, M., {Piot}, B., {Menick}, J., {Osband}, I.,
  {Graves}, A., {Mnih}, V., {Munos}, R., {Hassabis}, D., {Pietquin}, O.,
  {Blundell}, C., and {Legg}, S. (2017).
\newblock {Noisy Networks for Exploration}.
\newblock {\em ArXiv e-prints}.

\bibitem[\protect\astroncite{Fu et~al.}{2017}]{Fu2017}
Fu, J., Co-Reyes, J.~D., and Levine, S. (2017).
\newblock Ex2: Exploration with exemplar models for deep reinforcement
  learning.
\newblock In {\em the Annual Conference on Neural Information Processing
  Systems (NIPS)}.

\bibitem[\protect\astroncite{Ganin et~al.}{2016}]{Ganin2016}
Ganin, Y., Ustinova, E., Ajakan, H., Germain, P., Larochelle, H., Laviolette,
  F., Marchand, M., and Lempitsky, V. (2016).
\newblock Domain-adversarial training of neural networks.
\newblock {\em Journal of Machine Learning Research}, 17(59):1--35.

\bibitem[\protect\astroncite{Garc{\`i}a and Fern{\`a}ndez}{2015}]{Garcia2015}
Garc{\`i}a, J. and Fern{\`a}ndez, F. (2015).
\newblock A comprehensive survey on safe reinforcement learning.
\newblock {\em The Journal of Machine Learning Research}, 16:1437--1480.

\bibitem[\protect\astroncite{Gavrilovska et~al.}{2013}]{Gavrilovska2013}
Gavrilovska, L., Atanasovski, V., Macaluso, I., and DaSilva, L.~A. (2013).
\newblock Learning and reasoning in cognitive radio networks.
\newblock {\em IEEE Communications Surveys Tutorials}, 15(4):1761--1777.

\bibitem[\protect\astroncite{{Gehring} et~al.}{2017}]{Gehring2017}
{Gehring}, J., {Auli}, M., {Grangier}, D., {Yarats}, D., and {Dauphin}, Y.~N.
  (2017).
\newblock {Convolutional Sequence to Sequence Learning}.
\newblock {\em ArXiv e-prints}.

\bibitem[\protect\astroncite{Gelly et~al.}{2012}]{Gelly12}
Gelly, S., Schoenauer, M., Sebag, M., Teytaud, O., Kocsis, L., Silver, D., and
  Szepesv{\'a}ri, C. (2012).
\newblock The grand challenge of computer go: Monte carlo tree search and
  extensions.
\newblock {\em Communications of the ACM}, 55(3):106--113.

\bibitem[\protect\astroncite{Gelly and Silver}{2007}]{Gelly2007}
Gelly, S. and Silver, D. (2007).
\newblock Combining online and offline knowledge in uct.
\newblock In {\em the International Conference on Machine Learning (ICML)}.

\bibitem[\protect\astroncite{George et~al.}{2017}]{George2017}
George, D., Lehrach, W., Kansky, K., L{\'a}zaro-Gredilla, M., Laan, C., Marthi,
  B., Lou, X., Meng, Z., Liu, Y., Wang, H., Lavin, A., and Phoenix, D.~S.
  (2017).
\newblock A generative vision model that trains with high data efficiency and
  breaks text-based {CAPTCHA}s.
\newblock {\em Science}.

\bibitem[\protect\astroncite{Geramifard et~al.}{2015}]{Geramifard2015}
Geramifard, A., Dann, C., Klein, R.~H., Dabney, W., and How, J.~P. (2015).
\newblock Rlpy: A value-function-based reinforcement learning framework for
  education and research.
\newblock {\em Journal of Machine Learning Research}, 16:1573--1578.

\bibitem[\protect\astroncite{Geramifard et~al.}{2013}]{Geramifard2013}
Geramifard, A., Walsh, T.~J., Tellex, S., Chowdhary, G., Roy, N., and How,
  J.~P. (2013).
\newblock A tutorial on linear function approximators for dynamic programming
  and reinforcement learning.
\newblock {\em Foundations and Trends® in Machine Learning}, 6(4):375--451.

\bibitem[\protect\astroncite{Ghavamzadeh et~al.}{2006}]{Ghavamzadeh2006}
Ghavamzadeh, M., Mahadevan, S., and Makar, R. (2006).
\newblock Hierarchical multi-agent reinforcement learning.
\newblock {\em Autonomous Agents and Multi-Agent Systems}, 13(2):197--229.

\bibitem[\protect\astroncite{Ghavamzadeh et~al.}{2015}]{Ghavamzadeh2015}
Ghavamzadeh, M., Mannor, S., Pineau, J., and Tamar, A. (2015).
\newblock Bayesian reinforcement learning: a survey.
\newblock {\em Foundations and Trends in Machine Learning}, 8(5-6):359--483.

\bibitem[\protect\astroncite{Girshick}{2015}]{Girshick2015}
Girshick, R. (2015).
\newblock Fast {R-CNN}.
\newblock In {\em the IEEE International Conference on Computer Vision (ICCV)}.

\bibitem[\protect\astroncite{Glavic et~al.}{2017}]{Glavic2017}
Glavic, M., Fonteneau, R., and Ernst, D. (2017).
\newblock Reinforcement learning for electric power system decision and
  control: Past considerations and perspectives.
\newblock In {\em The 20th World Congress of the International Federation of
  Automatic Control}.

\bibitem[\protect\astroncite{Goldberg}{2017}]{Goldberg2017}
Goldberg, Y. (2017).
\newblock {\em Neural Network Methods for Natural Language Processing}.
\newblock Morgan \& Claypool Publishers.

\bibitem[\protect\astroncite{Goldberg and Kosorok}{2012}]{Goldberg2012}
Goldberg, Y. and Kosorok, M.~R. (2012).
\newblock Q-learning with censored data.
\newblock {\em Annals of Statistics}, 40(1):529--560.

\bibitem[\protect\astroncite{Goodfellow}{2017}]{Goodfellow2017}
Goodfellow, I. (2017).
\newblock {NIPS 2016 Tutorial: Generative Adversarial Networks}.
\newblock {\em ArXiv e-prints}.

\bibitem[\protect\astroncite{Goodfellow et~al.}{2016}]{Goodfellow2016}
Goodfellow, I., Bengio, Y., and Courville, A. (2016).
\newblock {\em Deep Learning}.
\newblock MIT Press.

\bibitem[\protect\astroncite{Goodfellow et~al.}{2014}]{Goodfellow2014}
Goodfellow, I., Pouget-Abadie, J., Mirza, M., Xu, B., Warde-Farley, D., Ozair,
  S., Courville, A., , and Bengio, Y. (2014).
\newblock Generative adversarial nets.
\newblock In {\em the Annual Conference on Neural Information Processing
  Systems (NIPS)}, page 2672?2680.

\bibitem[\protect\astroncite{{Graves} et~al.}{2017}]{Graves2017}
{Graves}, A., {Bellemare}, M.~G., {Menick}, J., {Munos}, R., and {Kavukcuoglu},
  K. (2017).
\newblock {Automated Curriculum Learning for Neural Networks}.
\newblock {\em ArXiv e-prints}.

\bibitem[\protect\astroncite{Graves et~al.}{2014}]{Graves2014}
Graves, A., Wayne, G., and Danihelka, I. (2014).
\newblock {Neural Turing Machines}.
\newblock {\em ArXiv e-prints}.

\bibitem[\protect\astroncite{Graves et~al.}{2016}]{Grave-DNC-2016}
Graves, A., Wayne, G., Reynolds, M., Harley, T., Danihelka, I.,
  Grabska-Barwi{\'n}ska, A., Colmenarejo, S.~G., Grefenstette, E., Ramalho, T.,
  Agapiou, J., nech Badia, A.~P., Hermann, K.~M., Zwols, Y., Ostrovski, G.,
  Cain, A., King, H., Summerfield, C., Blunsom, P., Kavukcuoglu, K., and
  Hassabis, D. (2016).
\newblock Hybrid computing using a neural network with dynamic external memory.
\newblock {\em Nature}, 538:471--476.

\bibitem[\protect\astroncite{Gregor et~al.}{2015}]{Gregor2015}
Gregor, K., Danihelka, I., Graves, A., Rezende, D., and Wierstra, D. (2015).
\newblock Draw: A recurrent neural network for image generation.
\newblock In {\em the International Conference on Machine Learning (ICML)}.

\bibitem[\protect\astroncite{Grondman et~al.}{2012}]{Grondman2012}
Grondman, I., Busoniu, L., Lopes, G.~A., and Babu{\v s}ka, R. (2012).
\newblock A survey of actor-critic reinforcement learning: Standard and natural
  policy gradients.
\newblock {\em IEEE Transactions on Systems, Man, and Cybernetics, Part C
  (Applications and Reviews)}, 42(6):1291--1307.

\bibitem[\protect\astroncite{{Gruslys} et~al.}{2017}]{Gruslys2017}
{Gruslys}, A., {Gheshlaghi Azar}, M., {Bellemare}, M.~G., and {Munos}, R.
  (2017).
\newblock {The Reactor: A Sample-Efficient Actor-Critic Architecture}.
\newblock {\em ArXiv e-prints}.

\bibitem[\protect\astroncite{Gu et~al.}{2016a}]{Gu-robotics-2016}
Gu, S., Holly, E., Lillicrap, T., and Levine, S. (2016a).
\newblock Deep reinforcement learning for robotic manipulation with
  asynchronous off-policy updates.
\newblock {\em ArXiv e-prints}.

\bibitem[\protect\astroncite{Gu et~al.}{2017}]{Gu2017QProp}
Gu, S., Lillicrap, T., Ghahramani, Z., Turner, R.~E., and Levine, S. (2017).
\newblock {Q-Prop}: Sample-efficient policy gradient with an off-policy critic.
\newblock In {\em the International Conference on Learning Representations
  (ICLR)}.

\bibitem[\protect\astroncite{{Gu} et~al.}{2017}]{Gu2017Interpolated}
{Gu}, S., {Lillicrap}, T., {Ghahramani}, Z., {Turner}, R.~E., {Sch{\"o}lkopf},
  B., and {Levine}, S. (2017).
\newblock Interpolated policy gradient: Merging on-policy and off-policy
  gradient estimation for deep reinforcement learning.
\newblock In {\em the Annual Conference on Neural Information Processing
  Systems (NIPS)}.

\bibitem[\protect\astroncite{Gu et~al.}{2016b}]{Gu2016}
Gu, S., Lillicrap, T., Sutskever, I., and Levine, S. (2016b).
\newblock Continuous deep {Q}-learning with model-based acceleration.
\newblock In {\em the International Conference on Machine Learning (ICML)}.

\bibitem[\protect\astroncite{{Gulcehre} et~al.}{2016}]{Gulcehre2016}
{Gulcehre}, C., {Chandar}, S., {Cho}, K., and {Bengio}, Y. (2016).
\newblock {Dynamic Neural Turing Machine with Soft and Hard Addressing
  Schemes}.
\newblock {\em ArXiv e-prints}.

\bibitem[\protect\astroncite{{Gulrajani} et~al.}{2017}]{Gulrajani2017}
{Gulrajani}, I., {Ahmed}, F., {Arjovsky}, M., {Dumoulin}, V., and {Courville},
  A. (2017).
\newblock Improved training of wasserstein gans.
\newblock In {\em the Annual Conference on Neural Information Processing
  Systems (NIPS)}.

\bibitem[\protect\astroncite{Gupta et~al.}{2017a}]{Gupta2017}
Gupta, A., Devin, C., Liu, Y., Abbeel, P., and Levine, S. (2017a).
\newblock Learning invariant feature spaces to transfer skills with
  reinforcement learning.
\newblock In {\em the International Conference on Learning Representations
  (ICLR)}.

\bibitem[\protect\astroncite{Gupta et~al.}{2017b}]{Gupta2017CMP}
Gupta, S., Davidson, J., Levine, S., Sukthankar, R., and Malik, J. (2017b).
\newblock Cognitive mapping and planning for visual navigation.
\newblock In {\em the IEEE Conference on Computer Vision and Pattern
  Recognition (CVPR)}.

\bibitem[\protect\astroncite{Guu et~al.}{2017}]{Guu2017}
Guu, K., Pasupat, P., Liu, E.~Z., and Liang, P. (2017).
\newblock From language to programs: Bridging reinforcement learning and
  maximum marginal likelihood.
\newblock In {\em the Association for Computational Linguistics annual meeting
  (ACL)}.

\bibitem[\protect\astroncite{Haarnoja et~al.}{2017}]{Haarnoja2017}
Haarnoja, T., Tang, H., Abbeel, P., and Levine, S. (2017).
\newblock Reinforcement learning with deep energy-based policies.
\newblock In {\em the International Conference on Machine Learning (ICML)}.

\bibitem[\protect\astroncite{Hadfield-Menell
  et~al.}{2016}]{Hadfield-Menell2016}
Hadfield-Menell, D., Dragan, A., Abbeel, P., and Russell, S. (2016).
\newblock Cooperative inverse reinforcement learning.
\newblock In {\em the Annual Conference on Neural Information Processing
  Systems (NIPS)}.

\bibitem[\protect\astroncite{Hadfield-Menell
  et~al.}{2017}]{Hadfield-Menell2017}
Hadfield-Menell, D., Milli, S., Abbeel, P., Russell, S., and Dragan, A. (2017).
\newblock Inverse reward design.
\newblock In {\em the Annual Conference on Neural Information Processing
  Systems (NIPS)}.

\bibitem[\protect\astroncite{Han et~al.}{2016}]{Han2016}
Han, S., Mao, H., and Dally, W.~J. (2016).
\newblock Deep compression: Compressing deep neural networks with pruning,
  trained quantization and {Huffman} coding.
\newblock In {\em the International Conference on Learning Representations
  (ICLR)}.

\bibitem[\protect\astroncite{{Harrison} et~al.}{2017}]{Harrison2017}
{Harrison}, B., {Ehsan}, U., and {Riedl}, M.~O. (2017).
\newblock {Rationalization: A Neural Machine Translation Approach to Generating
  Natural Language Explanations}.
\newblock {\em ArXiv e-prints}.

\bibitem[\protect\astroncite{Harutyunyan et~al.}{2018}]{Harutyunyan2018}
Harutyunyan, A., Vrancx, P., Bacon, P.-L., Precup, D., and Nowe, A. (2018).
\newblock Learning with options that terminate off-policy.
\newblock In {\em the AAAI Conference on Artificial Intelligence (AAAI)}.

\bibitem[\protect\astroncite{Hassabis et~al.}{2017}]{Hassabis2017}
Hassabis, D., Kumaran, D., Summerfield, C., and Botvinick, M. (2017).
\newblock Neuroscience-inspired artificial intelligence.
\newblock {\em Neuron}, 95:245--258.

\bibitem[\protect\astroncite{Hastie et~al.}{2009}]{Hastie2009}
Hastie, T., Tibshirani, R., and Friedman, J. (2009).
\newblock {\em The Elements of Statistical Learning: Data Mining, Inference,
  and Prediction}.
\newblock Springer.

\bibitem[\protect\astroncite{Hausknecht and Stone}{2015}]{Hausknecht2015}
Hausknecht, M. and Stone, P. (2015).
\newblock Deep recurrent {Q}-learning for partially observable {MDPs}.
\newblock In {\em the AAAI Conference on Artificial Intelligence (AAAI)}.

\bibitem[\protect\astroncite{Hausknecht and Stone}{2016}]{Hausknecht2016}
Hausknecht, M. and Stone, P. (2016).
\newblock Deep reinforcement learning in parameterized action space.
\newblock In {\em the International Conference on Learning Representations
  (ICLR)}.

\bibitem[\protect\astroncite{Haykin}{2005}]{Haykin2005}
Haykin, S. (2005).
\newblock Cognitive radio: brain-empowered wireless communications.
\newblock {\em IEEE Journal on Selected Areas in Communications},
  23(2):201--220.

\bibitem[\protect\astroncite{Haykin}{2008}]{Haykin2008}
Haykin, S. (2008).
\newblock {\em Neural Networks and Learning Machines (third edition)}.
\newblock Prentice Hall.

\bibitem[\protect\astroncite{He et~al.}{2016a}]{He2016}
He, D., Xia, Y., Qin, T., Wang, L., Yu, N., Liu, T.-Y., and Ma, W.-Y. (2016a).
\newblock Dual learning for machine translation.
\newblock In {\em the Annual Conference on Neural Information Processing
  Systems (NIPS)}.

\bibitem[\protect\astroncite{He et~al.}{2017}]{He2017}
He, F.~S., Liu, Y., Schwing, A.~G., and Peng, J. (2017).
\newblock Learning to play in a day: Faster deep reinforcement learning by
  optimality tightening.
\newblock In {\em the International Conference on Learning Representations
  (ICLR)}.

\bibitem[\protect\astroncite{He et~al.}{2016b}]{He2016-textgame}
He, J., Chen, J., He, X., Gao, J., Li, L., Deng, L., and Ostendorf, M. (2016b).
\newblock Deep reinforcement learning with a natural language action space.
\newblock In {\em the Association for Computational Linguistics annual meeting
  (ACL)}.

\bibitem[\protect\astroncite{He et~al.}{2016c}]{He2016-actionspace}
He, J., Ostendorf, M., He, X., Chen, J., Gao, J., Li, L., and Deng, L. (2016c).
\newblock Deep reinforcement learning with a combinatorial action space for
  predicting popular reddit threads.
\newblock In {\em Conference on Empirical Methods in Natural Language
  Processing (EMNLP)}.

\bibitem[\protect\astroncite{{He} et~al.}{2017}]{He2017Mask}
{He}, K., {Gkioxari}, G., {Doll{\'a}r}, P., and {Girshick}, R. (2017).
\newblock Mask {R-CNN}.
\newblock In {\em the IEEE International Conference on Computer Vision (ICCV)}.

\bibitem[\protect\astroncite{He et~al.}{2016d}]{He2016-ResNets}
He, K., Zhang, X., Ren, S., and Sun, J. (2016d).
\newblock Deep residual learning for image recognition.
\newblock In {\em the IEEE Conference on Computer Vision and Pattern
  Recognition (CVPR)}.

\bibitem[\protect\astroncite{He et~al.}{2017}]{He2017SRL}
He, L., Lee, K., Lewis, M., and Zettlemoyer, L. (2017).
\newblock Deep semantic role labeling: What works and what's next.
\newblock In {\em the Association for Computational Linguistics annual meeting
  (ACL)}.

\bibitem[\protect\astroncite{He and Deng}{2013}]{He2013}
He, X. and Deng, L. (2013).
\newblock Speech-centric information processing: An optimization-oriented
  approach.
\newblock {\em Proceedings of the IEEE | Vol. 101, No. 5, May 2013},
  101(5):1116--1135.

\bibitem[\protect\astroncite{Heaton et~al.}{2016}]{Heaton2016}
Heaton, J.~B., Polson, N.~G., and Witte, J.~H. (2016).
\newblock Deep learning for finance: deep portfolios.
\newblock {\em Applied Stochastic Models in Business and Industry}.

\bibitem[\protect\astroncite{{Heess} et~al.}{2017}]{Heess2017DPPO}
{Heess}, N., {TB}, D., {Sriram}, S., {Lemmon}, J., {Merel}, J., {Wayne}, G.,
  {Tassa}, Y., {Erez}, T., {Wang}, Z., {Eslami}, A., {Riedmiller}, M., and
  {Silver}, D. (2017).
\newblock {Emergence of Locomotion Behaviours in Rich Environments}.
\newblock {\em ArXiv e-prints}.

\bibitem[\protect\astroncite{Hein et~al.}{2017}]{Hein2017}
Hein, D., Depeweg, S., Tokic, M., Udluft, S., Hentschel, A., Runkler, T.~A.,
  and Sterzing, V. (2017).
\newblock A benchmark environment motivated by industrial control problems.
\newblock In {\em IEEE Symposium on Adaptive Dynamic Programming and
  Reinforcement Learning (IEEE ADPRL'17)}.

\bibitem[\protect\astroncite{Heinrich and Silver}{2016}]{Heinrich2016}
Heinrich, J. and Silver, D. (2016).
\newblock Deep reinforcement learning from self-play in imperfect-information
  games.
\newblock In {\em NIPS 2016 Deep Reinforcement Learning Workshop}.

\bibitem[\protect\astroncite{{Held} et~al.}{2017}]{Held2017}
{Held}, D., {Geng}, X., {Florensa}, C., and {Abbeel}, P. (2017).
\newblock {Automatic Goal Generation for Reinforcement Learning Agents}.
\newblock {\em ArXiv e-prints}.

\bibitem[\protect\astroncite{{Henaff} et~al.}{2017}]{Henaff2017}
{Henaff}, M., {Whitney}, W.~F., and {LeCun}, Y. (2017).
\newblock {Model-Based Planning in Discrete Action Spaces}.
\newblock {\em ArXiv e-prints}.

\bibitem[\protect\astroncite{{Hessel} et~al.}{2018}]{Hessel2018}
{Hessel}, M., {Modayil}, J., {van Hasselt}, H., {Schaul}, T., {Ostrovski}, G.,
  {Dabney}, W., {Horgan}, D., {Piot}, B., {Azar}, M., and {Silver}, D. (2018).
\newblock {Rainbow: Combining Improvements in Deep Reinforcement Learning}.
\newblock In {\em the AAAI Conference on Artificial Intelligence (AAAI)}.

\bibitem[\protect\astroncite{Hester and Stone}{2017}]{Hester2017texplore}
Hester, T. and Stone, P. (2017).
\newblock Intrinsically motivated model learning for developing curious robots.
\newblock {\em Artificial Intelligence}, 247:170--86.

\bibitem[\protect\astroncite{{Hester} et~al.}{2018}]{Hester2018}
{Hester}, T., {Vecerik}, M., {Pietquin}, O., {Lanctot}, M., {Schaul}, T.,
  {Piot}, B., {Horgan}, D., {Quan}, J., {Sendonaris}, A., {Dulac-Arnold}, G.,
  {Osband}, I., {Agapiou}, J., {Leibo}, J.~Z., and {Gruslys}, A. (2018).
\newblock Deep {Q}-learning from demonstrations.
\newblock In {\em the AAAI Conference on Artificial Intelligence (AAAI)}.

\bibitem[\protect\astroncite{Higgins et~al.}{2017}]{Higgins2017betaVAE}
Higgins, I., Matthey, L., Pal, A., Burgess, C., Glorot, X., Botvinick, M.,
  Mohamed, S., and Lerchner, A. (2017).
\newblock $\beta$-{VAE}: Learning basic visual concepts with a constrained
  variational framework.
\newblock In {\em the International Conference on Learning Representations
  (ICLR)}.

\bibitem[\protect\astroncite{Hinton et~al.}{2012}]{Hinton2012}
Hinton, G., Deng, L., Yu, D., Dahl, G.~E., rahman Mohamed, A., Jaitly, N.,
  Senior, A., Vanhoucke, V., Nguyen, P., Sainath, T.~N., , and Kingsbury, B.
  (2012).
\newblock Deep neural networks for acoustic modeling in speech recognition.
\newblock {\em IEEE Signal Processing Magazine}, 82.

\bibitem[\protect\astroncite{Hinton and Salakhutdinov}{2006}]{Hinton2006}
Hinton, G.~E. and Salakhutdinov, R.~R. (2006).
\newblock Reducing the dimensionality of data with neural networks.
\newblock {\em Science}, 313(5786):504--507.

\bibitem[\protect\astroncite{Hirschberg and Manning}{2015}]{Hirschberg2015}
Hirschberg, J. and Manning, C.~D. (2015).
\newblock Advances in natural language processing.
\newblock {\em Science}, 349(6245):261--266.

\bibitem[\protect\astroncite{Ho and Ermon}{2016}]{Ho2016}
Ho, J. and Ermon, S. (2016).
\newblock Generative adversarial imitation learning.
\newblock In {\em the Annual Conference on Neural Information Processing
  Systems (NIPS)}.

\bibitem[\protect\astroncite{Ho et~al.}{2016}]{Ho2016Imitation}
Ho, J., Gupta, J.~K., and Ermon, S. (2016).
\newblock Model-free imitation learning with policy optimization.
\newblock In {\em the International Conference on Machine Learning (ICML)}.

\bibitem[\protect\astroncite{Hochreiter and Schmidhuber}{1997}]{Hochreiter1997}
Hochreiter, S. and Schmidhuber, J. (1997).
\newblock Long short-term memory.
\newblock {\em Neural Computation}, 9:1735--1780.

\bibitem[\protect\astroncite{Hoshen}{2017}]{Hoshen2017}
Hoshen, Y. (2017).
\newblock Vain: Attentional multi-agent predictive modeling.
\newblock In {\em the Annual Conference on Neural Information Processing
  Systems (NIPS)}.

\bibitem[\protect\astroncite{Houthooft et~al.}{2016}]{Houthooft2016}
Houthooft, R., Chen, X., Duan, Y., Schulman, J., Turck, F.~D., and Abbeel, P.
  (2016).
\newblock Vime: Variational information maximizing exploration.
\newblock In {\em the Annual Conference on Neural Information Processing
  Systems (NIPS)}.

\bibitem[\protect\astroncite{{Hu} et~al.}{2017}]{Hu2017}
{Hu}, Z., {Yang}, Z., {Salakhutdinov}, R., and {Xing}, E.~P. (2017).
\newblock {On Unifying Deep Generative Models}.
\newblock {\em ArXiv e-prints}.

\bibitem[\protect\astroncite{Huang et~al.}{2017}]{Huang2017DenseNet}
Huang, G., Liu, Z., Weinberger, K.~Q., and van~der Maaten, L. (2017).
\newblock Densely connected convolutional networks.
\newblock In {\em the IEEE Conference on Computer Vision and Pattern
  Recognition (CVPR)}.

\bibitem[\protect\astroncite{{Huang} et~al.}{2017}]{Huang2017}
{Huang}, S., {Papernot}, N., {Goodfellow}, I., {Duan}, Y., and {Abbeel}, P.
  (2017).
\newblock {Adversarial Attacks on Neural Network Policies}.
\newblock {\em ArXiv e-prints}.

\bibitem[\protect\astroncite{{Huk Park} et~al.}{2016}]{Park2016}
{Huk Park}, D., {Hendricks}, L.~A., {Akata}, Z., {Schiele}, B., {Darrell}, T.,
  and {Rohrbach}, M. (2016).
\newblock {Attentive Explanations: Justifying Decisions and Pointing to the
  Evidence}.
\newblock {\em ArXiv e-prints}.

\bibitem[\protect\astroncite{Hull}{2014}]{Hull06}
Hull, J.~C. (2014).
\newblock {\em Options, Futures and Other Derivatives (9th edition)}.
\newblock Prentice Hall.

\bibitem[\protect\astroncite{Ian J.~Goodfellow}{2015}]{Goodfellow2015}
Ian J.~Goodfellow, Jonathon~Shlens, C.~S. (2015).
\newblock Explaining and harnessing adversarial examples.
\newblock In {\em the International Conference on Learning Representations
  (ICLR)}.

\bibitem[\protect\astroncite{Ioffe and Szegedy}{2015}]{Ioffe2015}
Ioffe, S. and Szegedy, C. (2015).
\newblock Batch normalization: Accelerating deep network training by reducing
  internal covariate shift.
\newblock In {\em the International Conference on Machine Learning (ICML)}.

\bibitem[\protect\astroncite{Islam et~al.}{2017}]{Islam2017}
Islam, R., Henderson, P., Gomrokchi, M., and Precup, D. (2017).
\newblock Reproducibility of benchmarked deep reinforcement learning tasks for
  continuous control.
\newblock In {\em ICML 2017 Reproducibility in Machine Learning Workshop}.

\bibitem[\protect\astroncite{{Jaderberg} et~al.}{2017}]{Jaderberg2017PBT}
{Jaderberg}, M., {Dalibard}, V., {Osindero}, S., {Czarnecki}, W.~M., {Donahue},
  J., {Razavi}, A., {Vinyals}, O., {Green}, T., {Dunning}, I., {Simonyan}, K.,
  {Fernando}, C., and {Kavukcuoglu}, K. (2017).
\newblock {Population Based Training of Neural Networks}.
\newblock {\em ArXiv e-prints}.

\bibitem[\protect\astroncite{Jaderberg et~al.}{2017}]{Jaderberg2017}
Jaderberg, M., Mnih, V., Czarnecki, W., Schaul, T., Leibo, J.~Z., Silver, D.,
  and Kavukcuoglu, K. (2017).
\newblock Reinforcement learning with unsupervised auxiliary tasks.
\newblock In {\em the International Conference on Learning Representations
  (ICLR)}.

\bibitem[\protect\astroncite{Jaderberg et~al.}{2015}]{Jaderberg2015}
Jaderberg, M., Simonyan, K., Zisserman, A., and Kavukcuoglu, K. (2015).
\newblock Spatial transformer networks.
\newblock In {\em the Annual Conference on Neural Information Processing
  Systems (NIPS)}.

\bibitem[\protect\astroncite{James et~al.}{2013}]{James2013}
James, G., Witten, D., Hastie, T., and Tibshirani, R. (2013).
\newblock {\em An Introduction to Statistical Learning with Applications in R}.
\newblock Springer.

\bibitem[\protect\astroncite{Jaques et~al.}{2017}]{Jaques-2017}
Jaques, N., Gu, S., Turner, R.~E., and Eck, D. (2017).
\newblock Tuning recurrent neural networks with reinforcement learning.
\newblock {\em Submitted to Int'l Conference on Learning Representations}.

\bibitem[\protect\astroncite{{Jiang} et~al.}{2016}]{Jiang2016}
{Jiang}, N., {Krishnamurthy}, A., {Agarwal}, A., {Langford}, J., and
  {Schapire}, R.~E. (2016).
\newblock {Contextual Decision Processes with Low Bellman Rank are
  PAC-Learnable}.
\newblock {\em ArXiv e-prints}.

\bibitem[\protect\astroncite{Jie et~al.}{2016}]{Jie2016}
Jie, Z., Liang, X., Feng, J., Jin, X., Lu, W.~F., and Yan, S. (2016).
\newblock Tree-structured reinforcement learning for sequential object
  localization.
\newblock In {\em the Annual Conference on Neural Information Processing
  Systems (NIPS)}.

\bibitem[\protect\astroncite{Johansson et~al.}{2016}]{Johansson2016}
Johansson, F.~D., Shalit, U., and Sontag, D. (2016).
\newblock Learning representations for counterfactual inference.
\newblock In {\em the International Conference on Machine Learning (ICML)}.

\bibitem[\protect\astroncite{Johnson et~al.}{2016}]{Johnson2016}
Johnson, M., Schuster, M., Le, Q.~V., Krikun, M., Wu, Y., Chen, Z., Thorat, N.,
  Vi{\'e}gas, F., Wattenberg, M., Corrado, G., Hughes, M., and Dean, J. (2016).
\newblock {Google's Multilingual Neural Machine Translation System: Enabling
  Zero-Shot Translation}.
\newblock {\em ArXiv e-prints}.

\bibitem[\protect\astroncite{Jordan and Mitchell}{2015}]{Jordan2015}
Jordan, M.~I. and Mitchell, T. (2015).
\newblock Machine learning: Trends, perspectives, and prospects.
\newblock {\em Science}, 349(6245):255--260.

\bibitem[\protect\astroncite{Joulin et~al.}{2017}]{Joulin2017}
Joulin, A., Grave, E., Bojanowski, P., and Mikolov, T. (2017).
\newblock Bag of tricks for efficient text classification.
\newblock In {\em Proceedings of the 15th Conference of the European Chapter of
  the Association for Computational Linguistics (EACL)}.

\bibitem[\protect\astroncite{Jurafsky and Martin}{2017}]{Jurafsky2017}
Jurafsky, D. and Martin, J.~H. (2017).
\newblock {\em Speech and Language Processing (3rd ed. draft)}.
\newblock Prentice Hall.

\bibitem[\protect\astroncite{{Justesen} et~al.}{2017}]{Justesen2017Survey}
{Justesen}, N., {Bontrager}, P., {Togelius}, J., and {Risi}, S. (2017).
\newblock {Deep Learning for Video Game Playing}.
\newblock {\em ArXiv e-prints}.

\bibitem[\protect\astroncite{Justesen and Risi}{2017}]{Justesen2017}
Justesen, N. and Risi, S. (2017).
\newblock Learning macromanagement in starcraft from replays using deep
  learning.
\newblock In {\em IEEE Conference on Computational Intelligence and Games
  (CIG)}.

\bibitem[\protect\astroncite{{Kadlec} et~al.}{2016}]{Kadlec2016}
{Kadlec}, R., {Schmid}, M., {Bajgar}, O., and {Kleindienst}, J. (2016).
\newblock {Text Understanding with the Attention Sum Reader Network}.
\newblock {\em ArXiv e-prints}.

\bibitem[\protect\astroncite{Kaelbling et~al.}{1996}]{Kaelbling1996}
Kaelbling, L.~P., Littman, M.~L., and Moore, A. (1996).
\newblock Reinforcement learning: A survey.
\newblock {\em Journal of Artificial Intelligence Research}, 4:237--285.

\bibitem[\protect\astroncite{Kaiser and Bengio}{2016}]{Kaiser2016}
Kaiser, L. and Bengio, S. (2016).
\newblock Can active memory replace attention?
\newblock In {\em the Annual Conference on Neural Information Processing
  Systems (NIPS)}.

\bibitem[\protect\astroncite{{Kaiser} et~al.}{2017a}]{Kaiser2017OneModel}
{Kaiser}, L., {Gomez}, A.~N., {Shazeer}, N., {Vaswani}, A., {Parmar}, N.,
  {Jones}, L., and {Uszkoreit}, J. (2017a).
\newblock {One Model To Learn Them All}.
\newblock {\em ArXiv e-prints}.

\bibitem[\protect\astroncite{{Kaiser} et~al.}{2017b}]{Kaiser2017}
{Kaiser}, {\L}., {Nachum}, O., {Roy}, A., and {Bengio}, S. (2017b).
\newblock {Learning to Remember Rare Events}.
\newblock In {\em the International Conference on Learning Representations
  (ICLR)}.

\bibitem[\protect\astroncite{Kakade}{2002}]{Kakade2002}
Kakade, S. (2002).
\newblock A natural policy gradient.
\newblock In {\em the Annual Conference on Neural Information Processing
  Systems (NIPS)}.

\bibitem[\protect\astroncite{Kalchbrenner and Blunsom}{2013}]{Kalchbrenner2013}
Kalchbrenner, N. and Blunsom, P. (2013).
\newblock Recurrent continuous translation models.
\newblock In {\em Conference on Empirical Methods in Natural Language
  Processing (EMNLP)}.

\bibitem[\protect\astroncite{Kandasamy et~al.}{2017}]{Kandasamy2017}
Kandasamy, K., Bachrach, Y., Tomioka, R., Tarlow, D., and Carter, D. (2017).
\newblock Batch policy gradient methods for improving neural conversation
  models.
\newblock In {\em the International Conference on Learning Representations
  (ICLR)}.

\bibitem[\protect\astroncite{{Kansky} et~al.}{2017}]{Kansky2017}
{Kansky}, K., {Silver}, T., {M{\'e}ly}, D.~A., {Eldawy}, M.,
  {L{\'a}zaro-Gredilla}, M., {Lou}, X., {Dorfman}, N., {Sidor}, S., {Phoenix},
  S., and {George}, D. (2017).
\newblock Schema networks: Zero-shot transfer with a generative causal model of
  intuitive physics.
\newblock In {\em the International Conference on Machine Learning (ICML)}.

\bibitem[\protect\astroncite{Karpathy et~al.}{2016}]{Karpathy2016}
Karpathy, A., Johnson, J., and Fei-Fei, L. (2016).
\newblock Visualizing and understanding recurrent networks.
\newblock In {\em ICLR 2016 Workshop}.

\bibitem[\protect\astroncite{Kavosh and Littman}{2017}]{Asadi2017}
Kavosh and Littman, M.~L. (2017).
\newblock A new softmax operator for reinforcement learning.
\newblock In {\em the International Conference on Machine Learning (ICML)}.

\bibitem[\protect\astroncite{{Kawaguchi} et~al.}{2017}]{Kawaguchi2017}
{Kawaguchi}, K., {Pack Kaelbling}, L., and {Bengio}, Y. (2017).
\newblock {Generalization in Deep Learning}.
\newblock {\em ArXiv e-prints}.

\bibitem[\protect\astroncite{Kempka et~al.}{2016}]{Kempka2016}
Kempka, M., Wydmuch, M., Runc, G., Toczek, J., and Jas{\'k}owski, W. (2016).
\newblock {ViZDoom: A Doom-based AI research platform for visual reinforcement
  learning}.
\newblock In {\em IEEE Conference on Computational Intelligence and Games}.

\bibitem[\protect\astroncite{Khandani et~al.}{2010}]{Khandani2010}
Khandani, A.~E., Kim, A.~J., and Lo, A.~W. (2010).
\newblock Consumer credit-risk models via machine-learning algorithms.
\newblock {\em Journal of Banking \& Finance}, 34:2767--2787.

\bibitem[\protect\astroncite{Killian et~al.}{2017}]{Killian2017}
Killian, T., Daulton, S., Konidaris, G., and Doshi-Velez, F. (2017).
\newblock Robust and efficient transfer learning with hidden-parameter markov
  decision processes.
\newblock In {\em the Annual Conference on Neural Information Processing
  Systems (NIPS)}.

\bibitem[\protect\astroncite{Kim et~al.}{2014}]{Kim2014}
Kim, B., massoud Farahmand, A., Pineau, J., and Precup, D. (2014).
\newblock Learning from limited demonstrations.
\newblock In {\em the Annual Conference on Neural Information Processing
  Systems (NIPS)}.

\bibitem[\protect\astroncite{{Kingma} et~al.}{2014}]{Kingma2014}
{Kingma}, D.~P., {Rezende}, D.~J., {Mohamed}, S., and {Welling}, M. (2014).
\newblock Semi-supervised learning with deep generative models.
\newblock In {\em the Annual Conference on Neural Information Processing
  Systems (NIPS)}.

\bibitem[\protect\astroncite{{Kirkpatrick} et~al.}{2017}]{Kirkpatrick2017}
{Kirkpatrick}, J., {Pascanu}, R., {Rabinowitz}, N., {Veness}, J., {Desjardins},
  G., {Rusu}, A.~A., {Milan}, K., {Quan}, J., {Ramalho}, T.,
  {Grabska-Barwinska}, A., {Hassabis}, D., {Clopath}, C., {Kumaran}, D., and
  {Hadsell}, R. (2017).
\newblock Overcoming catastrophic forgetting in neural networks.
\newblock {\em PNAS}, 114(13):3521--3526.

\bibitem[\protect\astroncite{{Klambauer} et~al.}{2017}]{Klambauer2017}
{Klambauer}, G., {Unterthiner}, T., {Mayr}, A., and {Hochreiter}, S. (2017).
\newblock {Self-Normalizing Neural Networks}.
\newblock {\em ArXiv e-prints}.

\bibitem[\protect\astroncite{{Klein} et~al.}{2017}]{Klein2017opennmt}
{Klein}, G., {Kim}, Y., {Deng}, Y., {Senellart}, J., and {Rush}, A.~M. (2017).
\newblock {OpenNMT: Open-Source Toolkit for Neural Machine Translation}.
\newblock {\em ArXiv e-prints}.

\bibitem[\protect\astroncite{Kober et~al.}{2013}]{Kober2013}
Kober, J., Bagnell, J.~A., and Peters, J. (2013).
\newblock Reinforcement learning in robotics: A survey.
\newblock {\em International Journal of Robotics Research}, 32(11):1238--1278.

\bibitem[\protect\astroncite{Koch et~al.}{2015}]{Koch2015}
Koch, G., Zemel, R., and Salakhutdinov, R. (2015).
\newblock Siamese neural networks for one-shot image recognition.
\newblock In {\em the International Conference on Machine Learning (ICML)}.

\bibitem[\protect\astroncite{Koh and Liang}{2017}]{Koh2017}
Koh, P.~W. and Liang, P. (2017).
\newblock Understanding black-box predictions via influence functions.
\newblock In {\em the International Conference on Machine Learning (ICML)}.

\bibitem[\protect\astroncite{Kompella et~al.}{2017}]{Kompella2017}
Kompella, V.~R., Stollenga, M., Luciw, M., and Schmidhuber, J. (2017).
\newblock Continual curiosity-driven skill acquisition from high-dimensional
  video inputs for humanoid robots.
\newblock {\em Artificial Intelligence}, 247:313--335.

\bibitem[\protect\astroncite{Kong et~al.}{2017}]{Kong2017}
Kong, X., Xin, B., Wang, Y., and Hua, G. (2017).
\newblock Collaborative deep reinforcement learning for joint object search.
\newblock In {\em the IEEE Conference on Computer Vision and Pattern
  Recognition (CVPR)}.

\bibitem[\protect\astroncite{Kosorok and Moodie}{2015}]{Kosorok2015}
Kosorok, M.~R. and Moodie, E. E.~M. (2015).
\newblock {\em Adaptive Treatment Strategies in Practice: Planning Trials and
  Analyzing Data for Personalized Medicine}.
\newblock ASA-SIAM Series on Statistics and Applied Probability.

\bibitem[\protect\astroncite{Kottur et~al.}{2017}]{Kottur2017}
Kottur, S., Moura, J.~M., Lee, S., and Batra, D. (2017).
\newblock Natural language does not emerge 'naturally' in multi-agent dialog.
\newblock In {\em Conference on Empirical Methods in Natural Language
  Processing (EMNLP)}.

\bibitem[\protect\astroncite{Krakovsky}{2016}]{Krakovsky2016}
Krakovsky, M. (2016).
\newblock Reinforcement renaissance.
\newblock {\em Communications of the ACM}, 59(8):12--14.

\bibitem[\protect\astroncite{Krizhevsky et~al.}{2012}]{Krizhevsky2012}
Krizhevsky, A., Sutskever, I., and Hinton, G.~E. (2012).
\newblock Imagenet classification with deep convolutional neural networks.
\newblock In {\em the Annual Conference on Neural Information Processing
  Systems (NIPS)}.

\bibitem[\protect\astroncite{Krull et~al.}{2017}]{Krull2017}
Krull, A., Brachmann, E., Nowozin, S., Michel, F., Shotton, J., and Rother, C.
  (2017).
\newblock Poseagent: Budget-constrained 6d object pose estimation via
  reinforcement learning.
\newblock In {\em the IEEE Conference on Computer Vision and Pattern
  Recognition (CVPR)}.

\bibitem[\protect\astroncite{Kuhn and Johnson}{2013}]{Kuhn2013}
Kuhn, M. and Johnson, K. (2013).
\newblock {\em Applied Predictive Modeling}.
\newblock Springer.

\bibitem[\protect\astroncite{Kulkarni et~al.}{2016}]{Kulkarni2016}
Kulkarni, T.~D., Narasimhan, K.~R., Saeedi, A., and Tenenbaum, J.~B. (2016).
\newblock Hierarchical deep reinforcement learning: Integrating temporal
  abstraction and intrinsic motivation.
\newblock In {\em the Annual Conference on Neural Information Processing
  Systems (NIPS)}.

\bibitem[\protect\astroncite{Kulkarni et~al.}{2015}]{Kulkarni2015DC-IGN}
Kulkarni, T.~D., Whitney, W., Kohli, P., and Tenenbaum, J.~B. (2015).
\newblock Deep convolutional inverse graphics network.
\newblock In {\em the Annual Conference on Neural Information Processing
  Systems (NIPS)}.

\bibitem[\protect\astroncite{Lagoudakis and Parr}{2003}]{Lagoudakis03}
Lagoudakis, M.~G. and Parr, R. (2003).
\newblock Least-squares policy iteration.
\newblock {\em The Journal of Machine Learning Research}, 4:1107 -- 1149.

\bibitem[\protect\astroncite{Lake et~al.}{2015}]{Lake2015}
Lake, B.~M., Salakhutdinov, R., and Tenenbaum, J.~B. (2015).
\newblock Human-level concept learning through probabilistic program induction.
\newblock {\em Science}, 350(6266):1332--1338.

\bibitem[\protect\astroncite{Lake et~al.}{2016}]{Lake2016}
Lake, B.~M., Ullman, T.~D., Tenenbaum, J.~B., and Gershman, S.~J. (2016).
\newblock Building machines that learn and think like people.
\newblock {\em Behavioral and Brain Sciences}, 24:1--101.

\bibitem[\protect\astroncite{Lamb et~al.}{2016}]{Lamb2016}
Lamb, A., Goyal, A., Zhang, Y., Zhang, S., Courville, A., and Bengio, Y.
  (2016).
\newblock Professor forcing: A new algorithm for training recurrent networks.
\newblock In {\em the Annual Conference on Neural Information Processing
  Systems (NIPS)}.

\bibitem[\protect\astroncite{Lample and Chaplot}{2017}]{Lample2017}
Lample, G. and Chaplot, D.~S. (2017).
\newblock Playing {FPS} games with deep reinforcement learning.
\newblock In {\em the AAAI Conference on Artificial Intelligence (AAAI)}.

\bibitem[\protect\astroncite{Lanctot et~al.}{2017}]{Lanctot2017}
Lanctot, M., Zambaldi, V., Gruslys, A., Lazaridou, A., Tuyls, K., Perolat, J.,
  Silver, D., and Graepel, T. (2017).
\newblock A unified game-theoretic approach to multiagent reinforcement
  learning.
\newblock In {\em the Annual Conference on Neural Information Processing
  Systems (NIPS)}.

\bibitem[\protect\astroncite{Le et~al.}{2012}]{Le2012}
Le, Q.~V., Ranzato, M., Monga, R., Devin, M., Chen, K., Corrado, G.~S., Dean,
  J., and Ng, A.~Y. (2012).
\newblock Building high-level features using large scale unsupervised learning.
\newblock In {\em the International Conference on Machine Learning (ICML)}.

\bibitem[\protect\astroncite{LeCun et~al.}{2015}]{LeCun2015}
LeCun, Y., Bengio, Y., and Hinton, G. (2015).
\newblock Deep learning.
\newblock {\em Nature}, 521:436--444.

\bibitem[\protect\astroncite{Lee et~al.}{2017}]{Lee2017}
Lee, A.~X., Levine, S., and Abbeel, P. (2017).
\newblock Learning visual servoing with deep features and trust region fitted
  {Q}-iteration.
\newblock In {\em the International Conference on Learning Representations
  (ICLR)}.

\bibitem[\protect\astroncite{{Lehman} et~al.}{2017}]{Lehman2017}
{Lehman}, J., {Chen}, J., {Clune}, J., and {Stanley}, K.~O. (2017).
\newblock {Safe Mutations for Deep and Recurrent Neural Networks through Output
  Gradients}.
\newblock {\em ArXiv e-prints}.

\bibitem[\protect\astroncite{Lei et~al.}{2016}]{Lei2016}
Lei, T., Barzilay, R., and Jaakkola, T. (2016).
\newblock Rationalizing neural predictions.
\newblock In {\em Conference on Empirical Methods in Natural Language
  Processing (EMNLP)}.

\bibitem[\protect\astroncite{{Leibo} et~al.}{2018}]{Leibo2018Psychlab}
{Leibo}, J.~Z., {de Masson d'Autume}, C., {Zoran}, D., {Amos}, D., {Beattie},
  C., {Anderson}, K., {Garc{\'{\i}}a Casta{\~n}eda}, A., {Sanchez}, M.,
  {Green}, S., {Gruslys}, A., {Legg}, S., {Hassabis}, D., and {Botvinick},
  M.~M. (2018).
\newblock {Psychlab: A Psychology Laboratory for Deep Reinforcement Learning
  Agents}.
\newblock {\em ArXiv e-prints}.

\bibitem[\protect\astroncite{Leibo et~al.}{2017}]{Leibo2017}
Leibo, J.~Z., Zambaldi, V., Lanctot, M., Marecki, J., and Graepel, T. (2017).
\newblock Multi-agent reinforcement learning in sequential social dilemmas.
\newblock In {\em the International Conference on Autonomous Agents \&
  Multiagent Systems (AAMAS)}.

\bibitem[\protect\astroncite{Levine et~al.}{2016a}]{Levine2016}
Levine, S., Finn, C., Darrell, T., and Abbeel, P. (2016a).
\newblock End-to-end training of deep visuomotor policies.
\newblock {\em The Journal of Machine Learning Research}, 17:1--40.

\bibitem[\protect\astroncite{Levine et~al.}{2016b}]{Levine2016-grasp}
Levine, S., Pastor, P., Krizhevsky, A., and Quillen, D. (2016b).
\newblock {Learning Hand-Eye Coordination for Robotic Grasping with Deep
  Learning and Large-Scale Data Collection}.
\newblock {\em ArXiv e-prints}.

\bibitem[\protect\astroncite{Lewis et~al.}{2017}]{Lewis2017}
Lewis, M., Yarats, D., Dauphin, Y.~N., Parikh, D., and Batra, D. (2017).
\newblock Deal or no deal? end-to-end learning for negotiation dialogues.
\newblock In {\em FAIR}.

\bibitem[\protect\astroncite{Leyton-Brown and Shoham}{2008}]{Leyton-Brown2008}
Leyton-Brown, K. and Shoham, Y. (2008).
\newblock {\em Essentials of Game Theory: A Concise, Multidisciplinary
  Introduction}.
\newblock Morgan \& Claypool Publishers.

\bibitem[\protect\astroncite{Li et~al.}{2017a}]{LiJiwei2017human}
Li, J., Miller, A.~H., Chopra, S., Ranzato, M., and Weston, J. (2017a).
\newblock Dialogue learning with human-in-the-loop.
\newblock In {\em the International Conference on Learning Representations
  (ICLR)}.

\bibitem[\protect\astroncite{Li et~al.}{2017b}]{LiJiwei2017Q}
Li, J., Miller, A.~H., Chopra, S., Ranzato, M., and Weston, J. (2017b).
\newblock Learning through dialogue interactions by asking questions.
\newblock In {\em the International Conference on Learning Representations
  (ICLR)}.

\bibitem[\protect\astroncite{Li et~al.}{2016a}]{LiJiwei2016-fast}
Li, J., Monroe, W., and Jurafsky, D. (2016a).
\newblock {A Simple, Fast Diverse Decoding Algorithm for Neural Generation}.
\newblock {\em ArXiv e-prints}.

\bibitem[\protect\astroncite{Li et~al.}{2016b}]{LiJiwei2016Understanding}
Li, J., Monroe, W., and Jurafsky, D. (2016b).
\newblock {Understanding Neural Networks through Representation Erasure}.
\newblock {\em ArXiv e-prints}.

\bibitem[\protect\astroncite{{Li} et~al.}{2017a}]{LiJiwei2017future}
{Li}, J., {Monroe}, W., and {Jurafsky}, D. (2017a).
\newblock {Learning to Decode for Future Success}.
\newblock {\em ArXiv e-prints}.

\bibitem[\protect\astroncite{Li et~al.}{2016c}]{LiJiwei2016}
Li, J., Monroe, W., Ritter, A., Galley, M., Gao, J., and Jurafsky, D. (2016c).
\newblock Deep reinforcement learning for dialogue generation.
\newblock In {\em Conference on Empirical Methods in Natural Language
  Processing (EMNLP)}.

\bibitem[\protect\astroncite{Li and Malik}{2017}]{Li2017}
Li, K. and Malik, J. (2017).
\newblock Learning to optimize.
\newblock In {\em the International Conference on Learning Representations
  (ICLR)}.

\bibitem[\protect\astroncite{{Li} and {Malik}}{2017}]{Li2017OptNN}
{Li}, K. and {Malik}, J. (2017).
\newblock {Learning to Optimize Neural Nets}.
\newblock {\em ArXiv e-prints}.

\bibitem[\protect\astroncite{Li et~al.}{2010}]{Li2010}
Li, L., Chu, W., Langford, J., and Schapire, R.~E. (2010).
\newblock A contextual-bandit approach to personalized news article
  recommendation.
\newblock In {\em the International World Wide Web Conference (WWW)}.

\bibitem[\protect\astroncite{{Li} et~al.}{2017b}]{Li2017TC}
{Li}, X., {Chen}, Y.-N., {Li}, L., and {Gao}, J. (2017b).
\newblock {End-to-End Task-Completion Neural Dialogue Systems}.
\newblock {\em ArXiv e-prints}.

\bibitem[\protect\astroncite{Li et~al.}{2015}]{Li2016-hybrid}
Li, X., Li, L., Gao, J., He, X., Chen, J., Deng, L., and He, J. (2015).
\newblock {Recurrent Reinforcement Learning: A Hybrid Approach}.
\newblock {\em ArXiv e-prints}.

\bibitem[\protect\astroncite{Li et~al.}{2016d}]{LiXiujun2016}
Li, X., Lipton, Z.~C., Dhingra, B., Li, L., Gao, J., and Chen, Y.-N. (2016d).
\newblock {A User Simulator for Task-Completion Dialogues}.
\newblock {\em ArXiv e-prints}.

\bibitem[\protect\astroncite{Li et~al.}{2017}]{Li2017InfoGAIL}
Li, Y., Song, J., and Ermon, S. (2017).
\newblock Infogail: Interpretable imitation learning from visual
  demonstrations.
\newblock In {\em the Annual Conference on Neural Information Processing
  Systems (NIPS)}.

\bibitem[\protect\astroncite{Li et~al.}{2009}]{Li2009Option}
Li, Y., Szepesv{\'a}ri, C., and Schuurmans, D. (2009).
\newblock Learning exercise policies for {A}merican options.
\newblock In {\em International Conference on Artificial Intelligence and
  Statistics (AISTATS09)}.

\bibitem[\protect\astroncite{Liang et~al.}{2017a}]{Liang2017}
Liang, C., Berant, J., Le, Q., Forbus, K.~D., and Lao, N. (2017a).
\newblock Neural symbolic machines: Learning semantic parsers on freebase with
  weak supervision.
\newblock In {\em the Association for Computational Linguistics annual meeting
  (ACL)}.

\bibitem[\protect\astroncite{Liang et~al.}{2017b}]{Liang2017NSM}
Liang, C., Berant, J., Le, Q., Forbus, K.~D., and Lao, N. (2017b).
\newblock Neural symbolic machines: Learning semantic parsers on freebase with
  weak supervision.
\newblock In {\em the Association for Computational Linguistics annual meeting
  (ACL)}.

\bibitem[\protect\astroncite{Liang et~al.}{2017c}]{Liang2017RayRLLib}
Liang, E., Liaw, R., Nishihara, R., Moritz, P., Fox, R., Gonzalez, J.,
  Goldberg, K., and Stoica, I. (2017c).
\newblock Ray rllib: A composable and scalable reinforcement learning library.
\newblock In {\em NIPS 2017 Deep Reinforcement Learning Symposium}.

\bibitem[\protect\astroncite{Liang et~al.}{2017d}]{Liang2017X}
Liang, X., Lee, L., and Xing, E.~P. (2017d).
\newblock Deep variation-structured reinforcement learning for visual
  relationship and attribute detection.
\newblock In {\em the IEEE Conference on Computer Vision and Pattern
  Recognition (CVPR)}.

\bibitem[\protect\astroncite{Liang et~al.}{2016}]{Liang2016}
Liang, Y., Machado, M.~C., Talvitie, E., and Bowling, M. (2016).
\newblock State of the art control of atari games using shallow reinforcement
  learning.
\newblock In {\em the International Conference on Autonomous Agents \&
  Multiagent Systems (AAMAS)}.

\bibitem[\protect\astroncite{Lillicrap et~al.}{2016}]{Lillicrap2016}
Lillicrap, T.~P., Hunt, J.~J., Pritzel, A., Heess, N., Erez, T., Tassa, Y.,
  Silver, D., and Wierstra, D. (2016).
\newblock Continuous control with deep reinforcement learning.
\newblock In {\em the International Conference on Learning Representations
  (ICLR)}.

\bibitem[\protect\astroncite{Lin}{1992}]{Lin1992}
Lin, L.-J. (1992).
\newblock Self-improving reactive agents based on reinforcement learning,
  planning and teaching.
\newblock {\em Machine learning}, 8(3):293--321.

\bibitem[\protect\astroncite{Lin et~al.}{2017}]{Lin2017}
Lin, Z., Gehring, J., Khalidov, V., and Synnaeve, G. (2017).
\newblock Stardata: A starcraft ai research dataset.
\newblock In {\em AAAI Conference on Artificial Intelligence and Interactive
  Digital Entertainment (AIIDE)}.

\bibitem[\protect\astroncite{Ling et~al.}{2017}]{Ling2017}
Ling, Y., Hasan, S.~A., Datla, V., Qadir, A., Lee, K., Liu, J., and Farri, O.
  (2017).
\newblock Diagnostic inferencing via improving clinical concept extraction with
  deep reinforcement learning: A preliminary study.
\newblock In {\em Machine Learning for Healthcare}.

\bibitem[\protect\astroncite{{Lipton}}{2016}]{Lipton2016Mythos}
{Lipton}, Z.~C. (2016).
\newblock {The Mythos of Model Interpretability}.
\newblock {\em ArXiv e-prints}.

\bibitem[\protect\astroncite{Lipton et~al.}{2016}]{Lipton2016}
Lipton, Z.~C., Gao, J., Li, L., Li, X., Ahmed, F., and Deng, L. (2016).
\newblock {Efficient Exploration for Dialogue Policy Learning with BBQ Networks
  \& Replay Buffer Spiking}.
\newblock {\em ArXiv e-prints}.

\bibitem[\protect\astroncite{Littman}{2015}]{Littman2015}
Littman, M.~L. (2015).
\newblock Reinforcement learning improves behaviour from evaluative feedback.
\newblock {\em Nature}, 521:445--451.

\bibitem[\protect\astroncite{Liu}{2012}]{Liu2012}
Liu, B. (2012).
\newblock {\em Sentiment Analysis and Opinion Mining}.
\newblock Morgan \& Claypool Publishers.

\bibitem[\protect\astroncite{Liu and Tomizuka}{2016}]{Liu2016CoRobot}
Liu, C. and Tomizuka, M. (2016).
\newblock Algorithmic safety measures for intelligent industrial co-robots.
\newblock In {\em IEEE International Conference on Robotics and Automation
  (ICRA)}.

\bibitem[\protect\astroncite{Liu and Tomizuka}{2017}]{Liu2017CoRobot}
Liu, C. and Tomizuka, M. (2017).
\newblock {\em Designing the robot behavior for safe human robot interactions,
  in Trends in Control and Decision-Making for Human-Robot Collaboration
  Systems (Y. Wang and F. Zhang (Eds.))}.
\newblock Springer.

\bibitem[\protect\astroncite{{Liu} et~al.}{2017}]{Liu2017arch}
{Liu}, C., {Zoph}, B., {Shlens}, J., {Hua}, W., {Li}, L.-J., {Fei-Fei}, L.,
  {Yuille}, A., {Huang}, J., and {Murphy}, K. (2017).
\newblock {Progressive Neural Architecture Search}.
\newblock {\em ArXiv e-prints}.

\bibitem[\protect\astroncite{Liu et~al.}{2017}]{Liu2017F}
Liu, F., Li, S., Zhang, L., Zhou, C., Ye, R., Wang, Y., and Lu, J. (2017).
\newblock {3DCNN-DQN-RNN}: A deep reinforcement learning framework for semantic
  parsing of large-scale 3d point clouds.
\newblock In {\em the IEEE International Conference on Computer Vision (ICCV)}.

\bibitem[\protect\astroncite{{Liu} et~al.}{2017}]{Liu2017H}
{Liu}, H., {Simonyan}, K., {Vinyals}, O., {Fernando}, C., and {Kavukcuoglu}, K.
  (2017).
\newblock {Hierarchical Representations for Efficient Architecture Search}.
\newblock {\em ArXiv e-prints}.

\bibitem[\protect\astroncite{Liu et~al.}{2017}]{Liu2017}
Liu, N., Li, Z., Xu, Z., Xu, J., Lin, S., Qiu, Q., Tang, J., and Wang, Y.
  (2017).
\newblock A hierarchical framework of cloud resource allocation and power
  management using deep reinforcement learning.
\newblock In {\em 37th IEEE International Conference on Distributed Computing
  (ICDCS 2017)}.

\bibitem[\protect\astroncite{{Liu} et~al.}{2016}]{Liu2016}
{Liu}, S., {Zhu}, Z., {Ye}, N., {Guadarrama}, S., and {Murphy}, K. (2016).
\newblock {Improved Image Captioning via Policy Gradient optimization of
  SPIDEr}.
\newblock {\em ArXiv e-prints}.

\bibitem[\protect\astroncite{{Liu} et~al.}{2017}]{Liu2017unsupervised}
{Liu}, Y., {Chen}, J., and {Deng}, L. (2017).
\newblock {Unsupervised Sequence Classification using Sequential Output
  Statistics}.
\newblock {\em ArXiv e-prints}.

\bibitem[\protect\astroncite{Liu et~al.}{2014}]{Liu2014Education}
Liu, Y.-E., Mandel, T., Brunskill, E., and Popovi{\'c}, Z. (2014).
\newblock Trading off scientific knowledge and user learning with multi-armed
  bandits.
\newblock In {\em Educational Data Mining (EDM)}.

\bibitem[\protect\astroncite{Lo}{2004}]{Lo04}
Lo, A.~W. (2004).
\newblock {The Adaptive Markets Hypothesis}: Market efficiency from an
  evolutionary perspective.
\newblock {\em Journal of Portfolio Management}, 30:15--29.

\bibitem[\protect\astroncite{Long et~al.}{2015}]{Long2015}
Long, M., Cao, Y., Wang, J., and Jordan, M.~I. (2015).
\newblock Learning transferable features with deep adaptation networks.
\newblock In {\em the International Conference on Machine Learning (ICML)}.

\bibitem[\protect\astroncite{Long et~al.}{2017}]{Long2017}
Long, M., Cao, Z., Wang, J., and Yu, P.~S. (2017).
\newblock Learning multiple tasks with multilinear relationship networks.
\newblock In {\em the Annual Conference on Neural Information Processing
  Systems (NIPS)}.

\bibitem[\protect\astroncite{Long et~al.}{2016}]{Long2016}
Long, M., Zhu, H., Wang, J., and Jordan, M.~I. (2016).
\newblock Unsupervised domain adaptation with residual transfer networks.
\newblock In {\em the Annual Conference on Neural Information Processing
  Systems (NIPS)}.

\bibitem[\protect\astroncite{Longstaff and Schwartz}{2001}]{Longstaff01}
Longstaff, F.~A. and Schwartz, E.~S. (2001).
\newblock {Valuing American options by simulation: a simple least-squares
  approach}.
\newblock {\em The Review of Financial Studies}, 14(1):113--147.

\bibitem[\protect\astroncite{{Loos} et~al.}{2017}]{Loos2017}
{Loos}, S., {Irving}, G., {Szegedy}, C., and {Kaliszyk}, C. (2017).
\newblock {Deep Network Guided Proof Search}.
\newblock {\em ArXiv e-prints}.

\bibitem[\protect\astroncite{{Lopez-Paz} and {Ranzato}}{2017}]{Lopez-Paz2017}
{Lopez-Paz}, D. and {Ranzato}, M. (2017).
\newblock {Gradient Episodic Memory for Continuum Learning}.
\newblock {\em ArXiv e-prints}.

\bibitem[\protect\astroncite{Lowe et~al.}{2017}]{Lowe2017}
Lowe, R., Wu, Y., Tamar, A., Harb, J., Abbeel, P., and Mordatch, I. (2017).
\newblock Multi-agent actor-critic for mixed cooperative-competitive
  environments.
\newblock In {\em the Annual Conference on Neural Information Processing
  Systems (NIPS)}.

\bibitem[\protect\astroncite{{Lu} et~al.}{2016}]{Lu2016}
{Lu}, J., {Xiong}, C., {Parikh}, D., and {Socher}, R. (2016).
\newblock {Knowing When to Look: Adaptive Attention via A Visual Sentinel for
  Image Captioning}.
\newblock {\em ArXiv e-prints}.

\bibitem[\protect\astroncite{Luenberger}{1997}]{Luenberger97}
Luenberger, D.~G. (1997).
\newblock {\em Investment Science}.
\newblock Oxford University Press.

\bibitem[\protect\astroncite{{Luo} et~al.}{2016}]{Luo2016}
{Luo}, Y., {Chiu}, C.-C., {Jaitly}, N., and {Sutskever}, I. (2016).
\newblock {Learning Online Alignments with Continuous Rewards Policy Gradient}.
\newblock {\em ArXiv e-prints}.

\bibitem[\protect\astroncite{Machado et~al.}{2017}]{Machado2017}
Machado, M.~C., Bellemare, M.~G., and Bowling, M. (2017).
\newblock A {Laplacian} framework for option discovery in reinforcement
  learning.
\newblock In {\em the International Conference on Machine Learning (ICML)}.

\bibitem[\protect\astroncite{{Machado} et~al.}{2017}]{Machado2017ALE}
{Machado}, M.~C., {Bellemare}, M.~G., {Talvitie}, E., {Veness}, J.,
  {Hausknecht}, M., and {Bowling}, M. (2017).
\newblock {Revisiting the Arcade Learning Environment: Evaluation Protocols and
  Open Problems for General Agents}.
\newblock {\em ArXiv e-prints}.

\bibitem[\protect\astroncite{{Madry} et~al.}{2017}]{Madry2017}
{Madry}, A., {Makelov}, A., {Schmidt}, L., {Tsipras}, D., and {Vladu}, A.
  (2017).
\newblock {Towards Deep Learning Models Resistant to Adversarial Attacks}.
\newblock {\em ArXiv e-prints}.

\bibitem[\protect\astroncite{{Mahler} et~al.}{2017}]{Mahler2017}
{Mahler}, J., {Liang}, J., {Niyaz}, S., {Laskey}, M., {Doan}, R., {Liu}, X.,
  {Aparicio Ojea}, J., and {Goldberg}, K. (2017).
\newblock {Dex-Net} 2.0: Deep learning to plan robust grasps with synthetic
  point clouds and analytic grasp metrics.
\newblock In {\em Robotics: Science and Systems (RSS)}.

\bibitem[\protect\astroncite{Mahmood et~al.}{2014}]{Mahmood2014}
Mahmood, A.~R., van Hasselt, H., and Sutton, R.~S. (2014).
\newblock Weighted importance sampling for off-policy learning with linear
  function approximation.
\newblock In {\em the Annual Conference on Neural Information Processing
  Systems (NIPS)}.

\bibitem[\protect\astroncite{Mandel et~al.}{2014}]{Mandel2014}
Mandel, T., Liu, Y.~E., Levine, S., Brunskill, E., and Popovi{\'c}, Z. (2014).
\newblock Offline policy evaluation across representations with applications to
  educational games.
\newblock In {\em the International Conference on Autonomous Agents \&
  Multiagent Systems (AAMAS)}.

\bibitem[\protect\astroncite{Manning et~al.}{2008}]{Manning2008}
Manning, C.~D., Raghavan, P., and Sch{\"u}tze, H. (2008).
\newblock {\em Introduction to Information Retrieval}.
\newblock Cambridge University Press.

\bibitem[\protect\astroncite{Mannion et~al.}{2016}]{Mannion2016}
Mannion, P., Duggan, J., and Howley, E. (2016).
\newblock An experimental review of reinforcement learning algorithms for
  adaptive traffic signal control.
\newblock {\em Autonomic Road Transport Support Systems, edited by McCluskey,
  T., Kotsialos, A., M{\"u}ller, J., Kl{\"u}gl, F., Rana, O., and Schumann R.,
  Springer International Publishing, Cham}, pages 47--66.

\bibitem[\protect\astroncite{Mao et~al.}{2016}]{Mao2016RM}
Mao, H., Alizadeh, M., Menache, I., and Kandula, S. (2016).
\newblock Resource management with deep reinforcement learning.
\newblock In {\em ACM Workshop on Hot Topics in Networks (HotNets)}.

\bibitem[\protect\astroncite{{Mao} et~al.}{2016}]{Mao2016}
{Mao}, X., {Li}, Q., {Xie}, H., {Lau}, R.~Y.~K., and {Wang}, Z. (2016).
\newblock {Least Squares Generative Adversarial Networks}.
\newblock {\em ArXiv e-prints}.

\bibitem[\protect\astroncite{Mathe et~al.}{2016}]{Mathe2016}
Mathe, S., Pirinen, A., and Sminchisescu, C. (2016).
\newblock Reinforcement learning for visual object detection.
\newblock In {\em the IEEE Conference on Computer Vision and Pattern
  Recognition (CVPR)}.

\bibitem[\protect\astroncite{{Matiisen} et~al.}{2017}]{Matiisen2017}
{Matiisen}, T., {Oliver}, A., {Cohen}, T., and {Schulman}, J. (2017).
\newblock {Teacher-Student Curriculum Learning}.
\newblock {\em ArXiv e-prints}.

\bibitem[\protect\astroncite{Maurer et~al.}{2016}]{Maurer2016}
Maurer, A., Pontil, M., and Romera-Paredes, B. (2016).
\newblock The benefit of multitask representation learning.
\newblock {\em The Journal of Machine Learning Research}, 17(81):1--32.

\bibitem[\protect\astroncite{McAllister and Rasmussen}{2017}]{McAllister2017}
McAllister, R. and Rasmussen, C.~E. (2017).
\newblock Data-efficient reinforcement learning in continuous-state {POMDPs}.
\newblock In {\em the Annual Conference on Neural Information Processing
  Systems (NIPS)}.

\bibitem[\protect\astroncite{{McCann} et~al.}{2017}]{McCann2017}
{McCann}, B., {Bradbury}, J., {Xiong}, C., and {Socher}, R. (2017).
\newblock {Learned in Translation: Contextualized Word Vectors}.
\newblock {\em ArXiv e-prints}.

\bibitem[\protect\astroncite{{Melis} et~al.}{2017}]{Melis2017}
{Melis}, G., {Dyer}, C., and {Blunsom}, P. (2017).
\newblock {On the State of the Art of Evaluation in Neural Language Models}.
\newblock {\em ArXiv e-prints}.

\bibitem[\protect\astroncite{{Merel} et~al.}{2017}]{Merel2017}
{Merel}, J., {Tassa}, Y., {TB}, D., {Srinivasan}, S., {Lemmon}, J., {Wang}, Z.,
  {Wayne}, G., and {Heess}, N. (2017).
\newblock {Learning human behaviors from motion capture by adversarial
  imitation}.
\newblock {\em ArXiv e-prints}.

\bibitem[\protect\astroncite{Mesnil et~al.}{2015}]{Mesnil2015}
Mesnil, G., Dauphin, Y., Yao, K., Bengio, Y., Deng, L., He, X., Heck, L., Tur,
  G., Hakkani-T{\"u}r, D., Yu, D., and Zweig, G. (2015).
\newblock Using recurrent neural networks for slot filling in spoken language
  understanding.
\newblock {\em IEEE/ACM Transactions on Audio, Speech, and Language
  Processing}, 23(3):530--539.

\bibitem[\protect\astroncite{Mestres et~al.}{2016}]{Mestres2016}
Mestres, A., Rodriguez-Natal, A., Carner, J., Barlet-Ros, P., Alarc{\'o}n, E.,
  Sol{\'e}, M., Munt{\'e}s, V., Meyer, D., Barkai, S., Hibbett, M.~J., Estrada,
  G., Ma{\`r}uf, K., Coras, F., Ermagan, V., Latapie, H., Cassar, C., Evans,
  J., Maino, F., Walrand, J., and Cabellos, A. (2016).
\newblock {Knowledge-Defined Networking}.
\newblock {\em ArXiv e-prints}.

\bibitem[\protect\astroncite{Mhamdi et~al.}{2017}]{MahdiElMhamdi2017}
Mhamdi, E. M.~E., Guerraoui, R., Hendrikx, H., and Maurer, A. (2017).
\newblock Dynamic safe interruptibility for decentralized multi-agent
  reinforcement learning.
\newblock In {\em the Annual Conference on Neural Information Processing
  Systems (NIPS)}.

\bibitem[\protect\astroncite{Mikolov et~al.}{2013}]{Mikolov2013}
Mikolov, T., Chen, K., Corrado, G., and Dean, J. (2013).
\newblock Efficient estimation of word representations in vector space.
\newblock In {\em the International Conference on Learning Representations
  (ICLR)}.

\bibitem[\protect\astroncite{{Mikolov} et~al.}{2017}]{Mikolov2017}
{Mikolov}, T., {Grave}, E., {Bojanowski}, P., {Puhrsch}, C., and {Joulin}, A.
  (2017).
\newblock {Advances in Pre-Training Distributed Word Representations}.
\newblock {\em ArXiv e-prints}.

\bibitem[\protect\astroncite{{Miller}}{2017}]{Miller2017}
{Miller}, T. (2017).
\newblock {Explanation in Artificial Intelligence: Insights from the Social
  Sciences}.
\newblock {\em ArXiv e-prints}.

\bibitem[\protect\astroncite{Miotto et~al.}{2017}]{Miotto2017}
Miotto, R., Wang, F., Wang, S., Jiang, X., and Dudley, J.~T. (2017).
\newblock Deep learning for healthcare: review, opportunities and challenges.
\newblock {\em Briefings in Bioinformatics}, pages 1--11.

\bibitem[\protect\astroncite{Mirhoseini et~al.}{2017}]{Mirhoseini2017}
Mirhoseini, A., Pham, H., Le, Q.~V., Steiner, B., Larsen, R., Zhou, Y., Kumar,
  N., and Mohammad~Norouzi, Samy~Bengio, J.~D. (2017).
\newblock Device placement optimization with reinforcement learning.
\newblock In {\em the International Conference on Machine Learning (ICML)}.

\bibitem[\protect\astroncite{Mirowski et~al.}{2017}]{Mirowski2017}
Mirowski, P., Pascanu, R., Viola, F., Soyer, H., Ballard, A., Banino, A.,
  Denil, M., Goroshin, R., Sifre, L., Kavukcuoglu, K., Kumaran, D., and
  Hadsell, R. (2017).
\newblock Learning to navigate in complex environments.
\newblock In {\em the International Conference on Learning Representations
  (ICLR)}.

\bibitem[\protect\astroncite{{Mitra} and {Craswell}}{2017}]{Mitra2017}
{Mitra}, B. and {Craswell}, N. (2017).
\newblock {Neural Models for Information Retrieval}.
\newblock {\em ArXiv e-prints}.

\bibitem[\protect\astroncite{Mnih et~al.}{2016}]{Mnih-A3C-2016}
Mnih, V., Badia, A.~P., Mirza, M., Graves, A., Harley, T., Lillicrap, T.~P.,
  Silver, D., and Kavukcuoglu, K. (2016).
\newblock Asynchronous methods for deep reinforcement learning.
\newblock In {\em the International Conference on Machine Learning (ICML)}.

\bibitem[\protect\astroncite{Mnih et~al.}{2014}]{Mnih-attention-2014}
Mnih, V., Heess, N., Graves, A., and Kavukcuoglu, K. (2014).
\newblock Recurrent models of visual attention.
\newblock In {\em the Annual Conference on Neural Information Processing
  Systems (NIPS)}.

\bibitem[\protect\astroncite{Mnih et~al.}{2015}]{Mnih-DQN-2015}
Mnih, V., Kavukcuoglu, K., Silver, D., Rusu, A.~A., Veness, J., Bellemare,
  M.~G., Graves, A., Riedmiller, M., Fidjeland, A.~K., Ostrovski, G., Petersen,
  S., Beattie, C., Sadik, A., Antonoglou, I., King, H., Kumaran, D., Wierstra,
  D., Legg, S., and Hassabis, D. (2015).
\newblock Human-level control through deep reinforcement learning.
\newblock {\em Nature}, 518(7540):529--533.

\bibitem[\protect\astroncite{Mo et~al.}{2016}]{Mo2016}
Mo, K., Li, S., Zhang, Y., Li, J., and Yang, Q. (2016).
\newblock {Personalizing a Dialogue System with Transfer Learning}.
\newblock {\em ArXiv e-prints}.

\bibitem[\protect\astroncite{Monroe}{2017}]{Monroe2017}
Monroe, D. (2017).
\newblock Deep learning takes on translation.
\newblock {\em Communications of the ACM}, 60(6):12--14.

\bibitem[\protect\astroncite{Moody and Saffell}{2001}]{Moody01}
Moody, J. and Saffell, M. (2001).
\newblock Learning to trade via direct reinforcement.
\newblock {\em IEEE Transactions on Neural Networks}, 12(4):875--889.

\bibitem[\protect\astroncite{Morav{\v c}{\' i}k et~al.}{2017}]{Moravcik2017}
Morav{\v c}{\' i}k, M., Schmid, M., Burch, N., Lis{\'y}, V., Morrill, D., Bard,
  N., Davis, T., Waugh, K., Johanson, M., and Bowling, M. (2017).
\newblock Deepstack: Expert-level artificial intelligence in heads-up no-limit
  poker.
\newblock {\em Science}.

\bibitem[\protect\astroncite{M{\"u}ller}{2002}]{Muller2002}
M{\"u}ller, M. (2002).
\newblock Computer go.
\newblock {\em Artificial Intelligence}, 134(1-2):145--179.

\bibitem[\protect\astroncite{Munos et~al.}{2016}]{Remi2016}
Munos, R., Stepleton, T., Harutyunyan, A., and Bellemare, M.~G. (2016).
\newblock Safe and efficient off-policy reinforcement learning.
\newblock In {\em the Annual Conference on Neural Information Processing
  Systems (NIPS)}.

\bibitem[\protect\astroncite{Murphy}{2012}]{Murphy2012}
Murphy, K.~P. (2012).
\newblock {\em Machine Learning: A Probabilistic Perspective}.
\newblock The MIT Press.

\bibitem[\protect\astroncite{Nachum et~al.}{2017}]{Nachum2017}
Nachum, O., Norouzi, M., and Schuurmans, D. (2017).
\newblock Improving policy gradient by exploring under-appreciated rewards.
\newblock In {\em the International Conference on Learning Representations
  (ICLR)}.

\bibitem[\protect\astroncite{{Nachum} et~al.}{2017}]{Nachum2017Gap}
{Nachum}, O., {Norouzi}, M., {Xu}, K., and {Schuurmans}, D. (2017).
\newblock Bridging the gap between value and policy based reinforcement
  learning.
\newblock In {\em the Annual Conference on Neural Information Processing
  Systems (NIPS)}.

\bibitem[\protect\astroncite{{Nair} et~al.}{2015}]{Nair2015}
{Nair}, A., {Srinivasan}, P., {Blackwell}, S., {Alcicek}, C., {Fearon}, R., {De
  Maria}, A., {Panneershelvam}, V., {Suleyman}, M., {Beattie}, C., {Petersen},
  S., {Legg}, S., {Mnih}, V., {Kavukcuoglu}, K., and {Silver}, D. (2015).
\newblock Massively parallel methods for deep reinforcement learning.
\newblock In {\em ICML 2015 Deep Learning Workshop}.

\bibitem[\protect\astroncite{Narasimhan et~al.}{2015}]{Narasimhan2015}
Narasimhan, K., Kulkarni, T., and Barzilay, R. (2015).
\newblock Language understanding for text-based games using deep reinforcement
  learning.
\newblock In {\em Conference on Empirical Methods in Natural Language
  Processing (EMNLP)}.

\bibitem[\protect\astroncite{Narasimhan et~al.}{2016}]{Narasimhan2016}
Narasimhan, K., Yala, A., and Barzilay, R. (2016).
\newblock Improving information extraction by acquiring external evidence with
  reinforcement learning.
\newblock In {\em Conference on Empirical Methods in Natural Language
  Processing (EMNLP)}.

\bibitem[\protect\astroncite{Nedi{\'c} and Bertsekas}{2003}]{Nedic2003}
Nedi{\'c}, A. and Bertsekas, D.~P. (2003).
\newblock Least squares policy evaluation algorithms with linear function
  approximation.
\newblock {\em Discrete Event Dynamic Systems: Theory and Applications},
  13:79--110.

\bibitem[\protect\astroncite{Neuneier}{1997}]{Neuneier1997}
Neuneier, R. (1997).
\newblock Enhancing q-learning for optimal asset allocation.
\newblock In {\em the Annual Conference on Neural Information Processing
  Systems (NIPS)}.

\bibitem[\protect\astroncite{{Neyshabur} et~al.}{2017}]{Neyshabur2017}
{Neyshabur}, B., {Tomioka}, R., {Salakhutdinov}, R., and {Srebro}, N. (2017).
\newblock {Geometry of Optimization and Implicit Regularization in Deep
  Learning}.
\newblock {\em ArXiv e-prints}.

\bibitem[\protect\astroncite{Ng and Russell}{2000}]{Ng2000}
Ng, A. and Russell, S. (2000).
\newblock Algorithms for inverse reinforcement learning.
\newblock In {\em the International Conference on Machine Learning (ICML)}.

\bibitem[\protect\astroncite{Nogueira and Cho}{2016}]{Nogueira2016}
Nogueira, R. and Cho, K. (2016).
\newblock {End-to-End Goal-Driven Web Navigation}.
\newblock {\em ArXiv e-prints}.

\bibitem[\protect\astroncite{{Nogueira} and {Cho}}{2017}]{Nogueira2017}
{Nogueira}, R. and {Cho}, K. (2017).
\newblock {Task-Oriented Query Reformulation with Reinforcement Learning}.
\newblock {\em ArXiv e-prints}.

\bibitem[\protect\astroncite{O'Donoghue et~al.}{2017}]{ODonoghue2017}
O'Donoghue, B., Munos, R., Kavukcuoglu, K., and Mnih, V. (2017).
\newblock {PGQ}: Combining policy gradient and {Q}-learning.
\newblock In {\em the International Conference on Learning Representations
  (ICLR)}.

\bibitem[\protect\astroncite{O'Donovan et~al.}{2015}]{ODonovan2015}
O'Donovan, P., Leahy, K., Bruton, K., and O'Sullivan, D. T.~J. (2015).
\newblock Big data in manufacturing: a systematic mapping study.
\newblock {\em Journal of Big Data}, 2(20).

\bibitem[\protect\astroncite{Oh et~al.}{2016}]{Oh2016}
Oh, J., Chockalingam, V., Singh, S., and Lee, H. (2016).
\newblock Control of memory, active perception, and action in minecraft.
\newblock In {\em the International Conference on Machine Learning (ICML)}.

\bibitem[\protect\astroncite{Oh et~al.}{2015}]{Oh2015}
Oh, J., Guo, X., Lee, H., Lewis, R., and Singh, S. (2015).
\newblock Action-conditional video prediction using deep networks in atari
  games.
\newblock In {\em the Annual Conference on Neural Information Processing
  Systems (NIPS)}.

\bibitem[\protect\astroncite{{Oh} et~al.}{2017}]{Oh2017VPN}
{Oh}, J., {Singh}, S., and {Lee}, H. (2017).
\newblock Value prediction network.
\newblock In {\em the Annual Conference on Neural Information Processing
  Systems (NIPS)}.

\bibitem[\protect\astroncite{Omidshafiei et~al.}{2017}]{Omidshafiei2017}
Omidshafiei, S., Pazis, J., Amato, C., How, J.~P., and Vian, J. (2017).
\newblock Deep decentralized multi-task multi-agent reinforcement learning
  under partial observability.
\newblock In {\em the International Conference on Machine Learning (ICML)}.

\bibitem[\protect\astroncite{Onta{\~n}{\'o}n et~al.}{2013}]{Ontanon2013}
Onta{\~n}{\'o}n, S., Synnaeve, G., Uriarte, A., Richoux, F., Churchill, D., and
  Preuss, M. (2013).
\newblock A survey of real-time strategy game ai research and competition in
  starcraft.
\newblock {\em IEEE Transactions on Computational Intelligence and AI in
  Games}, 5(4):293--311.

\bibitem[\protect\astroncite{Oquab et~al.}{2015}]{Oquab2015}
Oquab, M., Bottou, L., Laptev, I., and Sivic, J. (2015).
\newblock Is object localization for free? -- weakly-supervised learning with
  convolutional neural networks.
\newblock In {\em the IEEE Conference on Computer Vision and Pattern
  Recognition (CVPR)}.

\bibitem[\protect\astroncite{Osband et~al.}{2016}]{Osband2016}
Osband, I., Blundell, C., Pritzel, A., and Roy, B.~V. (2016).
\newblock Deep exploration via bootstrapped {DQN}.
\newblock In {\em the Annual Conference on Neural Information Processing
  Systems (NIPS)}.

\bibitem[\protect\astroncite{{Ostrovski} et~al.}{2017}]{Ostrovski2017}
{Ostrovski}, G., {Bellemare}, M.~G., {van den Oord}, A., and {Munos}, R.
  (2017).
\newblock {Count-Based Exploration with Neural Density Models}.
\newblock {\em ArXiv e-prints}.

\bibitem[\protect\astroncite{Pan and Yang}{2010}]{Pan2010}
Pan, S.~J. and Yang, Q. (2010).
\newblock A survey on transfer learning.
\newblock {\em IEEE Transactions on Knowledge and Data Engineering},
  22(10):1345 -- 1359.

\bibitem[\protect\astroncite{Papernot et~al.}{2017}]{Papernot2017}
Papernot, N., Abadi, M., Erlingsson, {\'U}., Goodfellow, I., and Talwar, K.
  (2017).
\newblock Semi-supervised knowledge transfer for deep learning from private
  training data.
\newblock In {\em the International Conference on Learning Representations
  (ICLR)}.

\bibitem[\protect\astroncite{{Papernot} et~al.}{2016}]{Papernot2016cleverhans}
{Papernot}, N., {Goodfellow}, I., {Sheatsley}, R., {Feinman}, R., and
  {McDaniel}, P. (2016).
\newblock {cleverhans v1.0.0: an adversarial machine learning library}.
\newblock {\em ArXiv e-prints}.

\bibitem[\protect\astroncite{Parisotto et~al.}{2016}]{Parisotto2016}
Parisotto, E., Ba, J.~L., and Salakhutdinov, R. (2016).
\newblock Actor-mimic: Deep multitask and transfer reinforcement learning.
\newblock In {\em the International Conference on Learning Representations
  (ICLR)}.

\bibitem[\protect\astroncite{Parisotto et~al.}{2017}]{Parisotto2017}
Parisotto, E., rahman Mohamed, A., Singh, R., Li, L., Zhou, D., and Kohli, P.
  (2017).
\newblock Neuro-symbolic program synthesis.
\newblock In {\em the International Conference on Learning Representations
  (ICLR)}.

\bibitem[\protect\astroncite{Pasunuru and Bansal}{2017}]{Pasunuru2017}
Pasunuru, R. and Bansal, M. (2017).
\newblock Reinforced video captioning with entailment rewards.
\newblock In {\em Conference on Empirical Methods in Natural Language
  Processing (EMNLP)}.

\bibitem[\protect\astroncite{{Paulus} et~al.}{2017}]{Paulus2017}
{Paulus}, R., {Xiong}, C., and {Socher}, R. (2017).
\newblock {A Deep Reinforced Model for Abstractive Summarization}.
\newblock {\em ArXiv e-prints}.

\bibitem[\protect\astroncite{{Pearl}}{2018}]{Pearl2018}
{Pearl}, J. (2018).
\newblock {Theoretical Impediments to Machine Learning With Seven Sparks from
  the Causal Revolution}.
\newblock {\em ArXiv e-prints}.

\bibitem[\protect\astroncite{{Pei} et~al.}{2017}]{Pei2017}
{Pei}, K., {Cao}, Y., {Yang}, J., and {Jana}, S. (2017).
\newblock {DeepXplore: Automated Whitebox Testing of Deep Learning Systems}.
\newblock {\em ArXiv e-prints}.

\bibitem[\protect\astroncite{{Peng} et~al.}{2017a}]{Peng2017Dialogue}
{Peng}, B., {Li}, X., {Li}, L., {Gao}, J., {Celikyilmaz}, A., {Lee}, S., and
  {Wong}, K.-F. (2017a).
\newblock Composite task-completion dialogue system via hierarchical deep
  reinforcement learning.
\newblock In {\em Conference on Empirical Methods in Natural Language
  Processing (EMNLP)}.

\bibitem[\protect\astroncite{{Peng} et~al.}{2017b}]{Peng2017}
{Peng}, P., {Yuan}, Q., {Wen}, Y., {Yang}, Y., {Tang}, Z., {Long}, H., and
  {Wang}, J. (2017b).
\newblock {Multiagent Bidirectionally-Coordinated Nets for Learning to Play
  StarCraft Combat Games}.
\newblock {\em ArXiv e-prints}.

\bibitem[\protect\astroncite{P{\'e}rez-D'Arpino and Shah}{2017}]{Perez2017}
P{\'e}rez-D'Arpino, C. and Shah, J.~A. (2017).
\newblock C-learn: Learning geometric constraints from demonstrations for
  multi-step manipulation in shared autonomy.
\newblock In {\em IEEE International Conference on Robotics and Automation
  (ICRA)}.

\bibitem[\protect\astroncite{Perolat et~al.}{2017}]{Perolat2017}
Perolat, J., Leibo, J.~Z., Zambaldi, V., Beattie, C., Tuyls, K., and Graepel,
  T. (2017).
\newblock A multi-agent reinforcement learning model of common-pool resource
  appropriation.
\newblock In {\em the Annual Conference on Neural Information Processing
  Systems (NIPS)}.

\bibitem[\protect\astroncite{Peters and Neumann}{2015}]{Peters2015}
Peters, J. and Neumann, G. (2015).
\newblock Policy search: Methods and applications.
\newblock {\em ICML 2015 Tutorial}.

\bibitem[\protect\astroncite{{Petroski Such} et~al.}{2017}]{Petroski2017}
{Petroski Such}, F., {Madhavan}, V., {Conti}, E., {Lehman}, J., {Stanley},
  K.~O., and {Clune}, J. (2017).
\newblock {Deep Neuroevolution: Genetic Algorithms Are a Competitive
  Alternative for Training Deep Neural Networks for Reinforcement Learning}.
\newblock {\em ArXiv e-prints}.

\bibitem[\protect\astroncite{Pfau and Vinyals}{2016}]{Pfau2016}
Pfau, D. and Vinyals, O. (2016).
\newblock {Connecting Generative Adversarial Networks and Actor-Critic
  Methods}.
\newblock {\em ArXiv e-prints}.

\bibitem[\protect\astroncite{{Phua} et~al.}{2010}]{Phua2010}
{Phua}, C., {Lee}, V., {Smith}, K., and {Gayler}, R. (2010).
\newblock {A Comprehensive Survey of Data Mining-based Fraud Detection
  Research}.
\newblock {\em ArXiv e-prints}.

\bibitem[\protect\astroncite{{Popov} et~al.}{2017}]{Popov2017}
{Popov}, I., {Heess}, N., {Lillicrap}, T., {Hafner}, R., {Barth-Maron}, G.,
  {Vecerik}, M., {Lampe}, T., {Tassa}, Y., {Erez}, T., and {Riedmiller}, M.
  (2017).
\newblock {Data-efficient Deep Reinforcement Learning for Dexterous
  Manipulation}.
\newblock {\em ArXiv e-prints}.

\bibitem[\protect\astroncite{Powell}{2011}]{Powell11}
Powell, W.~B. (2011).
\newblock {\em Approximate Dynamic Programming: Solving the curses of
  dimensionality (2nd Edition)}.
\newblock John Wiley and Sons.

\bibitem[\protect\astroncite{Prashanth et~al.}{2016}]{Prashanth2016}
Prashanth, L., Jie, C., Fu, M., Marcus, S., and Szepes{\'a}ri, C. (2016).
\newblock Cumulative prospect theory meets reinforcement learning: Prediction
  and control.
\newblock In {\em the International Conference on Machine Learning (ICML)}.

\bibitem[\protect\astroncite{Preuveneers and
  Ilie-Zudor}{2017}]{Preuveneers2017}
Preuveneers, D. and Ilie-Zudor, E. (2017).
\newblock The intelligent industry of the future: A survey on emerging trends,
  research challenges and opportunities in industry 4.0.
\newblock {\em Journal of Ambient Intelligence and Smart Environments},
  9(3):287--298.

\bibitem[\protect\astroncite{{Pritzel} et~al.}{2017}]{Pritzel2017}
{Pritzel}, A., {Uria}, B., {Srinivasan}, S., {Puigdom{\`e}nech}, A., {Vinyals},
  O., {Hassabis}, D., {Wierstra}, D., and {Blundell}, C. (2017).
\newblock {Neural Episodic Control}.
\newblock {\em ArXiv e-prints}.

\bibitem[\protect\astroncite{Provost and Fawcett}{2013}]{Provost2013}
Provost, F. and Fawcett, T. (2013).
\newblock {\em Data Science for Business}.
\newblock O'Reilly Media.

\bibitem[\protect\astroncite{Puterman}{2005}]{Puterman05}
Puterman, M.~L. (2005).
\newblock {\em Markov decision processes : discrete stochastic dynamic
  programming}.
\newblock Wiley-Interscience.

\bibitem[\protect\astroncite{{Radford} et~al.}{2017}]{Radford2017}
{Radford}, A., {Jozefowicz}, R., and {Sutskever}, I. (2017).
\newblock {Learning to Generate Reviews and Discovering Sentiment}.
\newblock {\em ArXiv e-prints}.

\bibitem[\protect\astroncite{{Raghu} et~al.}{2016}]{Raghu2016}
{Raghu}, M., {Poole}, B., {Kleinberg}, J., {Ganguli}, S., and {Sohl-Dickstein},
  J. (2016).
\newblock {Survey of Expressivity in Deep Neural Networks}.
\newblock {\em ArXiv e-prints}.

\bibitem[\protect\astroncite{Rahimi and Recht}{2007}]{Rahimi2007}
Rahimi, A. and Recht, B. (2007).
\newblock Random features for large-scale kernel machines.
\newblock In {\em the Annual Conference on Neural Information Processing
  Systems (NIPS)}.

\bibitem[\protect\astroncite{Rajendran et~al.}{2017}]{Rajendran2017}
Rajendran, J., Lakshminarayanan, A., Khapra, M.~M., P, P., and Ravindran, B.
  (2017).
\newblock Attend, adapt and transfer: Attentive deep architecture for adaptive
  transfer from multiple sources in the same domain.
\newblock {\em the International Conference on Learning Representations
  (ICLR)}.

\bibitem[\protect\astroncite{Ranzato et~al.}{2016}]{Ranzato2016}
Ranzato, M., Chopra, S., Auli, M., and Zaremba, W. (2016).
\newblock Sequence level training with recurrent neural networks.
\newblock In {\em the International Conference on Learning Representations
  (ICLR)}.

\bibitem[\protect\astroncite{Rao et~al.}{2017}]{Rao2017}
Rao, Y., Lu, J., and Zhou, J. (2017).
\newblock Attention-aware deep reinforcement learning for video face
  recognition.
\newblock In {\em the IEEE International Conference on Computer Vision (ICCV)}.

\bibitem[\protect\astroncite{Ravi and Larochelle}{2017}]{Ravi2017}
Ravi, S. and Larochelle, H. (2017).
\newblock Optimization as a model for few-shot learning.
\newblock In {\em the International Conference on Learning Representations
  (ICLR)}.

\bibitem[\protect\astroncite{Reed and de~Freitas}{2016}]{Reed2016}
Reed, S. and de~Freitas, N. (2016).
\newblock Neural programmer-interpreters.
\newblock In {\em the International Conference on Learning Representations
  (ICLR)}.

\bibitem[\protect\astroncite{Ren et~al.}{2015}]{Ren2015}
Ren, S., He, K., Girshick, R., and Sun, J. (2015).
\newblock Faster {R-CNN}: Towards real-time object detection with region
  proposal networks.
\newblock In {\em the Annual Conference on Neural Information Processing
  Systems (NIPS)}.

\bibitem[\protect\astroncite{Ren et~al.}{2017}]{Ren2017}
Ren, Z., Wang, X., Zhang, N., Lv, X., and Li, L.-J. (2017).
\newblock Deep reinforcement learning-based image captioning with embedding
  reward.
\newblock In {\em the IEEE Conference on Computer Vision and Pattern
  Recognition (CVPR)}.

\bibitem[\protect\astroncite{Rennie et~al.}{2017}]{Rennie2017}
Rennie, S.~J., Marcheret, E., Mroueh, Y., Ross, J., and Goel, V. (2017).
\newblock Self-critical sequence training for image captioning.
\newblock In {\em the IEEE Conference on Computer Vision and Pattern
  Recognition (CVPR)}.

\bibitem[\protect\astroncite{Rhinehart and Kitani}{2017}]{Rhinehart2017}
Rhinehart, N. and Kitani, K.~M. (2017).
\newblock First-person activity forecasting with online inverse reinforcement
  learning.
\newblock In {\em the IEEE International Conference on Computer Vision (ICCV)}.

\bibitem[\protect\astroncite{Ribeiro et~al.}{2016}]{Ribeiro2016}
Ribeiro, M.~T., Singh, S., and Guestrin, C. (2016).
\newblock "why should i trust you?" explaining the predictions of any
  classifier.
\newblock In {\em ACM International Conference on Knowledge Discovery and Data
  Mining (SIGKDD)}.

\bibitem[\protect\astroncite{Riedmiller}{2005}]{Riedmiller2005}
Riedmiller, M. (2005).
\newblock Neural fitted {Q} iteration - first experiences with a data efficient
  neural reinforcement learning method.
\newblock In {\em European Conference on Machine Learning (ECML)}.

\bibitem[\protect\astroncite{{Rockt{\"a}schel} and
  {Riedel}}{2017}]{Rocktaschel2017}
{Rockt{\"a}schel}, T. and {Riedel}, S. (2017).
\newblock {End-to-end Differentiable Proving}.
\newblock {\em ArXiv e-prints}.

\bibitem[\protect\astroncite{Roijers et~al.}{2013}]{Roijers2013}
Roijers, D.~M., Vamplew, P., Whiteson, S., and Dazeley, R. (2013).
\newblock A survey of multi-objective sequential decision-making.
\newblock {\em Journal of Artificial Intelligence Research}, 48:67--113.

\bibitem[\protect\astroncite{{Ruder}}{2017}]{Ruder2017}
{Ruder}, S. (2017).
\newblock {An Overview of Multi-Task Learning in Deep Neural Networks}.
\newblock {\em ArXiv e-prints}.

\bibitem[\protect\astroncite{Ruelens et~al.}{2016}]{Ruelens2016}
Ruelens, F., Claessens, B.~J., Vandael, S., Schutter, B.~D., Babu{\v s}ka, R.,
  and Belmans, R. (2016).
\newblock Residential demand response of thermostatically controlled loads
  using batch reinforcement learning.
\newblock {\em IEEE Transactions on Smart Grid}, PP(99):1--11.

\bibitem[\protect\astroncite{Russell and Norvig}{2009}]{Russell2009}
Russell, S. and Norvig, P. (2009).
\newblock {\em Artificial Intelligence: A Modern Approach (3rd edition)}.
\newblock Pearson.

\bibitem[\protect\astroncite{Sabour et~al.}{2017}]{Sabour2017}
Sabour, S., Frosst, N., and Hinton, G.~E. (2017).
\newblock Dynamic routing between capsules.
\newblock In {\em the Annual Conference on Neural Information Processing
  Systems (NIPS)}.

\bibitem[\protect\astroncite{Salakhutdinov}{2016}]{Salakhutdinov2016}
Salakhutdinov, R. (2016).
\newblock Foundations of unsupervised deep learning, a talk at {Deep Learning
  School}, https://www.bayareadlschool.org.
\newblock \url{https://www.youtube.com/watch?v=rK6bchqeaN8}.

\bibitem[\protect\astroncite{{Salimans} et~al.}{2017}]{Salimans2017}
{Salimans}, T., {Ho}, J., {Chen}, X., and {Sutskever}, I. (2017).
\newblock {Evolution Strategies as a Scalable Alternative to Reinforcement
  Learning}.
\newblock {\em ArXiv e-prints}.

\bibitem[\protect\astroncite{{Santoro} et~al.}{2017}]{Santoro2017}
{Santoro}, A., {Raposo}, D., {Barrett}, D.~G.~T., {Malinowski}, M., {Pascanu},
  R., {Battaglia}, P., and {Lillicrap}, T. (2017).
\newblock {A simple neural network module for relational reasoning}.
\newblock {\em ArXiv e-prints}.

\bibitem[\protect\astroncite{Saon et~al.}{2016}]{Saon2016}
Saon, G., Sercu, T., Rennie, S., and Kuo, H.-K.~J. (2016).
\newblock {The IBM 2016 English Conversational Telephone Speech Recognition
  System}.
\newblock In {\em Annual Meeting of the International Speech Communication
  Association (INTERSPEECH)}.

\bibitem[\protect\astroncite{Saria}{2014}]{Saria2014}
Saria, S. (2014).
\newblock A \$3 trillion challenge to computational scientists: Transforming
  healthcare delivery.
\newblock {\em IEEE Intelligent Systems}, 29(4):82--87.

\bibitem[\protect\astroncite{Schaul et~al.}{2015}]{Schaul2015}
Schaul, T., Horgan, D., Gregor, K., and Silver, D. (2015).
\newblock Universal value function approximators.
\newblock In {\em the International Conference on Machine Learning (ICML)}.

\bibitem[\protect\astroncite{Schaul et~al.}{2016}]{Schaul2016}
Schaul, T., Quan, J., Antonoglou, I., and Silver, D. (2016).
\newblock Prioritized experience replay.
\newblock In {\em the International Conference on Learning Representations
  (ICLR)}.

\bibitem[\protect\astroncite{Schmidhuber}{2015}]{Schmidhuber2015-DL}
Schmidhuber, J. (2015).
\newblock Deep learning in neural networks: {A}n overview.
\newblock {\em Neural Networks}, 61:85--117.

\bibitem[\protect\astroncite{{Schulman} et~al.}{2017}]{Schulman2017}
{Schulman}, J., {Abbeel}, P., and {Chen}, X. (2017).
\newblock {Equivalence Between Policy Gradients and Soft Q-Learning}.
\newblock {\em ArXiv e-prints}.

\bibitem[\protect\astroncite{Schulman et~al.}{2015}]{Schulman2015}
Schulman, J., Levine, S., Moritz, P., Jordan, M.~I., and Abbeel, P. (2015).
\newblock Trust region policy optimization.
\newblock In {\em the International Conference on Machine Learning (ICML)}.

\bibitem[\protect\astroncite{Schuurmans and Zinkevich}{2016}]{Schuurmans2016}
Schuurmans, D. and Zinkevich, M. (2016).
\newblock Deep learning games.
\newblock In {\em the Annual Conference on Neural Information Processing
  Systems (NIPS)}.

\bibitem[\protect\astroncite{{Segler} et~al.}{2017}]{Segler2017}
{Segler}, M.~H.~S., {Preuss}, M., and {Waller}, M.~P. (2017).
\newblock {Learning to Plan Chemical Syntheses}.
\newblock {\em ArXiv e-prints}.

\bibitem[\protect\astroncite{Serban et~al.}{2015}]{Serban2015}
Serban, I.~V., Lowe, R., Charlin, L., and Pineau, J. (2015).
\newblock A survey of available corpora for building data-driven dialogue
  systems.
\newblock {\em arXiv e-prints}, abs/1512.05742.

\bibitem[\protect\astroncite{{Serban} et~al.}{2017}]{Serban2017}
{Serban}, I.~V., {Sankar}, C., {Germain}, M., {Zhang}, S., {Lin}, Z.,
  {Subramanian}, S., {Kim}, T., {Pieper}, M., {Chandar}, S., {Ke}, N.~R.,
  {Mudumba}, S., {de Brebisson}, A., {Sotelo}, J.~M.~R., {Suhubdy}, D.,
  {Michalski}, V., {Nguyen}, A., {Pineau}, J., and {Bengio}, Y. (2017).
\newblock {A Deep Reinforcement Learning Chatbot}.
\newblock {\em ArXiv e-prints}.

\bibitem[\protect\astroncite{Shah et~al.}{2016}]{Shah2016}
Shah, P., Hakkani-T{\"u}r, D., and Heck, L. (2016).
\newblock Interactive reinforcement learning for task-oriented dialogue
  management.
\newblock In {\em NIPS 2016 Deep Learning for Action and Interaction Workshop}.

\bibitem[\protect\astroncite{Shalev-Shwartz et~al.}{2017}]{Shalev-Shwartz2017}
Shalev-Shwartz, S., Shamir, O., and Shammah, S. (2017).
\newblock Failures of gradient-based deep learning.
\newblock In {\em the International Conference on Machine Learning (ICML)}.

\bibitem[\protect\astroncite{Sharma et~al.}{2017}]{Sharma2017}
Sharma, S., Lakshminarayanan, A.~S., and Ravindran, B. (2017).
\newblock Learning to repeat: Fine grained action repetition for deep
  reinforcement learning.
\newblock In {\em the International Conference on Learning Representations
  (ICLR)}.

\bibitem[\protect\astroncite{She and Chai}{2017}]{She2017}
She, L. and Chai, J. (2017).
\newblock Interactive learning for acquisition of grounded verb semantics
  towards human-robot communication.
\newblock In {\em the Association for Computational Linguistics annual meeting
  (ACL)}.

\bibitem[\protect\astroncite{Shen et~al.}{2017}]{Shen2017}
Shen, Y., Huang, P.-S., Gao, J., and Chen, W. (2017).
\newblock Reasonet: Learning to stop reading in machine comprehension.
\newblock In {\em ACM International Conference on Knowledge Discovery and Data
  Mining (SIGKDD)}.

\bibitem[\protect\astroncite{Shoham and Leyton-Brown}{2009}]{Shoham2009}
Shoham, Y. and Leyton-Brown, K. (2009).
\newblock {\em Multiagent Systems: Algorithmic, Game-Theoretic, and Logical
  Foundations}.
\newblock Cambridge University Press.

\bibitem[\protect\astroncite{Shoham et~al.}{2007}]{Shoham2007}
Shoham, Y., Powers, R., and Grenager, T. (2007).
\newblock If multi-agent learning is the answer, what is the question?
\newblock {\em Artificial Intelligence}, 171:365--377.

\bibitem[\protect\astroncite{Shortreed et~al.}{2011}]{Shortreed2011}
Shortreed, S.~M., Laber, E., Lizotte, D.~J., Stroup, T.~S., Pineau, J., and
  Murphy, S.~A. (2011).
\newblock Informing sequential clinical decision-making through reinforcement
  learning: an empirical study.
\newblock {\em Machine Learning}, 84:109--136.

\bibitem[\protect\astroncite{Shrivastava et~al.}{2017}]{Shrivastava2017}
Shrivastava, A., Pfister, T., Tuzel, O., Susskind, J., Wang, W., and Webb, R.
  (2017).
\newblock Learning from simulated and unsupervised images through adversarial
  training.
\newblock In {\em the IEEE Conference on Computer Vision and Pattern
  Recognition (CVPR)}.

\bibitem[\protect\astroncite{{Shwartz-Ziv} and
  {Tishby}}{2017}]{Shwartz-Ziv2017}
{Shwartz-Ziv}, R. and {Tishby}, N. (2017).
\newblock {Opening the Black Box of Deep Neural Networks via Information}.
\newblock {\em ArXiv e-prints}.

\bibitem[\protect\astroncite{Silver}{2016}]{Silver2016Tutorial}
Silver, D. (2016).
\newblock Deep reinforcement learning, a tutorial at {ICML} 2016.
\newblock \url{http://icml.cc/2016/tutorials/deep_rl_tutorial.pdf}.

\bibitem[\protect\astroncite{Silver et~al.}{2016a}]{Silver-AlphaGo-2016}
Silver, D., Huang, A., Maddison, C.~J., Guez, A., Sifre, L., Van Den~Driessche,
  G., Schrittwieser, J., Antonoglou, I., Panneershelvam, V., Lanctot, M.,
  et~al. (2016a).
\newblock Mastering the game of go with deep neural networks and tree search.
\newblock {\em Nature}, 529(7587):484--489.

\bibitem[\protect\astroncite{{Silver} et~al.}{2017}]{Silver-AlphaZero-2017}
{Silver}, D., {Hubert}, T., {Schrittwieser}, J., {Antonoglou}, I., {Lai}, M.,
  {Guez}, A., {Lanctot}, M., {Sifre}, L., {Kumaran}, D., {Graepel}, T.,
  {Lillicrap}, T., {Simonyan}, K., and {Hassabis}, D. (2017).
\newblock {Mastering Chess and Shogi by Self-Play with a General Reinforcement
  Learning Algorithm}.
\newblock {\em ArXiv e-prints}.

\bibitem[\protect\astroncite{Silver et~al.}{2014}]{Silver-DPG-2014}
Silver, D., Lever, G., Heess, N., Degris, T., Wierstra, D., and Riedmiller, M.
  (2014).
\newblock Deterministic policy gradient algorithms.
\newblock In {\em the International Conference on Machine Learning (ICML)}.

\bibitem[\protect\astroncite{Silver et~al.}{2013}]{Silver2013}
Silver, D., Newnham, L., Barker, D., Weller, S., and McFall, J. (2013).
\newblock Concurrent reinforcement learning from customer interactions.
\newblock In {\em the International Conference on Machine Learning (ICML)}.

\bibitem[\protect\astroncite{Silver et~al.}{2017}]{Silver-AlphaGo-2017}
Silver, D., Schrittwieser, J., Simonyan, K., Antonoglou, I., Huang, A., Guez,
  A., Hubert, T., Baker, L., Lai, M., Bolton, A., Chen, Y., Lillicrap, T., Hui,
  F., Sifre, L., van~den Driessche, G., Graepel, T., and Hassabis, D. (2017).
\newblock Mastering the game of go without human knowledge.
\newblock {\em Nature}, 550:354--359.

\bibitem[\protect\astroncite{Silver et~al.}{2016b}]{Silver-Predictron-2016}
Silver, D., van Hasselt, H., Hessel, M., Schaul, T., Guez, A., Harley, T.,
  Dulac-Arnold, G., Reichert, D., Rabinowitz, N., Barreto, A., and Degris, T.
  (2016b).
\newblock The predictron: End-to-end learning and planning.
\newblock In {\em NIPS 2016 Deep Reinforcement Learning Workshop}.

\bibitem[\protect\astroncite{{Simeone}}{2017}]{Simeone2017}
{Simeone}, O. (2017).
\newblock {A Brief Introduction to Machine Learning for Engineers}.
\newblock {\em ArXiv e-prints}.

\bibitem[\protect\astroncite{{Smith}}{2017}]{Smith2017}
{Smith}, L.~N. (2017).
\newblock {Best Practices for Applying Deep Learning to Novel Applications}.
\newblock {\em ArXiv e-prints}.

\bibitem[\protect\astroncite{Smith et~al.}{2017}]{Smith2017multi}
Smith, V., Chiang, C.-K., Sanjabi, M., and Talwalkar, A. (2017).
\newblock Federated multi-task learning.
\newblock In {\em the Annual Conference on Neural Information Processing
  Systems (NIPS)}.

\bibitem[\protect\astroncite{{Snell} et~al.}{2017}]{Snell2017}
{Snell}, J., {Swersky}, K., and {Zemel}, R.~S. (2017).
\newblock {Prototypical Networks for Few-shot Learning}.
\newblock {\em ArXiv e-prints}.

\bibitem[\protect\astroncite{Socher et~al.}{2011}]{Socher2011}
Socher, R., Pennington, J., Huang, E.~H., Ng, A.~Y., and Manning, C.~D. (2011).
\newblock Semi-supervised recursive autoencoders for predicting sentiment
  distributions.
\newblock In {\em Conference on Empirical Methods in Natural Language
  Processing (EMNLP)}.

\bibitem[\protect\astroncite{Socher et~al.}{2013}]{Socher2013}
Socher, R., Perelygin, A., Wu, J., Chuang, J., Manning, C., Ng, A., and Potts,
  C. (2013).
\newblock Recursive deep models for semantic compositionality over a sentiment
  tree- bank.
\newblock In {\em Conference on Empirical Methods in Natural Language
  Processing (EMNLP)}.

\bibitem[\protect\astroncite{{Song} and {Roth}}{2017}]{Song2017Knowledge}
{Song}, Y. and {Roth}, D. (2017).
\newblock {Machine Learning with World Knowledge: The Position and Survey}.
\newblock {\em ArXiv e-prints}.

\bibitem[\protect\astroncite{Spring and Shrivastava}{2017}]{Spring2017}
Spring, R. and Shrivastava, A. (2017).
\newblock Scalable and sustainable deep learning via randomized hashing.
\newblock In {\em ACM International Conference on Knowledge Discovery and Data
  Mining (SIGKDD)}.

\bibitem[\protect\astroncite{Srivastava et~al.}{2014}]{Srivastava2014}
Srivastava, N., Hinton, G., Krizhevsky, A., Sutskever, I., and Salakhutdinov,
  R. (2014).
\newblock Dropout: A simple way to prevent neural networks from overfitting.
\newblock {\em The Journal of Machine Learning Research}, 15:1929--1958.

\bibitem[\protect\astroncite{Stadie et~al.}{2017}]{Stadie2017}
Stadie, B.~C., Abbeel, P., and Sutskever, I. (2017).
\newblock Third person imitation learning.
\newblock In {\em the International Conference on Learning Representations
  (ICLR)}.

\bibitem[\protect\astroncite{Stoica et~al.}{2017}]{Stoica2017}
Stoica, I., Song, D., Popa, R.~A., Patterson, D.~A., Mahoney, M.~W., Katz,
  R.~H., Joseph, A.~D., Jordan, M., Hellerstein, J.~M., Gonzalez, J., Goldberg,
  K., Ghodsi, A., Culler, D.~E., and Abbeel, P. (2017).
\newblock A berkeley view of systems challenges for {AI}.
\newblock {\em Technical Report No. UCB/EECS-2017-159}.

\bibitem[\protect\astroncite{Stone et~al.}{2016}]{Stone2016}
Stone, P., Brooks, R., Brynjolfsson, E., Calo, R., Etzioni, O., Hager, G.,
  Hirschberg, J., Kalyanakrishnan, S., Kamar, E., Kraus, S., Leyton-Brown, K.,
  Parkes, D., Press, W., Saxenian, A., Shah, J., Tambe, M., and Teller, A.
  (2016).
\newblock {\em Artificial Intelligence and Life in 2030 - One Hundred Year
  Study on Artificial Intelligence: Report of the 2015-2016 Study Panel}.
\newblock Stanford University, Stanford, CA.

\bibitem[\protect\astroncite{Stone and Veloso}{2000}]{Stone2000}
Stone, P. and Veloso, M. (2000).
\newblock Multiagent systems: A survey from a machine learning perspective.
\newblock {\em Autonomous Robots}, 8(3):345--383.

\bibitem[\protect\astroncite{{Strub} et~al.}{2017}]{Strub2017}
{Strub}, F., {de Vries}, H., {Mary}, J., {Piot}, B., {Courville}, A., and
  {Pietquin}, O. (2017).
\newblock {End-to-end optimization of goal-driven and visually grounded
  dialogue systems}.
\newblock {\em ArXiv e-prints}.

\bibitem[\protect\astroncite{Su et~al.}{2016a}]{Su2016-continuous}
Su, P.-H., Gasic, M., Mrksic, N., Rojas-Barahona, L., Ultes, S., Vandyke, D.,
  Wen, T.-H., and Young, S. (2016a).
\newblock {Continuously Learning Neural Dialogue Management}.
\newblock {\em ArXiv e-prints}.

\bibitem[\protect\astroncite{Su et~al.}{2016b}]{Su2016}
Su, P.-H., Gas{\v i}{\'c}, M., Mrks{\v i}{\'c}, N., Rojas-Barahona, L., Ultes,
  S., Vandyke, D., Wen, T.-H., and Young, S. (2016b).
\newblock On-line active reward learning for policy optimisation in spoken
  dialogue systems.
\newblock In {\em the Association for Computational Linguistics annual meeting
  (ACL)}.

\bibitem[\protect\astroncite{Sukhbaatar et~al.}{2016}]{Sukhbaatar2016}
Sukhbaatar, S., Szlam, A., and Fergus, R. (2016).
\newblock Learning multiagent communication with backpropagation.
\newblock In {\em the Annual Conference on Neural Information Processing
  Systems (NIPS)}.

\bibitem[\protect\astroncite{Sukhbaatar et~al.}{2015}]{Sukhbaatar2015}
Sukhbaatar, S., Weston, J., and Fergus, R. (2015).
\newblock End-to-end memory networks.
\newblock In {\em the Annual Conference on Neural Information Processing
  Systems (NIPS)}.

\bibitem[\protect\astroncite{Supan{\v c}i{\v c} and
  Ramanan}{2017}]{Supancic2017}
Supan{\v c}i{\v c}, III, J. and Ramanan, D. (2017).
\newblock Tracking as online decision-making: Learning a policy from streaming
  videos with reinforcement learning.
\newblock In {\em the IEEE International Conference on Computer Vision (ICCV)}.

\bibitem[\protect\astroncite{Surana et~al.}{2016}]{Surana2016}
Surana, A., Sarkar, S., and Reddy, K.~K. (2016).
\newblock Guided deep reinforcement learning for additive manufacturing control
  application.
\newblock In {\em NIPS 2016 Deep Reinforcement Learning Workshop}.

\bibitem[\protect\astroncite{Sutskever et~al.}{2014}]{Sutskever2014}
Sutskever, I., Vinyals, O., and Le, Q.~V. (2014).
\newblock Sequence to sequence learning with neural networks.
\newblock In {\em the Annual Conference on Neural Information Processing
  Systems (NIPS)}.

\bibitem[\protect\astroncite{Sutton}{2016}]{Sutton2016Course}
Sutton, R. (2016).
\newblock Reinforcement learning for artificial intelligence, course slides.
\newblock \url{http://www.incompleteideas.net/sutton/609\%20dropbox/}.

\bibitem[\protect\astroncite{Sutton}{1988}]{Sutton1988}
Sutton, R.~S. (1988).
\newblock Learning to predict by the methods of temporal differences.
\newblock {\em Machine Learning}, 3(1):9--44.

\bibitem[\protect\astroncite{Sutton}{1990}]{Sutton1990}
Sutton, R.~S. (1990).
\newblock Integrated architectures for learning, planning, and reacting based
  on approximating dynamic programming.
\newblock In {\em the International Conference on Machine Learning (ICML)}.

\bibitem[\protect\astroncite{Sutton}{1992}]{Sutton1992}
Sutton, R.~S. (1992).
\newblock Adapting bias by gradient descent: An incremental version of
  delta-bar-delta.
\newblock In {\em the AAAI Conference on Artificial Intelligence (AAAI)}.

\bibitem[\protect\astroncite{Sutton and Barto}{1998}]{Sutton98}
Sutton, R.~S. and Barto, A.~G. (1998).
\newblock {\em Reinforcement Learning: An Introduction}.
\newblock MIT Press.

\bibitem[\protect\astroncite{Sutton and Barto}{2018}]{Sutton2018}
Sutton, R.~S. and Barto, A.~G. (2018).
\newblock {\em Reinforcement Learning: An Introduction (2nd Edition, in
  preparation)}.
\newblock MIT Press.

\bibitem[\protect\astroncite{Sutton et~al.}{2009a}]{Sutton2009GTD-ICML}
Sutton, R.~S., Maei, H.~R., Precup, D., Bhatnagar, S., Silver, D.,
  Szepesv{\'a}ri, C., and Wiewiora, E. (2009a).
\newblock Fast gradient-descent methods for temporal-difference learning with
  linear function approximation.
\newblock In {\em the International Conference on Machine Learning (ICML)}.

\bibitem[\protect\astroncite{Sutton et~al.}{2016}]{Sutton2016}
Sutton, R.~S., Mahmood, A.~R., and White, M. (2016).
\newblock An emphatic approach to the problem of off-policy temporal-difference
  learning.
\newblock {\em The Journal of Machine Learning Research}, 17:1--29.

\bibitem[\protect\astroncite{Sutton et~al.}{2000}]{Sutton2000}
Sutton, R.~S., McAllester, D., Singh, S., and Mansour, Y. (2000).
\newblock Policy gradient methods for reinforcement learning with function
  approximation.
\newblock In {\em the Annual Conference on Neural Information Processing
  Systems (NIPS)}.

\bibitem[\protect\astroncite{Sutton et~al.}{2011}]{Sutton2011}
Sutton, R.~S., Modayil, J., Delp, M., Degris, T., Pilarski, P.~M., White, A.,
  and Precup, D. (2011).
\newblock Horde: A scalable real-time architecture for learning knowledge from
  unsupervised sensorimotor interaction, , proc. of 10th.
\newblock In {\em International Conference on Autonomous Agents and Multiagent
  Systems (AAMAS)}.

\bibitem[\protect\astroncite{Sutton et~al.}{1999}]{Sutton1999}
Sutton, R.~S., Precup, D., and Singh, S. (1999).
\newblock Between mdps and semi-mdps: A framework for temporal abstraction in
  reinforcement learning.
\newblock {\em Artificial Intelligence}, 112(1-2):181--211.

\bibitem[\protect\astroncite{Sutton et~al.}{2009b}]{Sutton2009GTD-NIPS}
Sutton, R.~S., Szepesv{\'a}ri, C., and Maei, H.~R. (2009b).
\newblock A convergent {O}($n$) algorithm for off-policy temporal-difference
  learning with linear function approximation.
\newblock In {\em the Annual Conference on Neural Information Processing
  Systems (NIPS)}.

\bibitem[\protect\astroncite{Synnaeve et~al.}{2016}]{Synnaeve2016}
Synnaeve, G., Nardelli, N., Auvolat, A., Chintala, S., Lacroix, T., Lin, Z.,
  Richoux, F., and Usunier, N. (2016).
\newblock {TorchCraft: a Library for Machine Learning Research on Real-Time
  Strategy Games}.
\newblock {\em ArXiv e-prints}.

\bibitem[\protect\astroncite{{Sze} et~al.}{2017}]{Sze2017}
{Sze}, V., {Chen}, Y.-H., {Yang}, T.-J., and {Emer}, J. (2017).
\newblock {Efficient Processing of Deep Neural Networks: A Tutorial and
  Survey}.
\newblock {\em ArXiv e-prints}.

\bibitem[\protect\astroncite{Szepesv{\'a}ri}{2010}]{Szepesvari2010}
Szepesv{\'a}ri, C. (2010).
\newblock {\em Algorithms for Reinforcement Learning}.
\newblock Morgan \& Claypool.

\bibitem[\protect\astroncite{Tamar et~al.}{2016}]{Tamar2016}
Tamar, A., Wu, Y., Thomas, G., Levine, S., and Abbeel, P. (2016).
\newblock Value iteration networks.
\newblock In {\em the Annual Conference on Neural Information Processing
  Systems (NIPS)}.

\bibitem[\protect\astroncite{Tang et~al.}{2017}]{Tang2017}
Tang, H., Houthooft, R., Foote, D., Stooke, A., Chen, X., Duan, Y., Schulman,
  J., Turck, F.~D., and Abbeel, P. (2017).
\newblock Exploration: A study of count-based exploration for deep
  reinforcement learning.
\newblock In {\em the Annual Conference on Neural Information Processing
  Systems (NIPS)}.

\bibitem[\protect\astroncite{Tanner and White}{2009}]{Tanner2009}
Tanner, B. and White, A. (2009).
\newblock {RL}-{G}lue : Language-independent software for
  reinforcement-learning experiments.
\newblock {\em Journal of Machine Learning Research}, 10:2133--2136.

\bibitem[\protect\astroncite{{Tassa} et~al.}{2018}]{Tassa2018}
{Tassa}, Y., {Doron}, Y., {Muldal}, A., {Erez}, T., {Li}, Y., {de Las Casas},
  D., {Budden}, D., {Abdolmaleki}, A., {Merel}, J., {Lefrancq}, A.,
  {Lillicrap}, T., and {Riedmiller}, M. (2018).
\newblock {DeepMind Control Suite}.
\newblock {\em ArXiv e-prints}.

\bibitem[\protect\astroncite{Taylor and Stone}{2009}]{Taylor09}
Taylor, M.~E. and Stone, P. (2009).
\newblock Transfer learning for reinforcement learning domains: A survey.
\newblock {\em Journal of Machine Learning Research}, 10:1633--1685.

\bibitem[\protect\astroncite{Tesauro}{1994}]{Tesauro1994}
Tesauro, G. (1994).
\newblock {TD-Gammon}, a self-teaching backgammon program, achieves
  master-level play.
\newblock {\em Neural Computation}, 6(2):215--219.

\bibitem[\protect\astroncite{Tessler et~al.}{2017}]{Tessler2017}
Tessler, C., Givony, S., Zahavy, T., Mankowitz, D.~J., and Mannor, S. (2017).
\newblock A deep hierarchical approach to lifelong learning in minecraft.
\newblock In {\em the AAAI Conference on Artificial Intelligence (AAAI)}.

\bibitem[\protect\astroncite{Theocharous et~al.}{2015}]{Theocharous2015}
Theocharous, G., Thomas, P.~S., and Ghavamzadeh, M. (2015).
\newblock Personalized ad recommendation systems for life-time value
  optimization with guarantees.
\newblock In {\em the International Joint Conference on Artificial Intelligence
  (IJCAI)}.

\bibitem[\protect\astroncite{{Tian} et~al.}{2017}]{Tian2017ELF}
{Tian}, Y., {Gong}, Q., {Shang}, W., {Wu}, Y., and {Zitnick}, L. (2017).
\newblock {ELF: An Extensive, Lightweight and Flexible Research Platform for
  Real-time Strategy Games}.
\newblock {\em ArXiv e-prints}.

\bibitem[\protect\astroncite{{Tram{\`e}r} et~al.}{2017}]{Tramer2017}
{Tram{\`e}r}, F., {Kurakin}, A., {Papernot}, N., {Boneh}, D., and {McDaniel},
  P. (2017).
\newblock {Ensemble Adversarial Training: Attacks and Defenses}.
\newblock {\em ArXiv e-prints}.

\bibitem[\protect\astroncite{Tran et~al.}{2017}]{Tran2017}
Tran, D., Hoffman, M.~D., Saurous, R.~A., Brevdo, E., Murphy, K., and Blei,
  D.~M. (2017).
\newblock Deep probabilistic programming.
\newblock In {\em the International Conference on Learning Representations
  (ICLR)}.

\bibitem[\protect\astroncite{Trischler et~al.}{2016}]{Trischler2016}
Trischler, A., Ye, Z., Yuan, X., and Suleman, K. (2016).
\newblock Natural language comprehension with the epireader.
\newblock In {\em Conference on Empirical Methods in Natural Language
  Processing (EMNLP)}.

\bibitem[\protect\astroncite{Tsitsiklis and {Van Roy}}{1997}]{Tsitsiklis97}
Tsitsiklis, J.~N. and {Van Roy}, B. (1997).
\newblock An analysis of temporal-difference learning with function
  approximation.
\newblock {\em IEEE Transactions on Automatic Control}, 42(5):674--690.

\bibitem[\protect\astroncite{Tsitsiklis and {Van Roy}}{2001}]{Tsitsiklis01}
Tsitsiklis, J.~N. and {Van Roy}, B. (2001).
\newblock Regression methods for pricing complex {American}-style options.
\newblock {\em IEEE Transactions on Neural Networks}, 12(4):694--703.

\bibitem[\protect\astroncite{Usunier et~al.}{2017}]{Usunier2016}
Usunier, N., Synnaeve, G., Lin, Z., and Chintala, S. (2017).
\newblock Episodic exploration for deep deterministic policies: An application
  to {StarCraft} micromanagement tasks.
\newblock In {\em the International Conference on Learning Representations
  (ICLR)}.

\bibitem[\protect\astroncite{van~der Pol and Oliehoek}{2017}]{vanDerPol2017}
van~der Pol, E. and Oliehoek, F.~A. (2017).
\newblock Coordinated deep reinforcement learners for traffic light control.
\newblock In {\em NIPS'16 Workshop on Learning, Inference and Control of
  Multi-Agent Systems}.

\bibitem[\protect\astroncite{van Hasselt et~al.}{2016a}]{vanHasselt2016}
van Hasselt, H., Guez, A., , and Silver, D. (2016a).
\newblock Deep reinforcement learning with double {Q}-learning.
\newblock In {\em the AAAI Conference on Artificial Intelligence (AAAI)}.

\bibitem[\protect\astroncite{van Hasselt
  et~al.}{2016b}]{vanHasselt2016-adaptive}
van Hasselt, H., Guez, A., Hessel, M., Mnih, V., and Silver, D. (2016b).
\newblock Learning values across many orders of magnitude.
\newblock In {\em the Annual Conference on Neural Information Processing
  Systems (NIPS)}.

\bibitem[\protect\astroncite{{van Seijen} et~al.}{2017}]{vanSeijen2017}
{van Seijen}, H., {Fatemi}, M., {Romoff}, J., {Laroche}, R., {Barnes}, T., and
  {Tsang}, J. (2017).
\newblock Hybrid reward architecture for reinforcement learning.
\newblock In {\em the Annual Conference on Neural Information Processing
  Systems (NIPS)}.

\bibitem[\protect\astroncite{{Vaswani} et~al.}{2017}]{Vaswani2017}
{Vaswani}, A., {Shazeer}, N., {Parmar}, N., {Uszkoreit}, J., {Jones}, L.,
  {Gomez}, A.~N., {Kaiser}, L., and {Polosukhin}, I. (2017).
\newblock Attention is all you need.
\newblock In {\em the Annual Conference on Neural Information Processing
  Systems (NIPS)}.

\bibitem[\protect\astroncite{Venkatraman et~al.}{2017}]{Venkatraman2017}
Venkatraman, A., Rhinehart, N., Sun, W., Pinto, L., Hebert, M., Boots, B.,
  Kitani, K.~M., and Bagnell, J.~A. (2017).
\newblock Predictive-state decoders: Encoding the future into recurrent
  networks.
\newblock In {\em the Annual Conference on Neural Information Processing
  Systems (NIPS)}.

\bibitem[\protect\astroncite{{Ve{\v c}er{\'{\i}}k} et~al.}{2017}]{Vecerik2017}
{Ve{\v c}er{\'{\i}}k}, M., {Hester}, T., {Scholz}, J., {Wang}, F., {Pietquin},
  O., {Piot}, B., {Heess}, N., {Roth{\"o}rl}, T., {Lampe}, T., and
  {Riedmiller}, M. (2017).
\newblock Leveraging demonstrations for deep reinforcement learning on robotics
  problems with sparse rewards.
\newblock In {\em the Annual Conference on Neural Information Processing
  Systems (NIPS)}.

\bibitem[\protect\astroncite{Vezhnevets et~al.}{2016}]{Vezhnevets2016}
Vezhnevets, A.~S., Mnih, V., Agapiou, J., Osindero, S., Graves, A., Vinyals,
  O., and Kavukcuoglu, K. (2016).
\newblock Strategic attentive writer for learning macro-actions.
\newblock In {\em the Annual Conference on Neural Information Processing
  Systems (NIPS)}.

\bibitem[\protect\astroncite{Vezhnevets et~al.}{2017}]{Vezhnevets2017}
Vezhnevets, A.~S., Osindero, S., Schaul, T., Heess, N., Jaderberg, M., Silver,
  D., and Kavukcuoglu, K. (2017).
\newblock Feudal networks for hierarchical reinforcement learning.
\newblock In {\em the International Conference on Machine Learning (ICML)}.

\bibitem[\protect\astroncite{Vinyals et~al.}{2016}]{Vinyals2016}
Vinyals, O., Blundell, C., Lillicrap, T., Kavukcuoglu, K., and Wierstra, D.
  (2016).
\newblock Matching networks for one shot learning.
\newblock In {\em the Annual Conference on Neural Information Processing
  Systems (NIPS)}.

\bibitem[\protect\astroncite{Vinyals et~al.}{2015}]{Vinyals2015}
Vinyals, O., Fortunato, M., and Jaitly, N. (2015).
\newblock Pointer networks.
\newblock In {\em the Annual Conference on Neural Information Processing
  Systems (NIPS)}.

\bibitem[\protect\astroncite{{Wang} and {Raj}}{2017}]{Wang2017Origin}
{Wang}, H. and {Raj}, B. (2017).
\newblock {On the Origin of Deep Learning}.
\newblock {\em ArXiv e-prints}.

\bibitem[\protect\astroncite{{Wang} et~al.}{2016}]{Wang2016LearnRL}
{Wang}, J.~X., {Kurth-Nelson}, Z., {Tirumala}, D., {Soyer}, H., {Leibo}, J.~Z.,
  {Munos}, R., {Blundell}, C., {Kumaran}, D., and {Botvinick}, M. (2016).
\newblock {Learning to reinforcement learn}.
\newblock {\em ArXiv e-prints}.

\bibitem[\protect\astroncite{Wang et~al.}{2016a}]{WangSida2016}
Wang, S.~I., Liang, P., and Manning, C.~D. (2016a).
\newblock Learning language games through interaction.
\newblock In {\em the Association for Computational Linguistics annual meeting
  (ACL)}.

\bibitem[\protect\astroncite{Wang et~al.}{2017a}]{Wang2017RNet}
Wang, W., Yang, N., Wei, F., Chang, B., and Zhou, M. (2017a).
\newblock Gated self-matching networks for reading comprehension and question
  answering.
\newblock In {\em the Association for Computational Linguistics annual meeting
  (ACL)}.

\bibitem[\protect\astroncite{Wang et~al.}{2017b}]{Wang2017}
Wang, Z., Bapst, V., Heess, N., Mnih, V., Munos, R., Kavukcuoglu, K., and
  de~Freitas, N. (2017b).
\newblock Sample efficient actor-critic with experience replay.
\newblock In {\em the International Conference on Learning Representations
  (ICLR)}.

\bibitem[\protect\astroncite{{Wang} et~al.}{2017}]{Wang2017Imitation}
{Wang}, Z., {Merel}, J., {Reed}, S., {Wayne}, G., {de Freitas}, N., and
  {Heess}, N. (2017).
\newblock {Robust Imitation of Diverse Behaviors}.
\newblock {\em ArXiv e-prints}.

\bibitem[\protect\astroncite{Wang et~al.}{2016b}]{Wang-Dueling-2016}
Wang, Z., Schaul, T., Hessel, M., van Hasselt, H., Lanctot, M., and de~Freitas,
  N. (2016b).
\newblock Dueling network architectures for deep reinforcement learning.
\newblock In {\em the International Conference on Machine Learning (ICML)}.

\bibitem[\protect\astroncite{Watkins and Dayan}{1992}]{Watkins1992}
Watkins, C. J. C.~H. and Dayan, P. (1992).
\newblock Q-learning.
\newblock {\em Machine Learning}, 8:279--292.

\bibitem[\protect\astroncite{Watter et~al.}{2015}]{Watter2015}
Watter, M., Springenberg, J.~T., Boedecker, J., and Riedmiller, M. (2015).
\newblock Embed to control: A locally linear latent dynamics model for control
  from raw images.
\newblock In {\em the Annual Conference on Neural Information Processing
  Systems (NIPS)}.

\bibitem[\protect\astroncite{{Watters} et~al.}{2017}]{Watters2017}
{Watters}, N., {Tacchetti}, A., {Weber}, T., {Pascanu}, R., {Battaglia}, P.,
  and {Zoran}, D. (2017).
\newblock Visual interaction networks: Learning a physics simulator from video.
\newblock In {\em the Annual Conference on Neural Information Processing
  Systems (NIPS)}.

\bibitem[\protect\astroncite{{Weber} et~al.}{2017}]{Weber2017}
{Weber}, T., {Racani{\`e}re}, S., {Reichert}, D.~P., {Buesing}, L., {Guez}, A.,
  {Jimenez Rezende}, D., {Puigdom{\`e}nech Badia}, A., {Vinyals}, O., {Heess},
  N., {Li}, Y., {Pascanu}, R., {Battaglia}, P., {Silver}, D., and {Wierstra},
  D. (2017).
\newblock Imagination-augmented agents for deep reinforcement learning.
\newblock In {\em the Annual Conference on Neural Information Processing
  Systems (NIPS)}.

\bibitem[\protect\astroncite{Weiss et~al.}{2016}]{Weiss2016}
Weiss, K., Khoshgoftaar, T.~M., and Wang, D. (2016).
\newblock A survey of transfer learning.
\newblock {\em Journal of Big Data}, 3(9).

\bibitem[\protect\astroncite{{Weiss} et~al.}{2017}]{Weiss2017}
{Weiss}, R.~J., {Chorowski}, J., {Jaitly}, N., {Wu}, Y., and {Chen}, Z. (2017).
\newblock {Sequence-to-Sequence Models Can Directly Transcribe Foreign Speech}.
\newblock {\em ArXiv e-prints}.

\bibitem[\protect\astroncite{Welleck et~al.}{2017}]{Welleck2017}
Welleck, S., Mao, J., Cho, K., and Zhang, Z. (2017).
\newblock Saliency-based sequential image attention with multiset prediction.
\newblock In {\em the Annual Conference on Neural Information Processing
  Systems (NIPS)}.

\bibitem[\protect\astroncite{Wen et~al.}{2015a}]{WenTH2015}
Wen, T.-H., Gasic, M., Mrksic, N., Su, P.-H., Vandyke, D., and Young, S.
  (2015a).
\newblock Semantically conditioned {LSTM}-based natural language generation for
  spoken dialogue systems.
\newblock In {\em Conference on Empirical Methods in Natural Language
  Processing (EMNLP)}.

\bibitem[\protect\astroncite{{Wen} et~al.}{2017}]{WenTH2017}
{Wen}, T.-H., {Vandyke}, D., {Mrksic}, N., {Gasic}, M., {Rojas-Barahona},
  L.~M., {Su}, P.-H., {Ultes}, S., and {Young}, S. (2017).
\newblock A network-based end-to-end trainable task-oriented dialogue system.
\newblock In {\em Proceedings of the 15th Conference of the European Chapter of
  the Association for Computational Linguistics (EACL)}.

\bibitem[\protect\astroncite{Wen et~al.}{2015b}]{Wen2015}
Wen, Z., O'Neill, D., and Maei, H. (2015b).
\newblock Optimal demand response using device-based reinforcement learning.
\newblock {\em IEEE Transactions on Smart Grid}, 6(5):2312--2324.

\bibitem[\protect\astroncite{Weston et~al.}{2015}]{Weston2015}
Weston, J., Chopra, S., and Bordes, A. (2015).
\newblock Memory networks.
\newblock In {\em the International Conference on Learning Representations
  (ICLR)}.

\bibitem[\protect\astroncite{White and White}{2016}]{White2016}
White, A. and White, M. (2016).
\newblock Investigating practical linear temporal difference learning.
\newblock In {\em the International Conference on Autonomous Agents \&
  Multiagent Systems (AAMAS)}.

\bibitem[\protect\astroncite{{Whye Teh} et~al.}{2017}]{WhyeTeh2017}
{Whye Teh}, Y., {Bapst}, V., {Czarnecki}, W.~M., {Quan}, J., {Kirkpatrick}, J.,
  {Hadsell}, R., {Heess}, N., and {Pascanu}, R. (2017).
\newblock Distral: Robust multitask reinforcement learning.
\newblock In {\em the Annual Conference on Neural Information Processing
  Systems (NIPS)}.

\bibitem[\protect\astroncite{Wiering and van Otterlo}{2012}]{Wiering2012}
Wiering, M. and van Otterlo, M. (2012).
\newblock {\em Reinforcement Learning: State-of-the-Art (edited book)}.
\newblock Springer.

\bibitem[\protect\astroncite{Williams et~al.}{2017}]{Williams2017}
Williams, J.~D., Asadi, K., and Zweig, G. (2017).
\newblock Hybrid code networks: practical and efficient end-to-end dialog
  control with supervised and reinforcement learning.
\newblock In {\em the Association for Computational Linguistics annual meeting
  (ACL)}.

\bibitem[\protect\astroncite{Williams and Zweig}{2016}]{Williams2016}
Williams, J.~D. and Zweig, G. (2016).
\newblock {End-to-end LSTM-based dialog control optimized with supervised and
  reinforcement learning}.
\newblock {\em ArXiv e-prints}.

\bibitem[\protect\astroncite{Williams}{1992}]{Williams1992}
Williams, R.~J. (1992).
\newblock Simple statistical gradient-following algorithms for connectionist
  reinforcement learning.
\newblock {\em Machine Learning}, 8(3):229--256.

\bibitem[\protect\astroncite{{Wilson} et~al.}{2017}]{Wilson2017}
{Wilson}, A.~C., {Roelofs}, R., {Stern}, M., {Srebro}, N., and {Recht}, B.
  (2017).
\newblock {The Marginal Value of Adaptive Gradient Methods in Machine
  Learning}.
\newblock {\em ArXiv e-prints}.

\bibitem[\protect\astroncite{Wu et~al.}{2017a}]{Wu2017De-animation}
Wu, J., Lu, E., Kohli, P., Freeman, B., and Tenenbaum, J. (2017a).
\newblock Learning to see physics via visual de-animation.
\newblock In {\em the Annual Conference on Neural Information Processing
  Systems (NIPS)}.

\bibitem[\protect\astroncite{Wu et~al.}{2017b}]{Wu2017De-rendering}
Wu, J., Tenenbaum, J.~B., and Kohli, P. (2017b).
\newblock Neural scene de-rendering.
\newblock In {\em the IEEE Conference on Computer Vision and Pattern
  Recognition (CVPR)}.

\bibitem[\protect\astroncite{Wu et~al.}{2015}]{Wu2015Galileo}
Wu, J., Yildirim, I., Lim, J.~J., Freeman, B., and Tenenbaum, J. (2015).
\newblock Galileo: Perceiving physical object properties by integrating a
  physics engine with deep learning.
\newblock In {\em the Annual Conference on Neural Information Processing
  Systems (NIPS)}.

\bibitem[\protect\astroncite{Wu et~al.}{2017c}]{Wu2017AdversarialNMT}
Wu, L., Xia, Y., Zhao, L., Tian, F., Qin, T., Lai, J., and Liu, T.-Y. (2017c).
\newblock {Adversarial Neural Machine Translation}.
\newblock {\em ArXiv e-prints}.

\bibitem[\protect\astroncite{{Wu} et~al.}{2017}]{Wu2017TRPO}
{Wu}, Y., {Mansimov}, E., {Liao}, S., {Grosse}, R., and {Ba}, J. (2017).
\newblock Scalable trust-region method for deep reinforcement learning using
  kronecker-factored approximation.
\newblock In {\em the Annual Conference on Neural Information Processing
  Systems (NIPS)}.

\bibitem[\protect\astroncite{Wu et~al.}{2016}]{Wu2016}
Wu, Y., Schuster, M., Chen, Z., Le, Q.~V., Norouzi, M., Macherey, W., Krikun,
  M., Cao, Y., Gao, Q., Macherey, K., Klingner, J., Shah, A., Johnson, M., Liu,
  X., Kaiser, L., Gouws, S., Kato, Y., Kudo, T., Kazawa, H., Stevens, K.,
  Kurian, G., Patil, N., Wang, W., Young, C., Smith, J., Riesa, J., Rudnick,
  A., Vinyals, O., Corrado, G., Hughes, M., and Dean, J. (2016).
\newblock Google's neural machine translation system: Bridging the gap between
  human and machine translation.
\newblock {\em ArXiv e-prints}.

\bibitem[\protect\astroncite{Wu and Tian}{2017}]{Wu2017}
Wu, Y. and Tian, Y. (2017).
\newblock Training agent for first-person shooter game with actor-critic
  curriculum learning.
\newblock In {\em the International Conference on Learning Representations
  (ICLR)}.

\bibitem[\protect\astroncite{Xiong et~al.}{2017a}]{Xiong2017}
Xiong, C., Zhong, V., and Socher, R. (2017a).
\newblock Dynamic coattention networks for question answering.
\newblock In {\em the International Conference on Learning Representations
  (ICLR)}.

\bibitem[\protect\astroncite{Xiong et~al.}{2017b}]{XiongW2017}
Xiong, W., Droppo, J., Huang, X., Seide, F., Seltzer, M., Stolcke, A., Yu, D.,
  and Zweig, G. (2017b).
\newblock The microsoft 2016 conversational speech recognition system.
\newblock In {\em The IEEE International Conference on Acoustics, Speech and
  Signal Processing (ICASSP)}.

\bibitem[\protect\astroncite{Xiong et~al.}{2017c}]{Xiong2017DeepPath}
Xiong, W., Hoang, T., and Wang, W.~Y. (2017c).
\newblock Deeppath: A reinforcement learning method for knowledge graph
  reasoning.
\newblock In {\em Conference on Empirical Methods in Natural Language
  Processing (EMNLP)}.

\bibitem[\protect\astroncite{{Xiong} et~al.}{2017}]{XiongW2017MS}
{Xiong}, W., {Wu}, L., {Alleva}, F., {Droppo}, J., {Huang}, X., and {Stolcke},
  A. (2017).
\newblock {The Microsoft 2017 Conversational Speech Recognition System}.
\newblock {\em ArXiv e-prints}.

\bibitem[\protect\astroncite{{Xu} et~al.}{2017}]{Xu2017NTP}
{Xu}, D., {Nair}, S., {Zhu}, Y., {Gao}, J., {Garg}, A., {Fei-Fei}, L., and
  {Savarese}, S. (2017).
\newblock {Neural Task Programming: Learning to Generalize Across Hierarchical
  Tasks}.
\newblock {\em ArXiv e-prints}.

\bibitem[\protect\astroncite{Xu et~al.}{2015}]{Xu2015}
Xu, K., Ba, J.~L., Kiros, R., Cho, K., Courville, A., Salakhutdinov, R., Zemel,
  R.~S., and Bengio, Y. (2015).
\newblock Show, attend and tell: Neural image caption generation with visual
  attention.
\newblock In {\em the International Conference on Machine Learning (ICML)}.

\bibitem[\protect\astroncite{Xu et~al.}{2014}]{Xu2014}
Xu, L.~D., He, W., and Li, S. (2014).
\newblock Internet of things in industries: A survey.
\newblock {\em IEEE Transactions on Industrial Informatics}, 10(4):2233--2243.

\bibitem[\protect\astroncite{Yahya et~al.}{2016}]{Yahya2016}
Yahya, A., Li, A., Kalakrishnan, M., Chebotar, Y., and Levine, S. (2016).
\newblock Collective robot reinforcement learning with distributed asynchronous
  guided policy search.
\newblock {\em ArXiv e-prints}.

\bibitem[\protect\astroncite{Yang and Mitchell}{2017}]{Yang2017KB}
Yang, B. and Mitchell, T. (2017).
\newblock Leveraging knowledge bases in lstms for improving machine reading.
\newblock In {\em the Association for Computational Linguistics annual meeting
  (ACL)}.

\bibitem[\protect\astroncite{Yang et~al.}{2016}]{Yang2016}
Yang, X., Chen, Y.-N., Hakkani-Tur, D., Crook, P., Li, X., Gao, J., and Deng,
  L. (2016).
\newblock {End-to-End Joint Learning of Natural Language Understanding and
  Dialogue Manager}.
\newblock {\em ArXiv e-prints}.

\bibitem[\protect\astroncite{{Yang} et~al.}{2015}]{Yang2015}
{Yang}, Z., {He}, X., {Gao}, J., {Deng}, L., and {Smola}, A. (2015).
\newblock {Stacked Attention Networks for Image Question Answering}.
\newblock {\em ArXiv e-prints}.

\bibitem[\protect\astroncite{Yang et~al.}{2017}]{YangZ2017}
Yang, Z., Hu, J., Salakhutdinov, R., and Cohen, W.~W. (2017).
\newblock Semi-supervised qa with generative domain-adaptive nets.
\newblock In {\em the Association for Computational Linguistics annual meeting
  (ACL)}.

\bibitem[\protect\astroncite{Yannakakis and Togelius}{2018}]{Yannakakis2018}
Yannakakis, G.~N. and Togelius, J. (2018).
\newblock {\em Artificial Intelligence and Games}.
\newblock Springer.

\bibitem[\protect\astroncite{Yao et~al.}{2014}]{Yao2014}
Yao, H., Szepesvari, C., Sutton, R.~S., Modayil, J., and Bhatnagar, S. (2014).
\newblock Universal option models.
\newblock In {\em the Annual Conference on Neural Information Processing
  Systems (NIPS)}.

\bibitem[\protect\astroncite{Yi et~al.}{2017}]{Yi2017DualGAN}
Yi, Z., Zhang, H., Tan, P., and Gong, M. (2017).
\newblock Dualgan: Unsupervised dual learning for image-to-image translation.
\newblock In {\em the IEEE International Conference on Computer Vision (ICCV)}.

\bibitem[\protect\astroncite{Yogatama et~al.}{2017}]{Yogatama2017}
Yogatama, D., Blunsom, P., Dyer, C., Grefenstette, E., and Ling, W. (2017).
\newblock Learning to compose words into sentences with reinforcement learning.
\newblock In {\em the International Conference on Learning Representations
  (ICLR)}.

\bibitem[\protect\astroncite{Yosinski et~al.}{2014}]{Yosinski2014}
Yosinski, J., Clune, J., Bengio, Y., and Lipson, H. (2014).
\newblock How transferable are features in deep neural networks?
\newblock In {\em the Annual Conference on Neural Information Processing
  Systems (NIPS)}.

\bibitem[\protect\astroncite{Young et~al.}{2013}]{Young2013}
Young, S., Ga{\v s}i{\'c}, M., Thomson, B., and Williams, J.~D. (2013).
\newblock {POMDP}-based statistical spoken dialogue systems: a review.
\newblock {\em PROC IEEE}, 101(5):1160--1179.

\bibitem[\protect\astroncite{{Young} et~al.}{2017}]{Young2017}
{Young}, T., {Hazarika}, D., {Poria}, S., and {Cambria}, E. (2017).
\newblock {Recent Trends in Deep Learning Based Natural Language Processing}.
\newblock {\em ArXiv e-prints}.

\bibitem[\protect\astroncite{Yu et~al.}{2017}]{Yu2017}
Yu, L., Zhang, W., Wang, J., and Yu, Y. (2017).
\newblock Seqgan: Sequence generative adversarial nets with policy gradient.
\newblock In {\em the AAAI Conference on Artificial Intelligence (AAAI)}.

\bibitem[\protect\astroncite{Yu et~al.}{2009}]{Yu2009}
Yu, Y.-L., Li, Y., Szepesv{\'a}ri, C., and Schuurmans, D. (2009).
\newblock A general projection property for distribution families.
\newblock In {\em the Annual Conference on Neural Information Processing
  Systems (NIPS)}.

\bibitem[\protect\astroncite{Yun et~al.}{2017}]{Yun2017}
Yun, S., Choi, J., Yoo, Y., Yun, K., and Young~Choi, J. (2017).
\newblock Action-decision networks for visual tracking with deep reinforcement
  learning.
\newblock In {\em the IEEE Conference on Computer Vision and Pattern
  Recognition (CVPR)}.

\bibitem[\protect\astroncite{Zagoruyko and Komodakis}{2017}]{Zagoruyko2017}
Zagoruyko, S. and Komodakis, N. (2017).
\newblock Paying more attention to attention: Improving the performance of
  convolutional neural networks via attention transfer.
\newblock In {\em the International Conference on Learning Representations
  (ICLR)}.

\bibitem[\protect\astroncite{Zaremba and Sutskever}{2015}]{Zaremba2015}
Zaremba, W. and Sutskever, I. (2015).
\newblock {Reinforcement Learning Neural Turing Machines - Revised}.
\newblock {\em ArXiv e-prints}.

\bibitem[\protect\astroncite{Zhang et~al.}{2017}]{Zhang2017}
Zhang, C., Bengio, S., Hardt, M., Recht, B., and Vinyals, O. (2017).
\newblock Understanding deep learning requires rethinking generalization.
\newblock In {\em the International Conference on Learning Representations
  (ICLR)}.

\bibitem[\protect\astroncite{{Zhang} et~al.}{2017a}]{Zhang2017Speak}
{Zhang}, H., {Yu}, H., and {Xu}, W. (2017a).
\newblock {Listen, Interact and Talk: Learning to Speak via Interaction}.
\newblock {\em ArXiv e-prints}.

\bibitem[\protect\astroncite{{Zhang} et~al.}{2017b}]{Zhang2017NMT}
{Zhang}, J., {Ding}, Y., {Shen}, S., {Cheng}, Y., {Sun}, M., {Luan}, H., and
  {Liu}, Y. (2017b).
\newblock {THUMT: An Open Source Toolkit for Neural Machine Translation}.
\newblock {\em ArXiv e-prints}.

\bibitem[\protect\astroncite{{Zhang} et~al.}{2018}]{Zhang2018sentiment}
{Zhang}, L., {Wang}, S., and {Liu}, B. (2018).
\newblock {Deep Learning for Sentiment Analysis : A Survey}.
\newblock {\em ArXiv e-prints}.

\bibitem[\protect\astroncite{{Zhang} and
  {Zhu}}{2018}]{Zhang2018Interpretability}
{Zhang}, Q. and {Zhu}, S.-C. (2018).
\newblock Visual interpretability for deep learning: a survey.
\newblock {\em Frontiers of Information Technology \& Electronic Engineering},
  19(1):27--39.

\bibitem[\protect\astroncite{Zhang and Lapata}{2017}]{Zhang2017sentence}
Zhang, X. and Lapata, M. (2017).
\newblock Sentence simplification with deep reinforcement learning.
\newblock In {\em Conference on Empirical Methods in Natural Language
  Processing (EMNLP)}.

\bibitem[\protect\astroncite{{Zhang} et~al.}{2016}]{Zhang2016}
{Zhang}, Y., {Mustafizur Rahman}, M., {Braylan}, A., {Dang}, B., {Chang},
  H.-L., {Kim}, H., {McNamara}, Q., {Angert}, A., {Banner}, E., {Khetan}, V.,
  {McDonnell}, T., {Thanh Nguyen}, A., {Xu}, D., {Wallace}, B.~C., and {Lease},
  M. (2016).
\newblock {Neural Information Retrieval: A Literature Review}.
\newblock {\em ArXiv e-prints}.

\bibitem[\protect\astroncite{{Zhang} et~al.}{2017c}]{Zhang2017Speech}
{Zhang}, Y., {Pezeshki}, M., {Brakel}, P., {Zhang}, S., {Yoshua Bengio}, C.~L.,
  and {Courville}, A. (2017c).
\newblock {Towards End-to-End Speech Recognition with Deep Convolutional Neural
  Networks}.
\newblock {\em ArXiv e-prints}.

\bibitem[\protect\astroncite{Zhao and Eskenazi}{2016}]{Zhao2016}
Zhao, T. and Eskenazi, M. (2016).
\newblock Towards end-to-end learning for dialog state tracking and management
  using deep reinforcement learning.
\newblock In {\em the Annual SIGdial Meeting on Discourse and Dialogue
  (SIGDIAL)}.

\bibitem[\protect\astroncite{{Zhong} et~al.}{2017}]{Zhong2017}
{Zhong}, Z., {Yan}, J., and {Liu}, C.-L. (2017).
\newblock {Practical Network Blocks Design with Q-Learning}.
\newblock {\em ArXiv e-prints}.

\bibitem[\protect\astroncite{Zhou et~al.}{2015}]{Zhou2015}
Zhou, B., Khosla, A., Lapedriza, A., Oliva, A., and Torralba, A. (2015).
\newblock Object detectors emerge in deep scene {CNN}s.
\newblock In {\em the International Conference on Learning Representations
  (ICLR)}.

\bibitem[\protect\astroncite{{Zhou} et~al.}{2017}]{Zhou2017Emotional}
{Zhou}, H., {Huang}, M., {Zhang}, T., {Zhu}, X., and {Liu}, B. (2017).
\newblock {Emotional Chatting Machine: Emotional Conversation Generation with
  Internal and External Memory}.
\newblock {\em ArXiv e-prints}.

\bibitem[\protect\astroncite{{Zhou} and {Tuzel}}{2017}]{Zhou2017Apple}
{Zhou}, Y. and {Tuzel}, O. (2017).
\newblock {VoxelNet: End-to-End Learning for Point Cloud Based 3D Object
  Detection}.
\newblock {\em ArXiv e-prints}.

\bibitem[\protect\astroncite{Zhou}{2016}]{Zhou2016}
Zhou, Z.-H. (2016).
\newblock {\em Machine Learning (in Chinese)}.
\newblock Tsinghua University Press, Beijing, China.

\bibitem[\protect\astroncite{{Zhou} and {Feng}}{2017}]{Zhou2017}
{Zhou}, Z.-H. and {Feng}, J. (2017).
\newblock Deep forest: Towards an alternative to deep neural networks.
\newblock In {\em the International Joint Conference on Artificial Intelligence
  (IJCAI)}.

\bibitem[\protect\astroncite{Zhu et~al.}{2017a}]{Zhu2017CycleGAN}
Zhu, J.-Y., Park, T., Isola, P., and Efros, A.~A. (2017a).
\newblock Unpaired image-to-image translation using cycle-consistent
  adversarial networks.
\newblock In {\em the IEEE International Conference on Computer Vision (ICCV)}.

\bibitem[\protect\astroncite{Zhu and Goldberg}{2009}]{Zhu2009}
Zhu, X. and Goldberg, A.~B. (2009).
\newblock {\em Introduction to semi-supervised learning}.
\newblock Morgan \& Claypool.

\bibitem[\protect\astroncite{Zhu et~al.}{2017b}]{Zhu2017}
Zhu, Y., Mottaghi, R., Kolve, E., Lim, J.~J., Gupta, A., Li, F.-F., and
  Farhadi, A. (2017b).
\newblock Target-driven visual navigation in indoor scenes using deep
  reinforcement learning.
\newblock In {\em IEEE International Conference on Robotics and Automation
  (ICRA)}.

\bibitem[\protect\astroncite{Zinkevich}{2017}]{Zinkevich2017}
Zinkevich, M. (2017).
\newblock {\em Rules of Machine Learning: Best Practices for ML Engineering}.
\newblock http://martin.zinkevich.org/rules\_of\_ml/rules\_of\_ml.pdf.

\bibitem[\protect\astroncite{Zoph and Le}{2017}]{Zoph2017}
Zoph, B. and Le, Q.~V. (2017).
\newblock Neural architecture search with reinforcement learning.
\newblock In {\em the International Conference on Learning Representations
  (ICLR)}.

\bibitem[\protect\astroncite{{Zoph} et~al.}{2017}]{Zoph2017Transfer}
{Zoph}, B., {Vasudevan}, V., {Shlens}, J., and {Le}, Q.~V. (2017).
\newblock {Learning Transferable Architectures for Scalable Image Recognition}.
\newblock {\em ArXiv e-prints}.

\end{thebibliography}
\bibliographystyle{apa}

\end{document}